\documentclass{article}
\usepackage{arxiv}

\usepackage[utf8]{inputenc}
\usepackage[T1]{fontenc}
\usepackage{natbib}  
\usepackage{hyperref}
\usepackage{url}
\usepackage{booktabs}
\usepackage{amsmath}
\usepackage{amsfonts}
\usepackage{bm}
\usepackage{nicefrac}
\usepackage{microtype}
\usepackage{graphicx}
\usepackage{xcolor}
\usepackage{multirow}
\usepackage{subcaption}
\usepackage{pifont}
\usepackage{enumitem}
\usepackage{longtable}
\usepackage{placeins}

\graphicspath{{./}}

\setlist[itemize]{leftmargin=1.1em, itemsep=1pt, topsep=2pt, parsep=0pt}
\newcommand{\ci}[2]{{\scriptsize\,[$#1$,\,$#2$]}}

\title{3D-DefectBench: A Controlled Factorial Study of Vision-Language Model Evaluation Pipelines for Fine-Grained 3D Generation Defects}
\shorttitle{3D-DefectBench}
\renewcommand{\headerdate}{A Preprint}
\date{%
  {\small
  \textsuperscript{1}Roblox Corporation\\
  \textsuperscript{2}Berkeley AI Research Lab \& Department of Industrial Engineering and Operations Research, University of California, Berkeley\\
  \textsuperscript{3}Computer Science Department, Stanford University\\
  \textsuperscript{4}Division of Biostatistics, University of California, Berkeley
  }%
}

\author{%
  \begin{minipage}{\dimexpr\textwidth-2\tabcolsep\relax}\centering
  Zhenyu Zhao\textsuperscript{1,\textdagger}, \enspace
  Nanshan Jia\textsuperscript{1,2,*}, \enspace
  Jihyeon Je\textsuperscript{1,3,*}, \enspace
  Yifu Tang\textsuperscript{1,2,*}, \enspace
  Alvin Chan\textsuperscript{1}, \enspace
  Michael Spedden\textsuperscript{1}, \enspace
  Michael V. Palleschi\textsuperscript{1}, \enspace
  Sui Huang\textsuperscript{1}, \enspace
  Jingshen Wang\textsuperscript{4}, \enspace
  Zeyu Zheng\textsuperscript{2}
  \end{minipage}%
}

\begin{document}
\maketitle
\begingroup
\renewcommand{\thefootnote}{}
\footnotetext{%
\begin{tabular}{@{}l@{\hspace{0.3em}}p{0.9\textwidth}@{}}
\textsuperscript{*} & This work was done while Nanshan Jia, Jihyeon Je, and Yifu Tang were interns at Roblox Corporation.\\
\textsuperscript{\textdagger} & Correspondence: \texttt{zzhao@roblox.com}.
\end{tabular}%
}
\endgroup

\begin{abstract}
Automated evaluation is essential for scaling generative 3D systems, where exhaustive human review is costly and slow. Yet the reliability of an automated judge depends on the full evaluation pipeline---not only the underlying vision-language model (VLM), but also how the asset is rendered, which visual evidence is provided, how the task is specified, and how human reference labels are constructed. We introduce \textbf{3D-DefectBench}, a large-scale instantiation of a methodology for rigorous evaluation-pipeline analysis. The benchmark complements holistic ratings and pairwise preferences with nine fine-grained binary defects spanning geometry, texture, and prompt adherence---each additionally carrying an optional human \emph{severity} (Major) annotation---providing actionable diagnostics for generator development and detailed visibility into judge behavior. Using a balanced factorial design, we vary the VLM, camera protocol, visual input, and prompt schema across 84 inference designs, then use a cost-aware staged study to validate the resulting design conclusions on a broader frontier-model set. Model choice is the dominant source of variation in agreement with human labels, but the remaining pipeline factors also influence agreement, interact with the model, and can alter the best configuration for a given judge. Within the tested design space, a compact six-view RGB protocol performs comparably to denser view sets and configurations augmented with depth or normal channels, making it a strong cost-effective default. Under this fixed design, the best of 12 VLMs still trail trained human labelers, while texture agreement drops sharply when moving from expert-agreement labels to noisier silver labels. The severity annotations further let us show that binary judge calls recover most of the defects humans flag as \emph{severe} (recall rises) even though a binary protocol cannot grade severity on its own. These results show that automated judges should be evaluated as complete pipelines and calibrated across human reference regimes, rather than benchmarked solely as standalone models. We release labels, prompts, predictions, and Croissant metadata on \href{https://huggingface.co/datasets/zzhao0500/3D-DefectBench}{HuggingFace}.
\end{abstract}


\FloatBarrier

\section{Introduction}
\label{sec:intro}

Evaluating generative 3D systems at scale is difficult. Both the input and the output are open ended: a text prompt can describe an effectively unbounded set of objects, while the resulting textured mesh may vary in geometry, appearance, composition, and adherence to the prompt. In production and model-development settings, evaluation may cover large volumes of generations, repeated checkpoint comparisons, online failure analysis, and even annotation of training data for evaluators or downstream models. Exhaustive human review at this scale is costly and slow, motivating the use of vision-language models (VLMs) as automated judges that inspect multi-view renders together with the generation prompt \citep{duggal2025eval3d,wu2024gpt,zhang2025hi3deval}. However, evaluating VLM judges is fundamentally a measurement problem, not simply a model benchmarking problem.

Using a VLM as a judge involves more than choosing a VLM model. Human labelers can inspect a 3D asset interactively by rotating and zooming the mesh, whereas a VLM receives a constructed representation of that asset. Its agreement with human labels may therefore depend not only on the VLM's capability, but also on how the asset is rendered, which visual channels are shown, how the task and output format are specified, and how the reference labels are constructed. Evaluating judges under different settings can consequently confound model quality with differences in the surrounding evaluation setup. We therefore treat a VLM judge as a configurable \emph{evaluation pipeline} and study how each component contributes to agreement with human labels.

Evaluating a VLM judge against humans is itself a measurement problem. A simple approach is to compare each VLM prediction with a single human annotation, but this implicitly assumes that the annotation is a reliable target. In practice, agreement can vary with annotator expertise, rubric interpretation, and example difficulty, so a low VLM--human agreement score may reflect limitations in the reference labels as well as in the judge. To separate these sources of variation, we construct two complementary reference regimes: a large \textbf{silver} set annotated by trained vendor labelers and a smaller \textbf{expert} set labeled by the 3D artists who developed the rubric. Each asset receives multiple independent annotations. This design lets us estimate inter-labeler agreement, compare VLMs under silver and expert references, and examine whether judge performance deteriorates on defects or examples that humans also label inconsistently.

Reliable calibration alone, however, is not sufficient. Many existing 3D-generation evaluations summarize an asset with a holistic rating or pairwise preference \citep{Gu_2025_CVPR,yang2024llplace3dindoorscene,yang2024scenecraft,yang2024holodeck}. Such signals remain valuable for ranking generators and summarizing overall quality, but they provide limited insight into why a generation succeeds or fails and can therefore be difficult to act on during model development. A system can improve its average preference score while producing more fused parts, misplaced textures, or missing prompt-specified components---failures that require different data and modeling interventions. We therefore complement holistic evaluation with fine-grained defect assessment. Given a text prompt and a generated textured mesh, the judge predicts a nine-dimensional binary vector spanning geometry, texture, and prompt adherence. Human labelers additionally mark each flagged defect as \emph{Major (Severe)} or not, so the reference labels carry a per-defect severity signal alongside the binary flags---a dataset feature we use to test whether judges preferentially recover the defects humans deem serious (Section~\ref{sec:severity}). These defect-level judgments provide actionable diagnostics for generator improvement, enable more detailed quantification of model performance, and reveal which categories of defects VLM judges can and cannot assess reliably.

Studying these questions requires varying pipeline components jointly. Camera protocol, visual input, prompt schema, and judge model may interact: a rendering or prompting choice that helps one model may hurt another. We therefore adopt a balanced factorial design that crosses the practitioner-controllable pipeline factors and estimates both their main effects and interactions with a defect-level logistic factor model. Because evaluating every configuration with every frontier VLM would be prohibitively expensive, we further employ a staged experimental design. We first screen all 84 inference configurations using a diverse set of cost-effective VLMs, then validate the resulting design recommendations on a broader set of frontier models, before fixing a near-optimal pipeline configuration for comprehensive comparison against expert labels, held-out silver labels, and trained human labelers.

\begin{figure}[tbp]
\centering
\includegraphics[width=\linewidth]{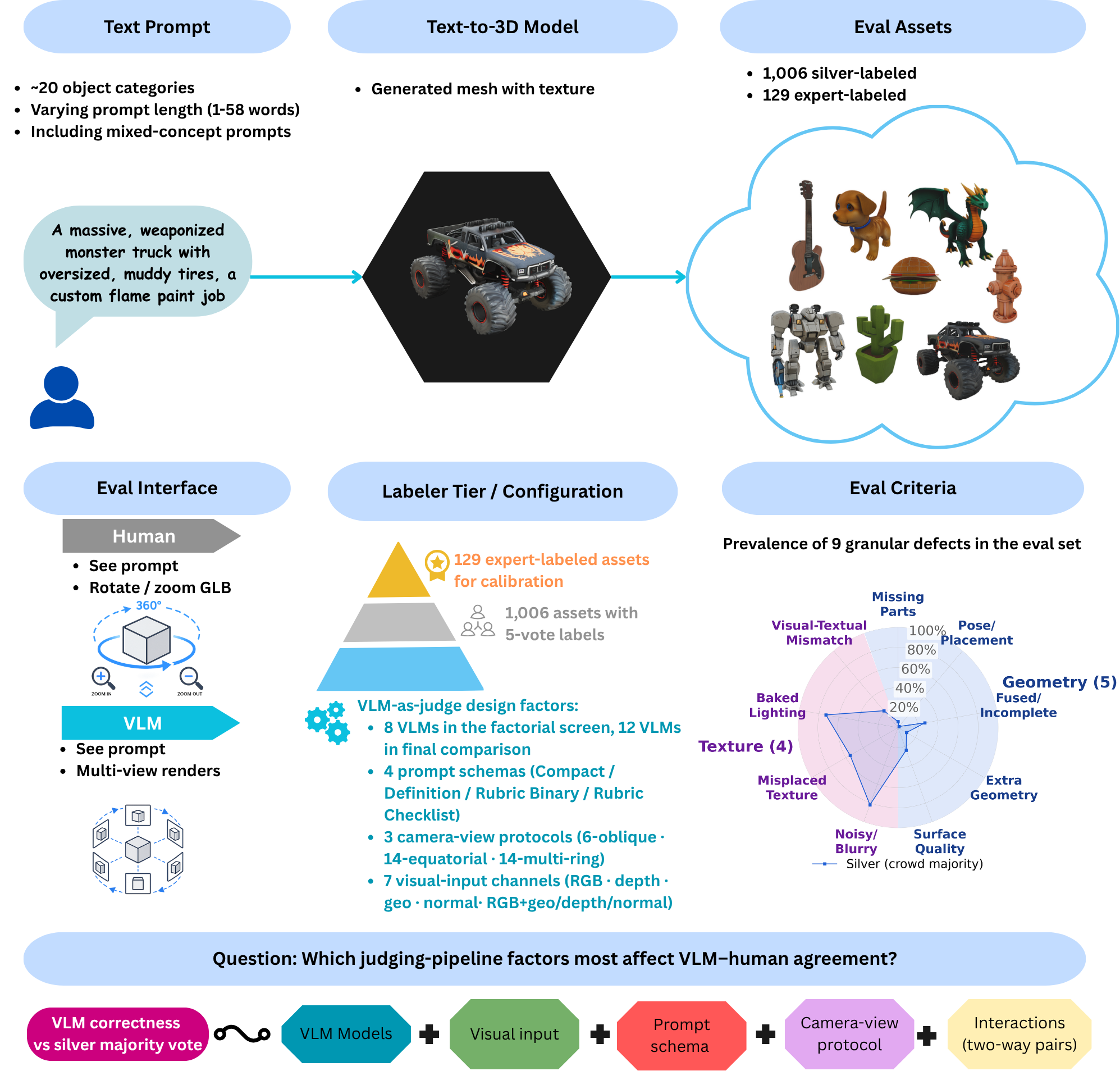}
\caption{\textbf{Overview of 3D-DefectBench.} Text prompts are converted into textured GLB meshes by two anonymized text-to-3D generator arms. Human reviewers inspect interactive meshes, whereas VLM judges receive rendered multi-view images together with the prompt. We treat the VLM judge as a configurable evaluation pipeline, systematically varying the judge model, camera protocol, visual input, and prompt schema through a balanced factorial design before evaluating frontier VLMs under a fixed near-optimal pipeline. Multiple annotations from trained vendor and expert labelers provide complementary reference-label regimes for calibration.}
\label{fig:overview}
\end{figure}

This study is enabled by \textbf{3D-DefectBench}, a benchmark designed specifically for studying VLM judges rather than merely ranking them. It contains text prompts, generated textured meshes, and fine-grained defect annotations from complementary human reference sources. The benchmark includes a large silver set labeled by trained vendor annotators and a smaller expert set of 129 assets independently reviewed by two expert 3D artists. From an initial pool of 1,049 collected assets we retain the 1,006 with complete five-vote silver relabeling; this 1,006-asset silver set is the released benchmark and the basis for all analyses in the remainder of the paper. Assets are drawn from multiple text-to-3D generator arms, enabling us to study judge reliability, calibration across reference-label regimes, and the stability of downstream generator-comparison conclusions.

We organize the study around three questions:
\begin{enumerate}[leftmargin=1.5em,itemsep=1pt,topsep=2pt]
\item \textbf{Pipeline sensitivity:} How much do the judge model, camera protocol, visual input, and prompt schema---and their interactions---affect agreement with human labels?
\item \textbf{Deployment design:} Which evaluation pipeline provides the best reliability--cost trade-off, and do those conclusions transfer across judges and to a different 3D generator?
\item \textbf{Human calibration:} Under a fixed pipeline, how closely do VLM judges agree with trained humans, how robust are their rankings across silver and expert reference regimes, and do they support consistent downstream generator-comparison decisions?
\end{enumerate}

Our contributions are:
\begin{itemize}
\item \textbf{A fine-grained 3D defect benchmark and evaluation task.} We introduce a nine-category defect taxonomy spanning geometry, texture, and prompt adherence, together with silver and expert reference labels and a per-defect human \emph{severity} (Major) annotation. The resulting benchmark complements holistic quality metrics by providing actionable diagnostics for generator development and detailed characterization of VLM judges.

\item \textbf{A factorial framework for studying VLM evaluation pipelines.} We jointly vary the judge model, camera protocol, visual input, and prompt schema across 84 inference designs and use a defect-level logistic factor model to estimate their main effects and interactions, treating automated judging as a configurable measurement pipeline rather than a standalone model.

\item \textbf{A cost-aware staged methodology for evaluating automated judges.} We combine broad factorial screening with cost-effective VLMs, targeted validation on a wider frontier-model set, and comprehensive fixed-pipeline comparison. This separates the question of which evaluation pipeline to deploy from which VLM to use.

\item \textbf{Calibration against human and reference-label variability.} We evaluate 12 VLM judges against expert and silver reference regimes, benchmark them against trained human labelers under matched protocols, analyze defect-level strengths and weaknesses, and study how judge choice influences downstream generator-comparison conclusions.
\end{itemize}

Although our empirical study focuses on text-to-3D generation, we believe the underlying methodology is broadly applicable to GenAI evaluation. In many domains, an automated judge operates as a configurable evaluation pipeline whose reliability depends jointly on the judge model, input construction, prompt design, output schema, and the quality of human reference labels. The combination of balanced factorial experiments, staged evaluation, multiple reference-label regimes, and calibration against human labelers provides a general framework for studying automated judges beyond 3D generation.

\section{Related work}
\label{sec:related}

\paragraph{Automated evaluation with LLM/VLM judges.}
LLM- and VLM-based judges automate evaluation in text, code, and multimodal tasks \citep{NEURIPS2024_5a7c9475,lin2024evaluating,liu2023gevalnlgevaluationusing,NEURIPS2023_91f18a12}. These methods raise a common measurement concern: a judge's score can conflate model capability with prompt design, input representation, and label quality. This concern is central in 3D generation, where the visual evidence is not a natural image but a rendered representation of a mesh. We therefore treat rendering, view layout, prompt format, and output schema as part of the evaluated judging system rather than fixed implementation details. A parallel line of work studies the \emph{reliability} of LLM/VLM judges themselves---position and verbosity biases, self-consistency, and calibration \citep{NEURIPS2023_91f18a12,lin2024evaluating}; we complement it by measuring, for a 3D defect-judging task, how much of a judge's agreement is governed by pipeline choices rather than the underlying model.

\paragraph{Visual evaluator benchmarks.}
MMBench and MMMU assess multimodal reasoning \citep{liu2024mmbench,yue2024mmmu}; Q-Bench targets low-level visual perception \citep{wu2024qbench}; VISCO benchmarks fine-grained, step-wise binary critique of visual-reasoning chains \citep{wu2025visco}; GenAI-Bench and VQAScore target 2D text-to-visual generation \citep{li2024genaibench,lin2024evaluating}. These benchmarks primarily compare model capability under a fixed input and evaluation protocol. Of them, VISCO is closest to our fine-grained binary labeling in spirit---it assigns a binary correctness label per reasoning step---though it critiques natural-image reasoning chains rather than defects on rendered 3D assets. In contrast, our benchmark evaluates VLMs inside a configurable 3D judging pipeline and reports how system-level choices affect defect-level accuracy.

\paragraph{3D generation evaluation.}
Prior work scores text-to-3D and image-to-3D outputs with a range of protocols rather than a single VLM-judge recipe: VLM/MLLM judges producing pairwise or hierarchical scores \citep{wu2024gpt,zhang2025hi3deval}, interpretable multi-criterion metrics computed from foundation models \citep{duggal2025eval3d}, dedicated non-VLM metrics motivated by concerns that direct GPT scoring inherits model biases \citep{su2024gt23d}, trained human-preference reward models fit to large crowd-plus-expert vote collections \citep{zhang20253dgen}, and learned rank metrics fit to human mean-opinion scores with GPT-4o used only for prompt construction \citep{t23dcompbench2025}. Several of these collect substantial human annotation for validation, and 3DGen-Bench in particular gathers both crowd and expert votes \citep{zhang20253dgen}. CLIP alignment \citep{pmlr-v139-radford21a} and 2D perceptual metrics \citep{zhang2018perceptual,NIPS2017_8a1d6947} are not designed for localized geometry/texture defects. Earlier text-to-shape retrieval and text-driven mesh stylization methods~\citep{text2shape,text2mesh} provide complementary alignment signals but likewise do not target per-defect judgments. Two choices set 3D-DefectBench apart from this body of work. \emph{First, the unit of measurement.} Recent 3D evaluators have moved toward localized and categorical assessment: Eval3D enumerates inconsistency \emph{types}---structural, semantic, and geometric inconsistency, and text--3D misalignment~\citep{duggal2025eval3d}---and Hi3DEval scores geometry and texture quality at the \emph{part} level, localizing flaws below the whole-object level~\citep{zhang2025hi3deval}; others report multi-dimensional quality on absolute scales, such as T23D-CompBench's twelve sub-components~\citep{t23dcompbench2025} and DB-3DME's coarse 1--3 ordinal mesh ratings~\citep{jia2026db}. Our unit is finer and more categorical still: a \emph{binary} label for each of nine independently defined failure modes---five geometry (form/surface quality, fused or incomplete parts, pose/placement, missing parts, extra geometry) and four texture---recording \emph{whether} each specific defect is present rather than rating quality on a continuous or ordinal scale. This defect-vector granularity, rather than localization per se (which Hi3DEval and Eval3D already provide in coarser or continuous form), is what reduces judge evaluation to a per-defect classification problem. \emph{Second, the object of study.} Prior 3D benchmarks are aimed primarily at ranking \emph{generators}, and where they examine the judging pipeline they do so as a validation step for a chosen configuration: GPTEval3D, for instance, ablates a few judge-side inputs---how GPT-4V is prompted and its outputs ensembled, the number of views shown at once, and RGB versus surface-normal renders---to justify its final setup~\citep{wu2024gpt}, while other evaluators ablate their own learned scoring components~\citep{zhang2025hi3deval,t23dcompbench2025}. We instead hold the generated assets fixed and make the \emph{judge pipeline} the object of a systematic factorial study, varying rendering style, camera-view layout, visual input, and prompt/output schema to measure how much defect-level agreement depends on pipeline design rather than the model alone. The per-defect binary taxonomy makes this possible by turning judge evaluation into a classification problem with a well-defined agreement metric, enabling attribution to specific pipeline factors and their interactions. Large-scale silver labels, expert annotations on the same schema, broad model comparisons, and a held-out trained human evaluated on the same per-defect metric further validate and stress-test the analysis. The benchmark is distinguished primarily by its localized defect unit and its treatment of the judge as a measurement system.

Taken together, prior work establishes automated judging, multimodal benchmarking, and structured 3D evaluation, often validating individual judge-design choices in service of ranking generators. Our contribution is to study the judging pipeline itself. By scoring localized binary defects rather than aggregate quality, we systematically quantify how rendering, view layout, visual input, prompt schema, and output format affect agreement. Staged model validation, multiple human-reference regimes, and calibration against trained human labelers then connect this analysis to the practical question of which evaluation pipeline to trust.

\section{3D-DefectBench: fine-grained evaluation task and human reference systems}
\label{sec:dataset}

\begin{table}[t]
\caption{Canonical accounting for splits, API calls, and scored defect cells. Screening scores all 1{,}006 silver assets on every design: geometry on all 84 configs and texture on the 48 RGB-bearing configs (color is required to judge texture). The 500/506-asset figures used for configuration selection and the silver holdout are a disjoint partition of these 1{,}006 assets with complete five-vote relabeling, not a limit on screening coverage. The final row is the both-experts-agree subset of the \emph{Expert eval} (\texttt{c004}) predictions above---no additional inference---counted as unique reference cells rather than per-model scored cells, so the per-run columns do not apply.}
\label{tab:accounting}
\centering
\footnotesize
\begin{tabular}{lrrrrrr}
\toprule
Phase & Assets & Configs & Models & Aspects & Calls & Scored cells \\
\midrule
Screening (factor analysis) & 1,006 & 84 & 8 & 2 & 1,170,984 & 5,359,982 \\
Silver holdout eval & 506 & 1 & 12 & 2 & 12,144 & 54,648 \\
Expert eval (c004) & 129 & 1 & 12 & 2 & 3,096 & 13,932 \\
Expert agreement-only cells & 129 & --- & --- & --- & --- & 877 \\
\bottomrule
\end{tabular}
\end{table}

\paragraph{Task and taxonomy.}
Each example is a text prompt and a generated textured GLB mesh. Human labelers inspect the mesh interactively with rotation and zoom; VLM judges receive rendered views plus the prompt. The target is a nine-dimensional binary defect vector: five geometry defects (form/surface quality, fused or incomplete parts, pose/placement, missing parts, and extra geometry) and four texture defects (noise/blur, misplaced or overlapping texture, baked lighting/shadow artifacts, and visual-textual mismatch). Missing parts, pose/placement, and visual-textual mismatch are prompt-conditioned because the labeler must compare the asset to the text prompt. The released tables use stable machine-readable column names; Appendix~\ref{app:taxonomy} maps them to the reader-facing labels used here.

\begin{figure}[tbp]
\centering
\includegraphics[width=\linewidth]{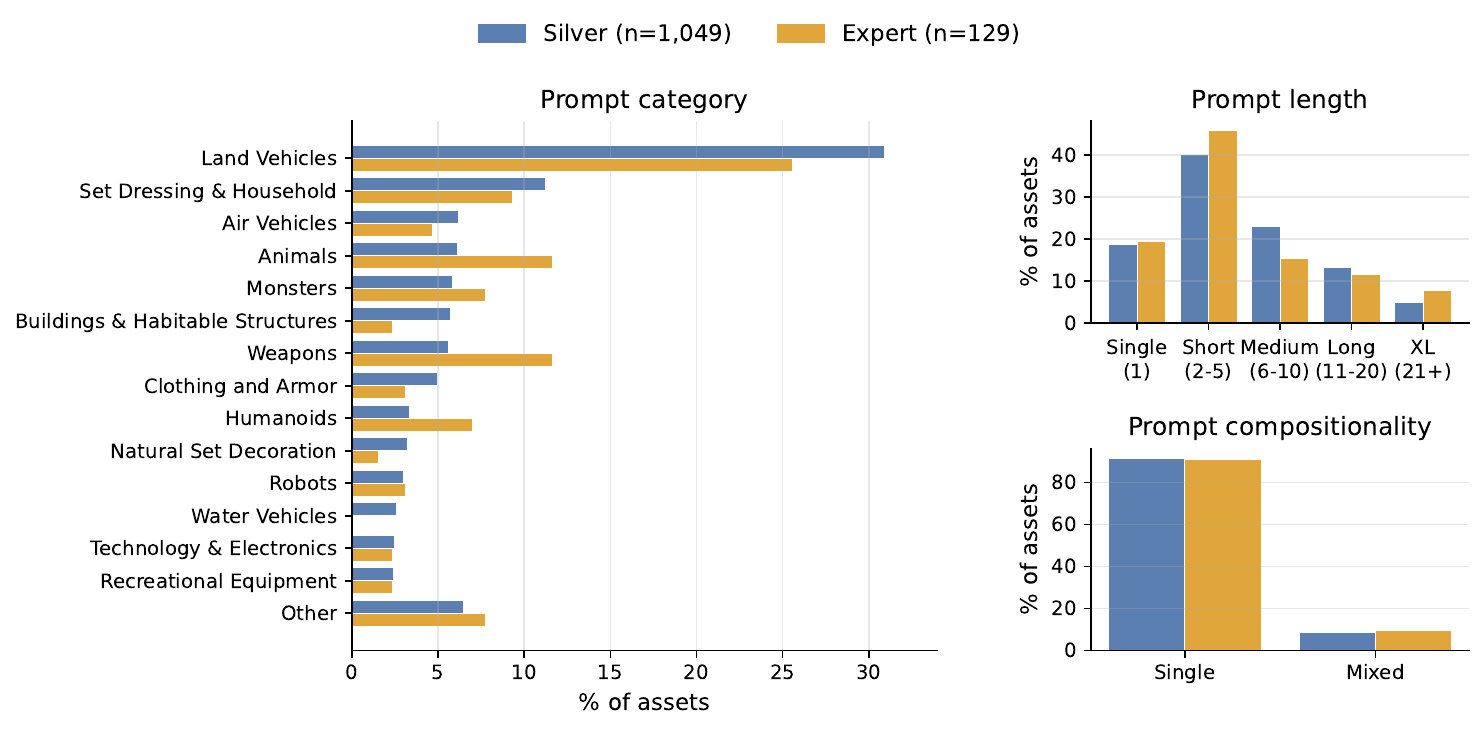}
\caption{\textbf{Prompt distribution across the silver and expert splits.} \textbf{Left:} prompt category (the 14 largest categories by silver share; less-frequent categories are grouped into ``Other''). \textbf{Top-right:} prompt length in words. \textbf{Bottom-right:} prompt compositionality (single- vs.\ mixed-concept). Both splits are dominated by short (2--5 word), single-concept prompts describing everyday game objects, with land vehicles the most common category; the silver and expert splits have closely matched distributions on all three axes.}
\label{fig:prompt_dist}
\end{figure}

\paragraph{Generator provenance.}
Assets are generated by two variants of Roblox’s Cube text-to-3D system~\citep{roblox2025cube}, referred to throughout the benchmark as models A and B. We resample the generated assets to construct the evaluation set, so its defect distribution is designed to support judge-pipeline evaluation and should not be interpreted as representative of the generators’ production performance. Prompts are short object descriptions, with median lengths of 24 and 14 characters in the silver and expert splits, respectively. Figure~\ref{fig:prompt_dist} shows that their category, length, and compositionality distributions are closely matched across splits. Benchmark conclusions are therefore scoped to this sampled generator mix and to evaluating the judging pipeline, rather than to estimating the performance of any single production system.

\paragraph{Labeler pool and training.}
Two distinct populations produce the reference labels. The \textbf{expert} labelers are two 3D artists embedded with the generation-modeling team, who routinely perform 3D-generation evaluation tasks; they authored the defect rubrics used throughout the benchmark and trained the silver labelers against them. The \textbf{silver} labelers are vendor annotators who work with the team and were trained and calibrated on the same expert-authored rubric. Crucially, both populations are experienced rather than first-time raters for this task: prior to the data used here, the same labelers completed earlier pilot evaluations, calibration rounds, and a formal evaluation on a separate model-checkpoint-selection effort. The released labels therefore reflect trained, calibrated annotators rather than a cold-start crowd.

\paragraph{Splits and labeling.}
The \textbf{silver} split comprises \textbf{1{,}006} generated assets with complete five-vote relabeling by trained vendor annotators, retained from an initial pool of 1{,}049 collected assets (the funnel is detailed below). This evaluation set was formed by a sampling procedure over the generator outputs---prior to VLM evaluation---that reshapes the defect-prevalence distribution relative to the raw generator population. Silver ground truth is the per-defect majority vote across labelers. These labels support the silver leave-one-labeler-out benchmark in Section~\ref{sec:human} on the 506-asset validation holdout. The \textbf{expert} split contains \textbf{129} assets with two labels per asset from the expert 3D reviewers; a small number of assets also appear in the silver split and are kept as independent labeling passes. To control label quality, the silver labeling was carried out in two passes---an initial pooled pass (at least three labels per asset; mean 3.91, maximum 13, pooling annotations across the evaluation runs in which an asset was reused) and a subsequent five-vote relabeling of the 1{,}049 collected assets---and a controlled comparison confirms the second pass is the higher-quality one, raising both inter-labeler agreement and alignment with the expert golden, most clearly on texture. We retain the \textbf{1{,}006} assets with complete five-vote coverage as the released silver benchmark and use these improved five-vote labels for all analyses in the paper; we further verified that every design conclusion we report is unchanged between the two label sets.

\begin{table}[tbp]
\caption{Per-defect silver prevalence ($n{=}1{,}006$ assets), mean crowd-label agreement, and chance-corrected inter-annotator agreement: silver Fleiss' $\kappa$ ($\kappa_{\mathrm{silver}}$) and expert Cohen's $\kappa$ ($\kappa_{\mathrm{exp}}$, $n{=}129$). Texture $\kappa$ is markedly lower than geometry---the chance-corrected form of the texture label-noise ceiling. Fleiss' $\kappa$ deflates under extreme class imbalance (pose/placement); a prevalence-robust Randolph free-marginal $\kappa$ is in Appendix~\ref{app:repro}.}
\label{tab:perdef}
\centering
\footnotesize
\begin{tabular}{llrrrr}
\toprule
Aspect & Defect & Prevalence & Mean agr. & $\kappa_{\mathrm{silver}}$ & $\kappa_{\mathrm{exp}}$ \\
\midrule
Geometry & Missing parts & 0.060 & 0.904 & 0.198 & 0.380 \\
Geometry & Pose or placement mismatch & 0.015 & 0.954 & 0.132 & 0.485 \\
Geometry & Fused or incomplete parts & 0.271 & 0.814 & 0.281 & 0.424 \\
Geometry & Extra geometry & 0.097 & 0.929 & 0.441 & 0.566 \\
Geometry & Form or surface quality & 0.239 & 0.775 & 0.145 & 0.380 \\
\midrule
Texture & Noise, blur, or grain & 0.819 & 0.755 & 0.054 & 0.359 \\
Texture & Misplaced or overlapping texture & 0.552 & 0.748 & 0.168 & 0.284 \\
Texture & Baked lighting or shadow & 0.731 & 0.772 & 0.155 & 0.277 \\
Texture & Visual-textual mismatch & 0.220 & 0.820 & 0.256 & 0.316 \\
\bottomrule
\end{tabular}
\end{table}

\begin{figure}[tbp]
\centering
\begin{subfigure}{\linewidth}\centering
\includegraphics[width=\linewidth]{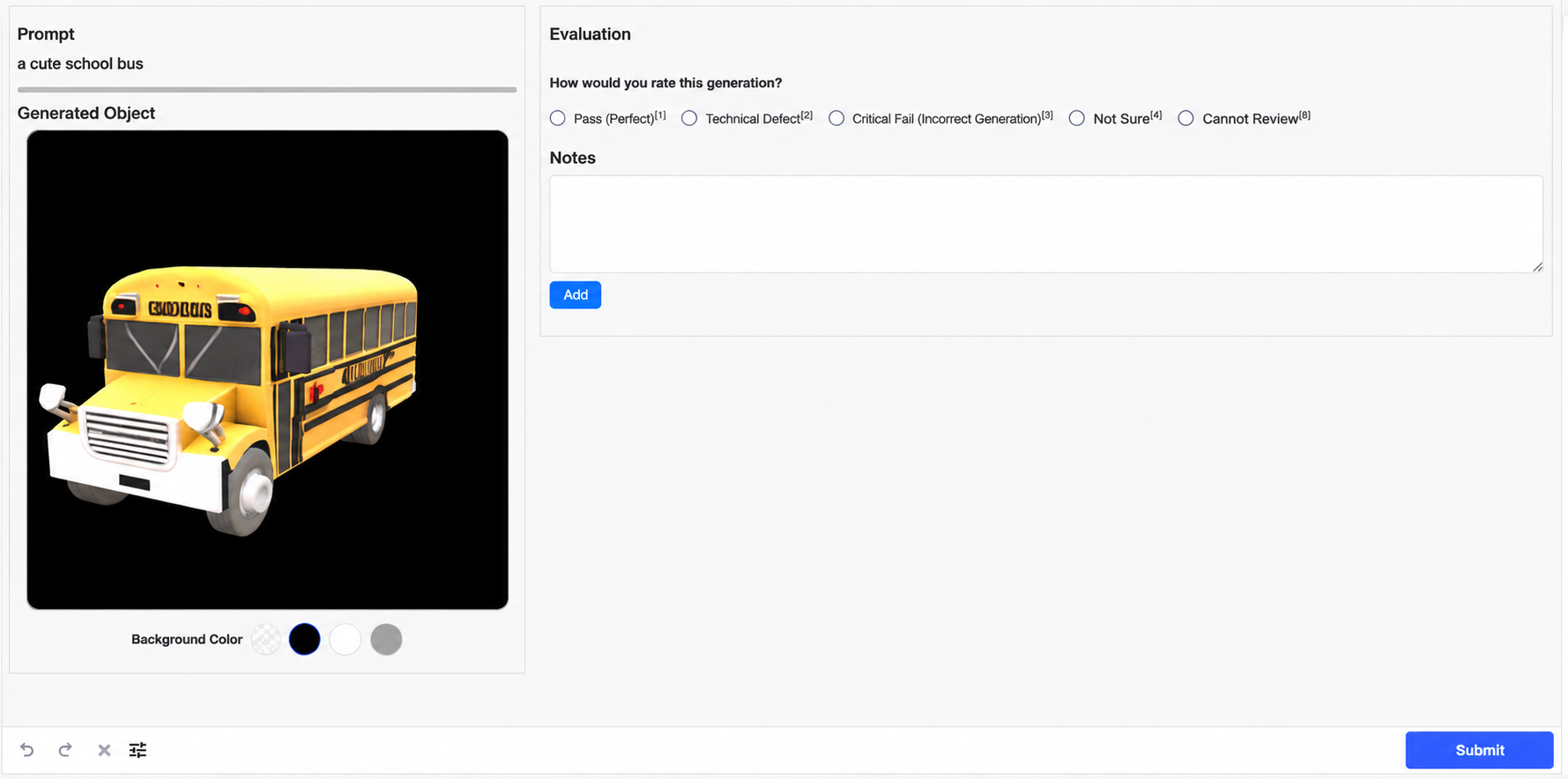}
\caption*{\scriptsize (a) Initial state: the labeler orbits/zooms the textured mesh and picks one top-level verdict.}
\end{subfigure}

\vspace{3pt}
\begin{subfigure}{\linewidth}\centering
\includegraphics[width=\linewidth]{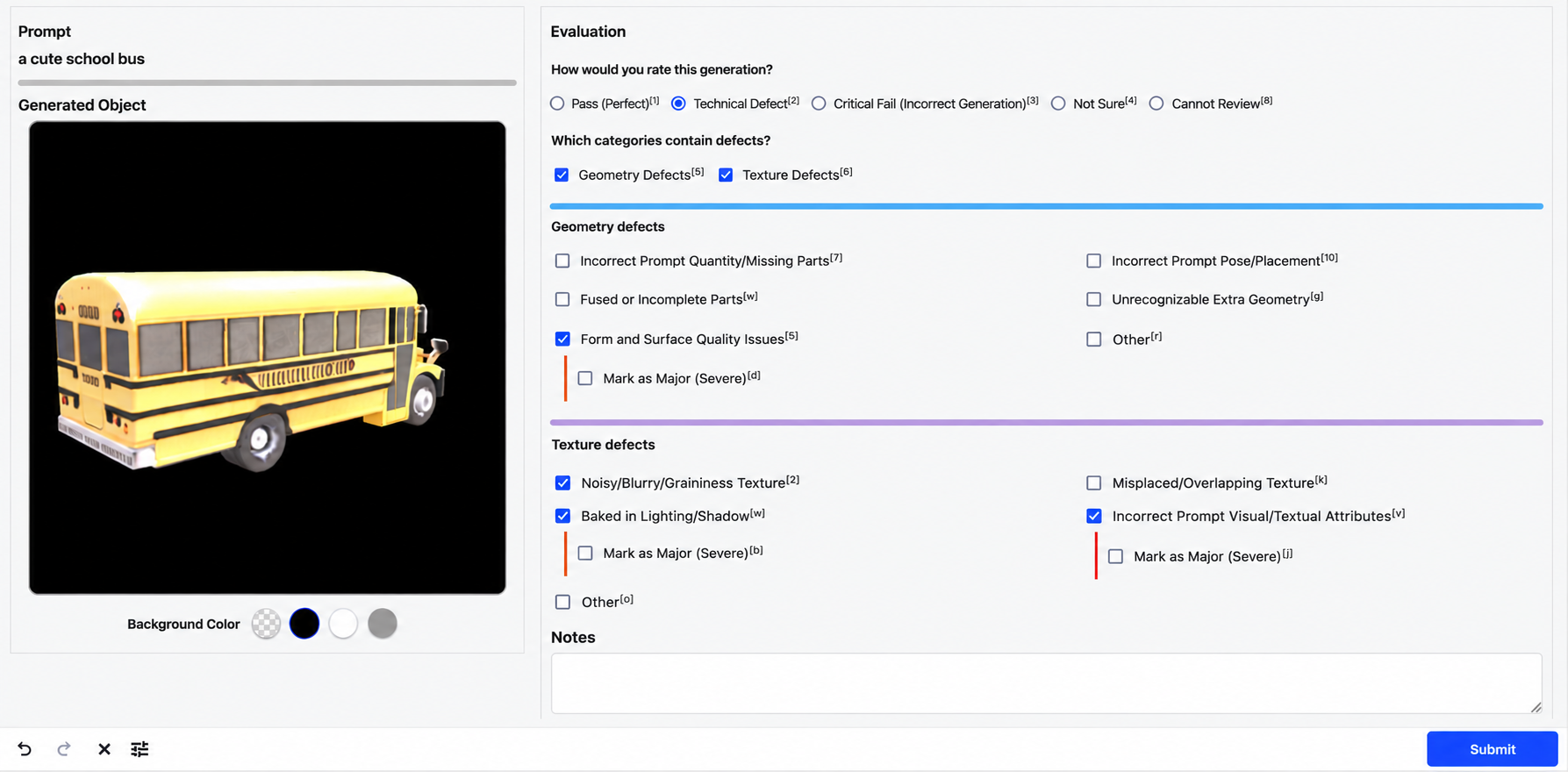}
\caption*{\scriptsize (b) After selecting \emph{Technical Defect}, geometry and texture checklists expand; checking a defect reveals its severity toggle.}
\end{subfigure}
\caption{Interactive GLB labeling UI used for crowd and expert annotation. Labelers orbit, zoom, and toggle each defect category independently. \textbf{(a)} The labeler first selects one top-level verdict (\emph{Pass}, \emph{Technical Defect}, \emph{Critical Fail}, \emph{Not Sure}, \emph{Cannot Review}). \textbf{(b)} Choosing \emph{Technical Defect} progressively discloses the geometry- and texture-defect checklists; each flagged defect then exposes a \emph{Mark as Major (Severe)} toggle, so every submission carries a per-defect binary flag \emph{and} an optional severity signal. The one exception is \emph{Noisy/Blurry/Graininess Texture}, which has no severity toggle: it is treated as inherently minor (anything appearance-critical should instead be recorded under another defect).}
\label{fig:ui}
\end{figure}

\paragraph{Labeling interface.}
  Figure~\ref{fig:ui} shows the human labeling UI: labelers view the textured mesh with orbit controls, rotate and zoom freely, and mark each
  defect category independently. A submission marked \emph{Pass (Perfect)} sets all defect labels to $0$, whereas \emph{Critical Failure} flags a
  generation that is unrecognizable with respect to its prompt and so cannot receive granular defect labels; \emph{Cannot Review} indicates the object is not loaded in the UI and \emph{Not
  Sure} flags the labeler is uncertain to score, and labelers are trained to use \emph{Not Sure} sparingly. These non-granular
  verdicts are dropped at the \emph{submission} level---we retain only per-labeler submissions marked as a defect or as \emph{Pass}---not at the
  asset level: an asset is removed only when too few granular submissions remain (below the per-tier label threshold of three for silver and two
  for expert assets), so an asset can still enter the evaluation set when a minority of its labelers marked it critical or unsure.
  Crucially, the interface collects \emph{two} signals per defect: a binary \emph{present/absent} flag and, for every defect except
  Noisy/Blurry/Graininess Texture, an optional \emph{Mark as Major (Severe)} toggle (Figure~\ref{fig:ui}b). The entire benchmark below is scored
  against the \emph{binary} per-defect majority vote; we exploit the severity signal only once, in Section~\ref{sec:severity}, to test whether VLM
  judges recover the subset of defects that human majorities deem severe.

\section{A factorial framework for analyzing VLM evaluation pipelines}
\label{sec:design}
\begin{figure}[tbp]
\centering
\begin{subfigure}{\linewidth}
\centering
\begin{subfigure}{0.32\linewidth}\includegraphics[width=\linewidth]{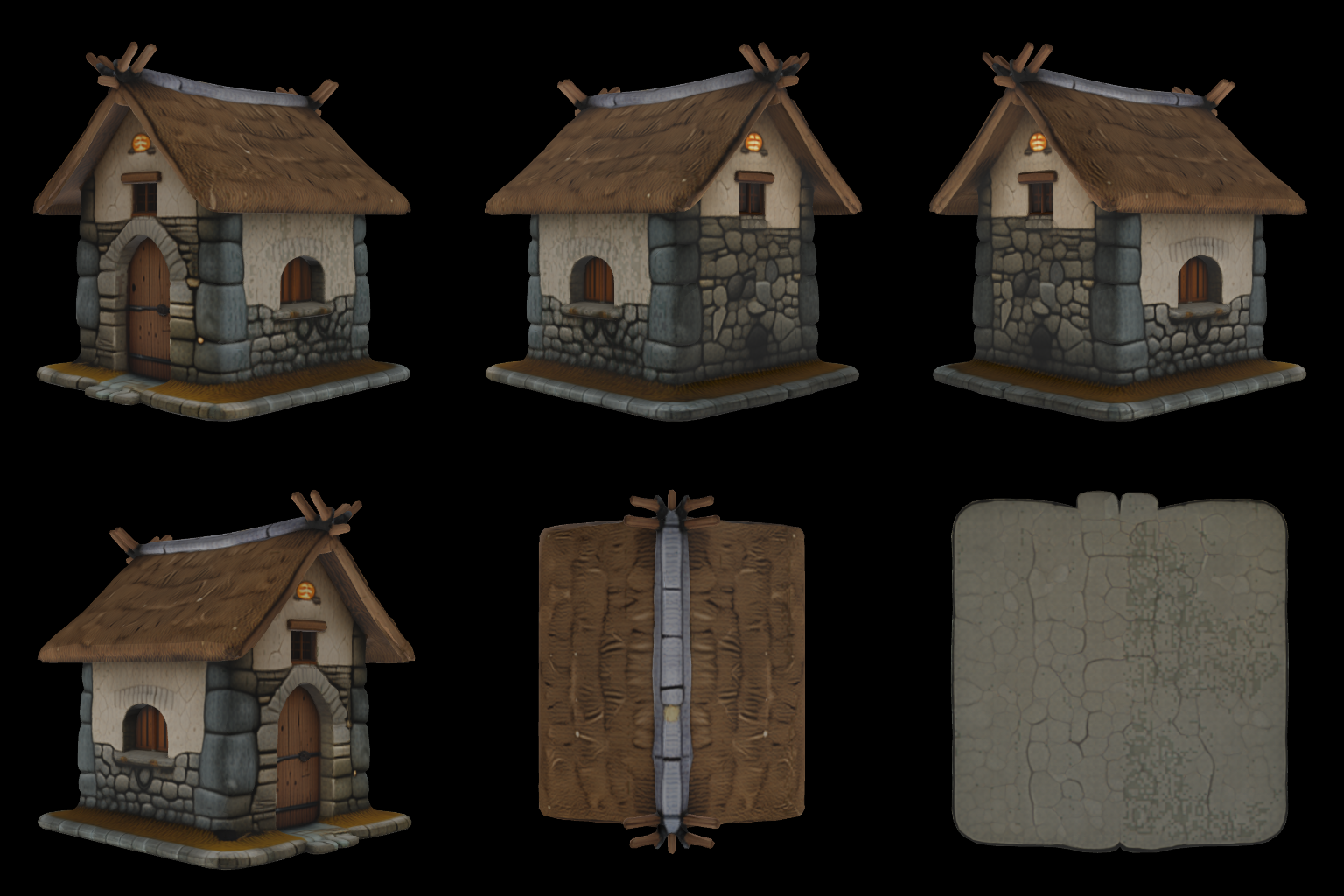}\caption*{\scriptsize 6-view oblique}\end{subfigure}\hfill
\begin{subfigure}{0.32\linewidth}\includegraphics[width=\linewidth]{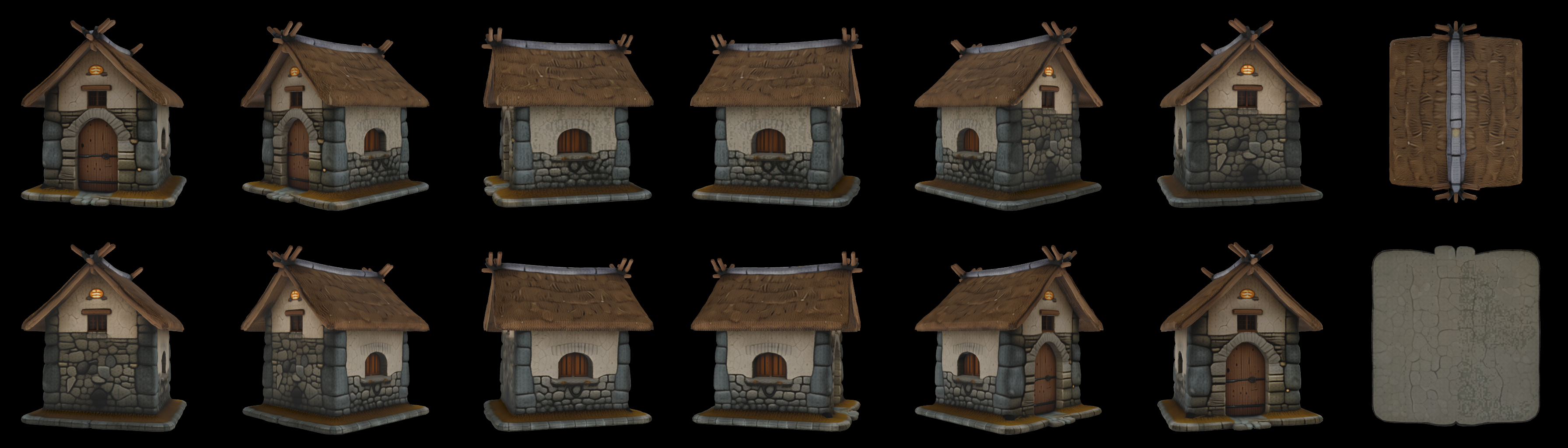}\caption*{\scriptsize 14-view equatorial}\end{subfigure}\hfill
\begin{subfigure}{0.32\linewidth}\includegraphics[width=\linewidth]{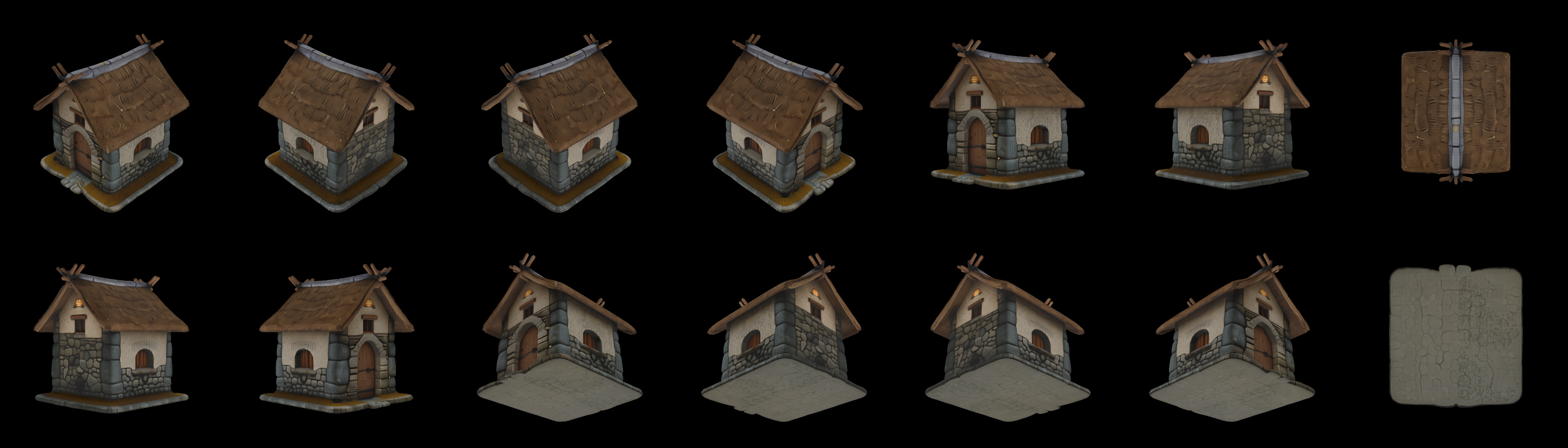}\caption*{\scriptsize 14-view multi-ring}\end{subfigure}
\caption*{\textbf{(a) Camera-view protocols} (RGB).}
\end{subfigure}

\vspace{2pt}
\begin{subfigure}{\linewidth}
\centering
\begin{subfigure}{0.24\linewidth}\includegraphics[width=\linewidth]{figures/fig_example_cam_6oblique.png}\caption*{\scriptsize RGB}\end{subfigure}\hfill
\begin{subfigure}{0.24\linewidth}\includegraphics[width=\linewidth]{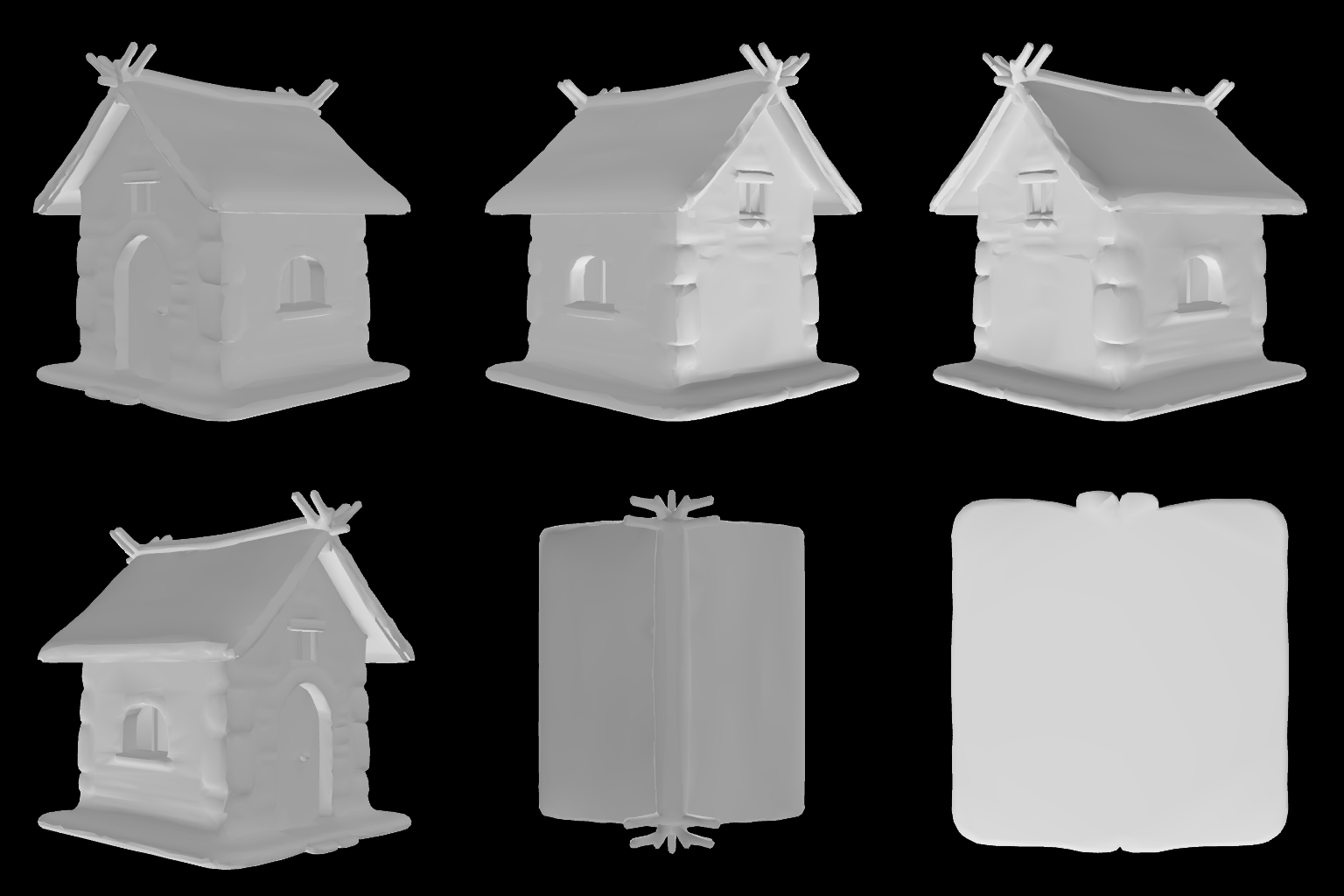}\caption*{\scriptsize geometry-only}\end{subfigure}\hfill
\begin{subfigure}{0.24\linewidth}\includegraphics[width=\linewidth]{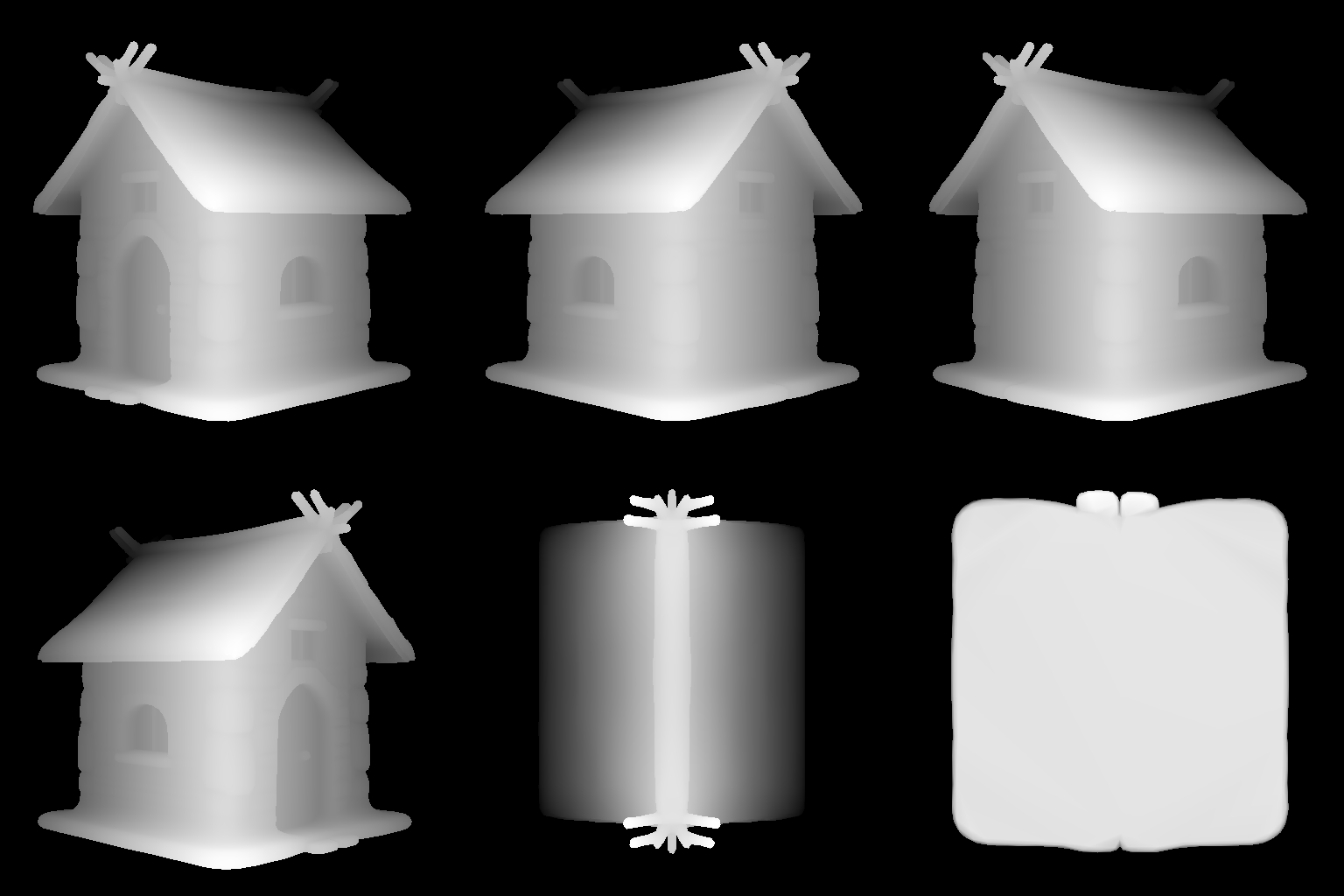}\caption*{\scriptsize depth}\end{subfigure}\hfill
\begin{subfigure}{0.24\linewidth}\includegraphics[width=\linewidth]{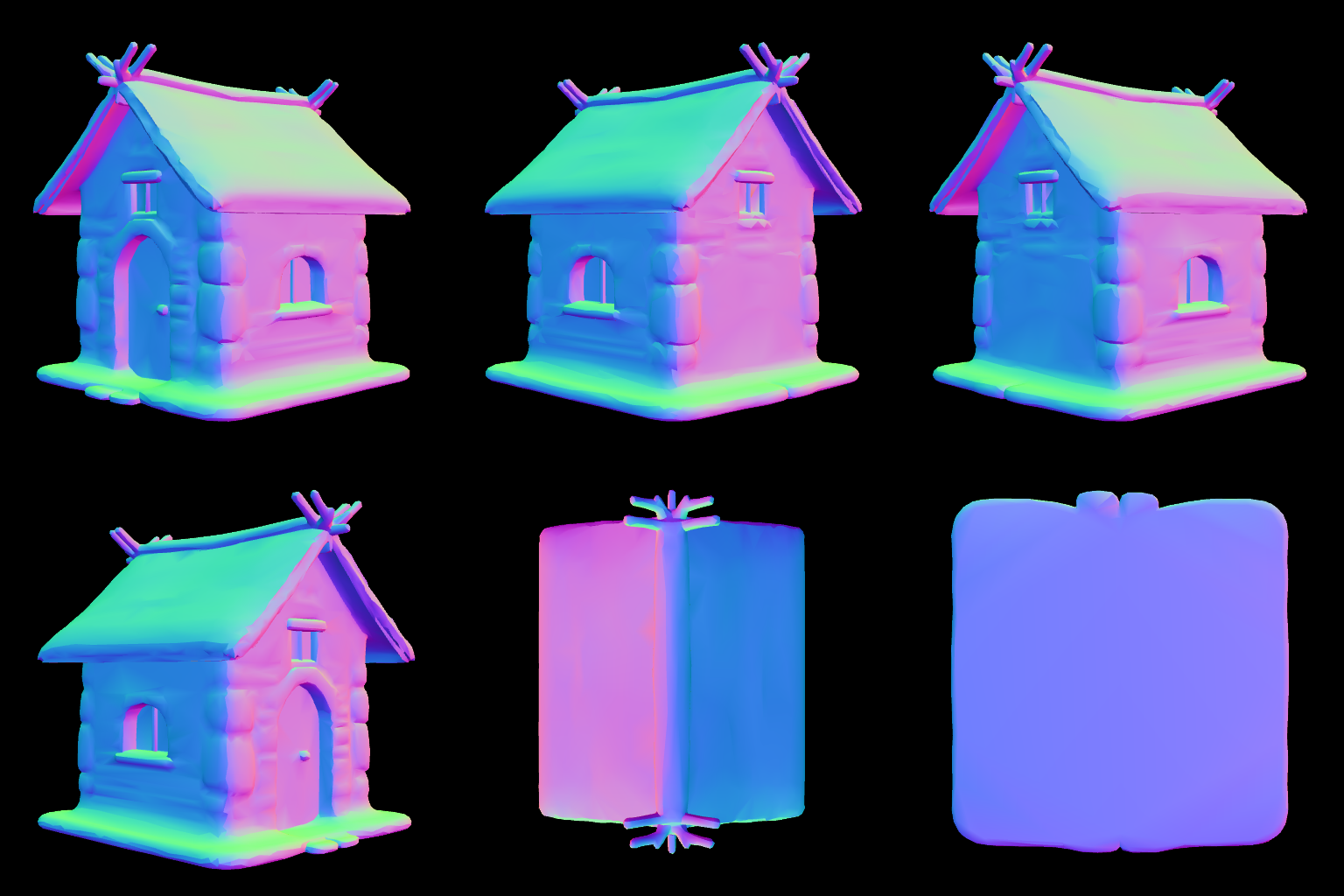}\caption*{\scriptsize normal map}\end{subfigure}
\caption*{\textbf{(b) Visual-input channels} (six-view oblique).}
\end{subfigure}
\caption{\textbf{Visual evidence factors on one example asset} (a stone cottage with a thatched roof). (a) The three camera-view protocols; (b) the four base visual-input channels rendered with the six-view oblique protocol. Each panel is the single grid image passed to the VLM.}
\label{fig:example_inputs}
\end{figure}

We operationalize the pipeline view by varying four components that together turn a 3D asset into a structured defect judgment: the VLM model and three practitioner-controllable choices---camera-view protocol, visual input, and prompt schema. Their full crossing forms a grid of \textbf{84} inference designs:
\begin{itemize}
\item \textbf{Camera-view protocol} (3 levels): (i) a compact \emph{six-view oblique} turntable---four oblique equatorial views spaced around the asset plus one top-down and one bottom-up view; (ii) a \emph{14-view equatorial} turntable---twelve equatorial views plus a top-down and bottom-up view; and (iii) a \emph{14-view multi-ring} turntable---oblique views sampled across multiple elevation rings plus a top-down and bottom-up view. All protocols pack the views into a single multi-view grid image (Figure~\ref{fig:example_inputs}).
\item \textbf{Visual input} (7 levels): RGB renders alone or paired with geometry-only, depth, or normal-map channels; depth-only; geometry-only; or normal-map-only views.
\item \textbf{Prompt schema} (4 levels): Compact Binary, Definition-Guided Binary, Rubric-Guided Binary, Rubric-Guided Checklist (Table~\ref{tab:prompt_schemas}).
\end{itemize}
Depth and normal channels are visualization encodings, not metric buffers: depth is per-frame min--max normalized over the visible mesh and shown inverted (white near, black far), and normals are view-space, flat-shaded, mapped by $(\mathbf{n}\cdot 0.5+0.5)$; encoding details are in Appendix~\ref{app:repro}.
The full mapping from released configuration IDs to these human-readable descriptions is in Appendix~\ref{app:config_grid}. Figure~\ref{fig:example_inputs} illustrates the camera-view protocols and visual-input channels on a single example asset. We refer to the configuration carried forward into model comparison as \texttt{c004} (six-view oblique RGB turntable with a rubric-guided checklist prompt); the factor analysis in Section~\ref{sec:stage1} motivates this choice as a well-performing default rather than a uniquely optimal design. We pack all views into one grid image so each judgment is a single-image call with a fixed token cost and consistent cross-view alignment, rather than a multi-image or multi-turn exchange whose cost and context handling vary by model.

\paragraph{Rendering details.}
All views are rendered offscreen with \texttt{pyrender}/\texttt{trimesh} at $512\times512$ per view and packed into a single grid (six-view oblique $\to$ $3\times2$, $1536\times1024$). Each mesh is centered at the origin, normalized to a unit sphere, and framed by an auto-fit perspective camera ($\mathrm{yfov}{=}30^\circ$, unit aspect) with a $1.10$ distance margin held fixed across the views of a protocol. Lighting is flat ambient only (no directional light or HDRI) on a black background; images are 8-bit sRGB PNGs with no explicit tonemapping. These settings were held fixed for every configuration; we did not vary background or lighting, so sensitivity of judge agreement to those choices is left to future work. Full renderer parameters and per-protocol camera angles are in Appendix~\ref{app:repro}.

\paragraph{Prompt schemas.}
The prompt-schema factor has four levels, applied in parallel to the geometry and texture aspects. All four share an identical preamble (evaluation scope, per-view inspection instruction, and the binary marking rule) and differ only in how much rubric detail and structure they add; Table~\ref{tab:prompt_schemas} summarizes them. The schemas form an increasing ladder of guidance, from a one-line definition per defect (Compact Binary) to a full rubric with named failure examples emitted as a per-defect checklist (Rubric-Guided Checklist, used by \texttt{c004}). The full text of the selected geometry prompt is reproduced in Appendix~\ref{app:prompts}.

\begin{table}[h]
\caption{The four prompt schemas in the screening grid (geometry/texture parallel). Schemas progressively add rubric detail; \texttt{c004} uses Rubric-Guided Checklist. Character counts are for the geometry prompt.}
\label{tab:prompt_schemas}
\centering
\footnotesize
\begin{tabular}{p{0.22\linewidth}p{0.46\linewidth}rr}
\toprule
Schema & Design intent & Output & Chars \\
\midrule
Compact Binary & One-sentence definition per defect category; minimal guidance. & Single-line rating string & 3,324 \\
Definition-Guided Binary & Adds an explicit ``examples of what to count'' bullet list per category. & Single-line rating string & 5,221 \\
Rubric-Guided Binary & Full inclusion criteria plus concrete named failure examples per category. & Single-line rating string & 7,551 \\
Rubric-Guided Checklist & Same rubric as rubric-guided binary, emitted as a per-defect checklist. & Per-defect checklist & 8,176 \\
\bottomrule
\end{tabular}
\end{table}

\paragraph{Model factor.}
  The factor-analysis phase uses a full factorial design over all 84 inference designs, scored by
  eight VLMs chosen to span distinct model families and capability tiers rather than to enumerate an exhaustive model list: Gemini~3.1~Flash-Lite,
  GPT-5~Mini, Qwen~3.5~397B-A17B, Mistral~Small~3.1~24B, GPT-5.4, Gemini~3.1~Pro, Claude~Sonnet~4.6, and Claude~Haiku~4.5. From this analysis we
  carry forward one well-performing configuration (\texttt{c004}) and, conditional on that fixed design, broaden the model set for a more thorough
  comparison: the model-comparison phase evaluates 12 VLMs by adding Gemini~2.5~Pro, Claude~Opus~4.7, GPT-4o, and Qwen2.5-VL-7B, and additionally probes
  repeated-run stability across seeds.

\paragraph{Metrics.}
The primary metric is \textbf{macro Matthews correlation coefficient (MCC)} over the five geometry or four texture defects, computed on successfully parsed predictions. MCC is robust to class imbalance and is informative for rare defects such as pose errors. We also report macro F1. Silver-label analyses use crowd majority vote. For the expert split we report \emph{two complementary targets} rather than a single one: agreement on cells where \emph{both} experts agree, which provides clean, strong defect signals (877 unique cells: 539 geometry, 338 texture); and an \emph{either-expert union} target, where a defect is positive if either expert flags it (1{,}161 cells), which captures softer defect identification and gives full coverage on the disagreement cells where two labelers admit no majority (expert label modes are detailed in Appendix~\ref{subsec:expert_modes}). Bootstrap standard errors and 95\% confidence intervals are clustered by asset. The defect-level factor model that turns this grid into importance estimates is defined at the head of Section~\ref{sec:stage1}, where it is used.

\section{Screening the evaluation-pipeline design space: which factors govern reliability?}
\label{sec:stage1}

This phase is a full factorial sweep of all 84 inference designs on the 1{,}006 silver assets, scored with eight VLMs that span open-weight and closed-weight families and several capability tiers ($\sim$5.4M scored defect cells; geometry on all 84 configs, texture on the 48 RGB-bearing configs; full accounting in Table~\ref{tab:accounting}). We use them as a representative cross-section of judge types and ask: \emph{which design factors---and which interactions---most affect how well a VLM judge agrees with human labels?} We answer this with the defect-level logistic factor model of Eq.~\eqref{eq:factor}, then use the result to motivate a default configuration for model comparison. Because the analysis is observational over a finite grid, we read it as prioritizing engineering choices, not as causal claims beyond the tested designs.

\paragraph{The factor model.}
For each defect $d$ we fit a logistic regression whose binary dependent variable $y_{di}=1$ when the VLM prediction matches the reference (silver majority) label on parse-ok row $i$. Writing the four factor families as dummy-coded design blocks---VLM model ($\mathbf{m}$), camera-view protocol ($\mathbf{c}$), visual input ($\mathbf{v}$), and prompt schema ($\mathbf{p}$)---the model is
\begin{equation}
\operatorname{logit}\,\Pr(y_{di}{=}1)
= \beta_{0}
+ \bm{\beta}_{m}^{\top}\mathbf{m}_i
+ \bm{\beta}_{c}^{\top}\mathbf{c}_i
+ \bm{\beta}_{v}^{\top}\mathbf{v}_i
+ \bm{\beta}_{p}^{\top}\mathbf{p}_i
+ \!\!\sum_{(f,g)\in\mathcal{P}}\!\! \bm{\beta}_{fg}^{\top}\,(\mathbf{f}_i\!\otimes\!\mathbf{g}_i),
\label{eq:factor}
\end{equation}
where each factor block is dummy-coded against a held-out reference level, and $\mathcal{P}$ is the set of six two-way factor pairs. We rank each factor family by its McFadden partial pseudo-$R^2$---the drop in fit when that family's design block is removed---and test it with an omnibus likelihood-ratio test; every confidence interval we rely on uses an asset-cluster bootstrap. Full estimation, multiple-testing, and clustering details are in Appendix~\ref{subsec:stats}.

\subsection{The model dominates, but pipeline factors are not negligible}

\begin{figure}[tbp]
\centering
\includegraphics[width=\linewidth]{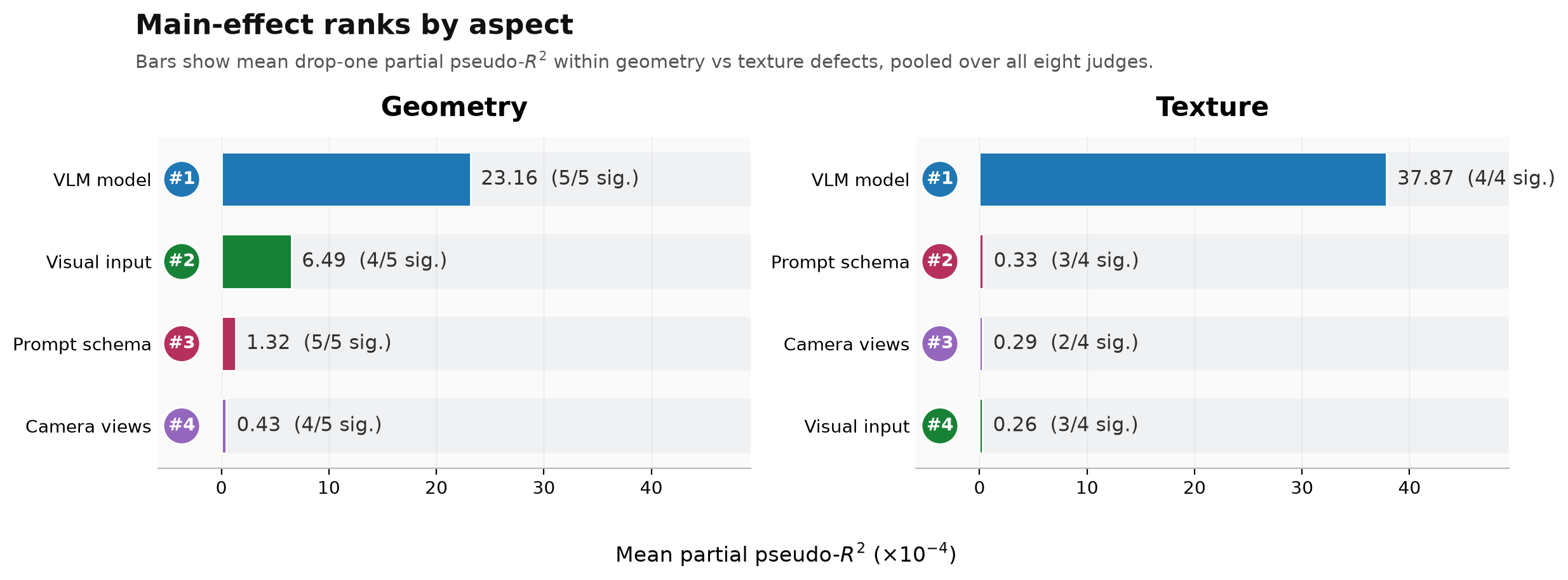}
\caption{\textbf{Main-effect factor importance.} Mean partial pseudo-$R^2$ per factor family by aspect (main effects only, interaction block excluded). VLM model dominates, but visual input and prompt schema remain non-negligible.}
\label{fig:stage1_main_effect_rank}
\end{figure}

\begin{figure}[tbp]
\centering
\includegraphics[width=\linewidth]{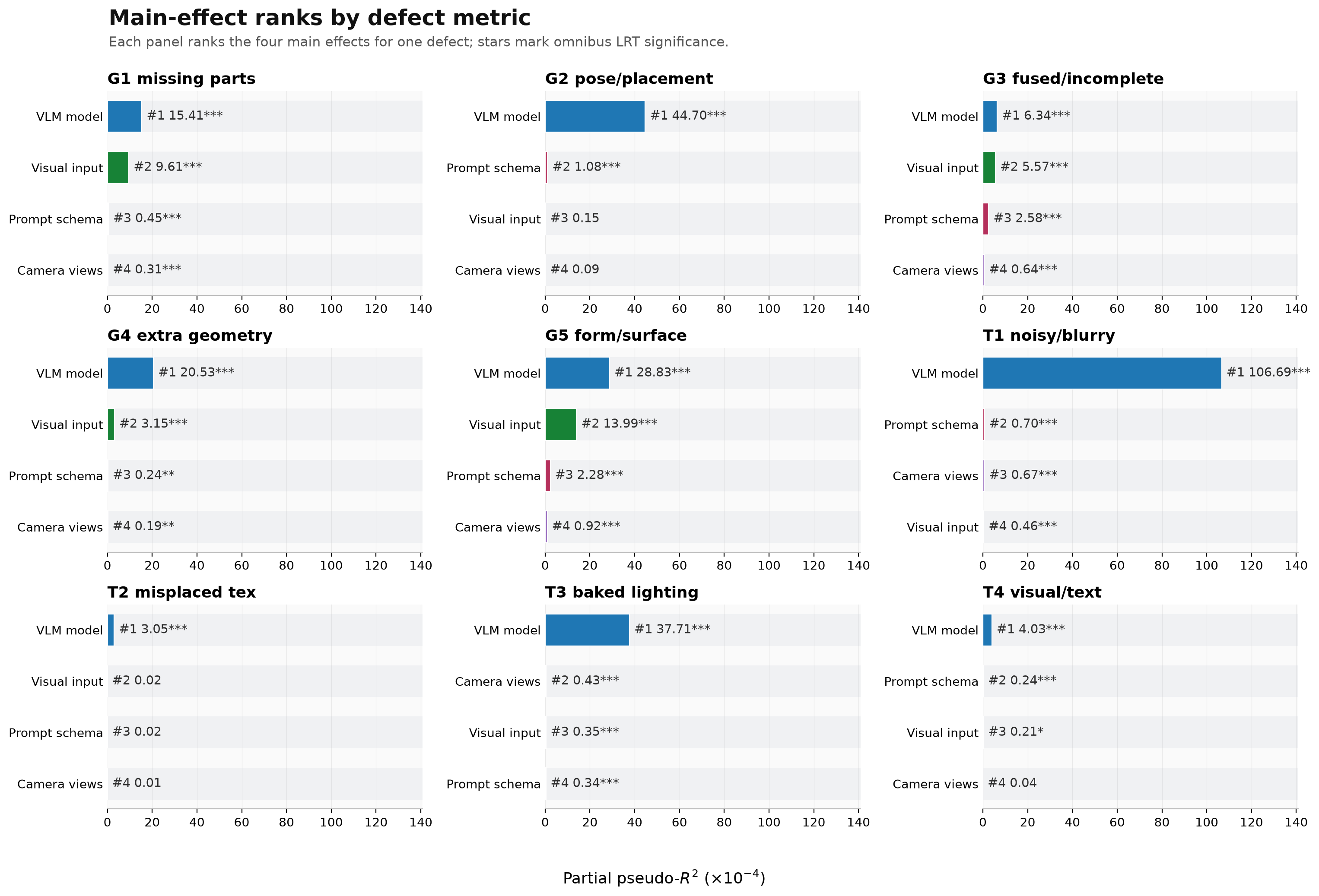}
\caption{\textbf{Per-defect factor-family importance.} Partial pseudo-$R^2$ per defect (main effects only); the family ordering (VLM model $\gg$ visual input, prompt schema, camera protocol) holds for most individual defects rather than being an artifact of macro-averaging.}
\label{fig:stage1_factor_importance}
\end{figure}

\begin{figure}[tbp]
\centering
\includegraphics[width=\linewidth]{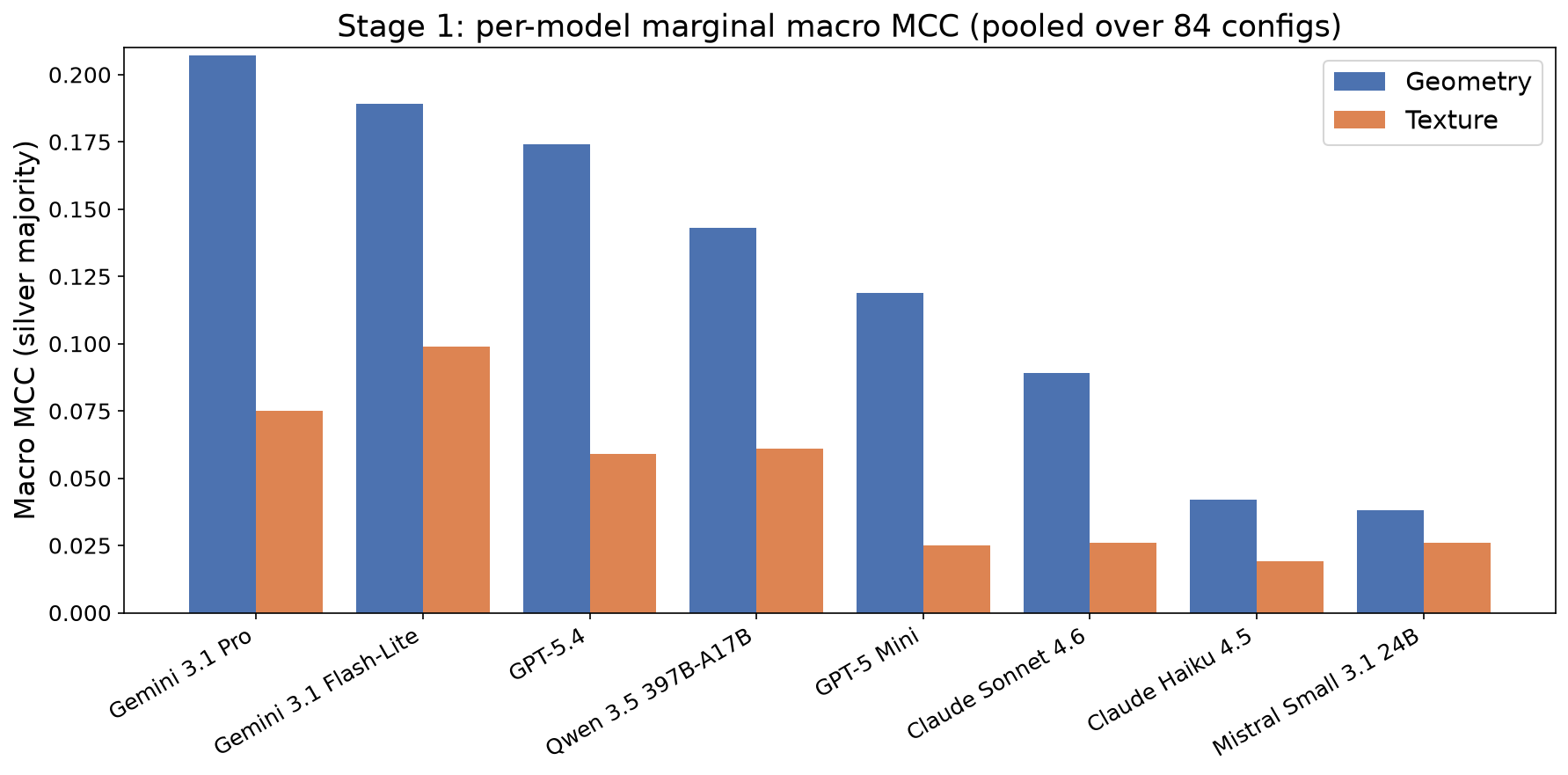}
\caption{Per-model marginal macro MCC pooled over all 84 inference designs (geometry) and the 48 RGB-bearing designs (texture), on the silver majority labels. Values match Table~\ref{tab:stage1_models}.}
\label{fig:stage1_models}
\end{figure}

\emph{Model choice explains far more agreement variance than any single pipeline factor---but the pipeline factors are not negligible.} Ranking the four main-effect families by partial pseudo-$R^2$ and omnibus likelihood-ratio tests, \textbf{VLM model} explains the most variance (mean partial pseudo-$R^2$ $3.0\times10^{-3}$), followed by \textbf{visual input}, \textbf{prompt schema}, and \textbf{camera-view protocol} (Figure~\ref{fig:stage1_main_effect_rank}). These main effects are significant for 9/9, 7/9, 8/9, and 6/9 defects at $p{<}0.05$, with identical counts after Benjamini--Hochberg FDR correction across 45 omnibus tests; the per-defect breakdown is in Figure~\ref{fig:stage1_factor_importance}. The same ordering appears in raw per-model marginals: Gemini~3.1~Pro leads on geometry (0.207), ahead of Gemini~3.1~Flash-Lite (0.189) and GPT-5.4 (0.174), while texture marginals stay low across the board---the best judge (Gemini~3.1~Flash-Lite) reaches only 0.099 (Figure~\ref{fig:stage1_models}, Table~\ref{tab:stage1_models}). Operationally, the best-vs-worst \emph{level} gap is $\sim$0.17 macro MCC for VLM model (geometry), $\sim$0.02 for both visual input and camera protocol, and $\sim$0.01 for prompt schema (Table~\ref{tab:effect_sizes}).

\begin{table}[tbp]
\caption{Per-model marginal macro MCC in the factor-analysis phase, on the silver majority labels (geometry pooled over all 84 inference designs; texture over the 48 RGB-bearing designs). All eight judges are scored on the identical 84-config grid.}
\label{tab:stage1_models}
\centering
\footnotesize
\begin{tabular}{lrr}
\toprule
Model & Geometry & Texture \\
\midrule
Gemini 3.1 Pro & 0.207 & 0.075 \\
Gemini 3.1 Flash-Lite & 0.189 & 0.099 \\
GPT-5.4 & 0.174 & 0.059 \\
Qwen 3.5 397B-A17B & 0.143 & 0.061 \\
GPT-5 Mini & 0.119 & 0.025 \\
Claude Sonnet 4.6 & 0.089 & 0.026 \\
Claude Haiku 4.5 & 0.042 & 0.019 \\
Mistral Small 3.1 24B & 0.038 & 0.026 \\
\bottomrule
\end{tabular}
\end{table}

\subsection{Color is essential; rubric guidance helps the hard geometry defects}

\begin{figure}[tbp]
\centering
\includegraphics[width=\linewidth]{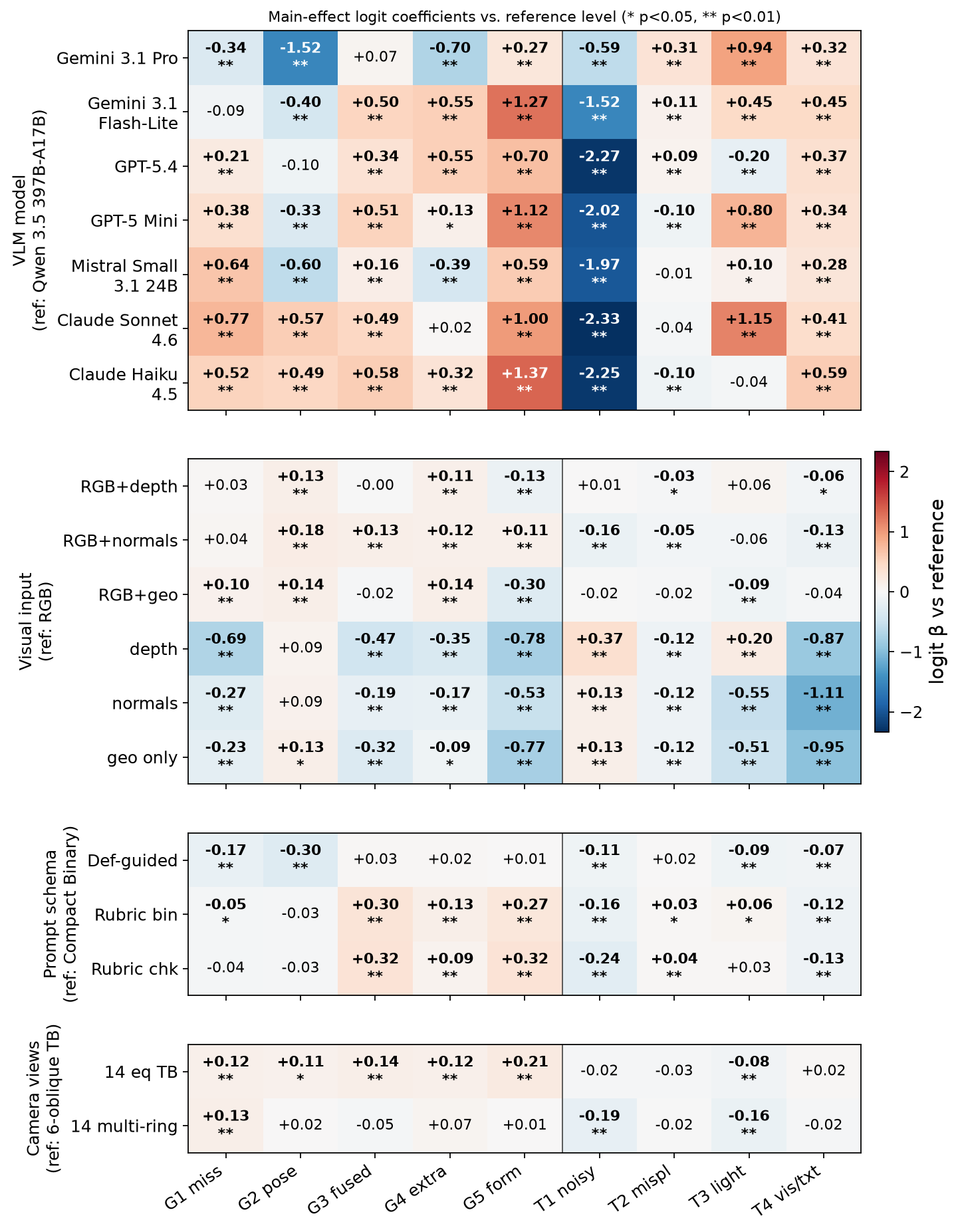}
\caption{\textbf{Main-effect logit coefficients} from Eq.~\eqref{eq:factor}, per factor family (rows) and defect (columns; G1--G5 geometry, T1--T4 texture). Positive (red) values increase agreement vs.\ the reference level; \texttt{*}/\texttt{**} mark bootstrap significance at $p{<}0.05$/$0.01$. VLM-model reference is Qwen~3.5~397B-A17B.}
\label{fig:stage1_coef}
\end{figure}

\emph{RGB is worth far more than any added geometric channel, and rubric detail gives a small but consistent lift on the hardest geometry defects.} Beyond family importance, the fitted coefficients show \emph{which} level of each factor helps or hurts each defect. Figure~\ref{fig:stage1_coef} reports the main-effect logit coefficients relative to the reference level. Two patterns stand out. First, on geometry defects the color-free inputs (depth-only, normals-only, geometry-only) carry strongly negative
  coefficients---stripping color hurts geometry agreement---whereas RGB augmented with a geometry or normals channel is roughly neutral to mildly
  helpful. This likely reflects the reference itself: the silver labels are produced by human annotators inspecting the \emph{textured} (RGB)
  render, so a judge shown only geometric channels is scored against evidence the humans never used, depressing agreement independent of any
  change in judging ability. In other words, part of RGB's advantage may be \emph{input-condition alignment} with the reference rather than perception alone; disentangling the two would require a human-side modality ablation, which we leave to future work (Section~\ref{sec:limitations}). Second, the rubric-guided prompts add small but consistently positive coefficients on the harder geometry defects (fused/incomplete, extra geometry, form/surface) relative to the compact binary prompt. Texture coefficients are uniformly small, consistent with the low texture ceiling.

\subsection{Factors interact with the judge: the precise optimum can be judge-specific}

\begin{figure}[tbp]
\centering
\includegraphics[width=0.7\linewidth]{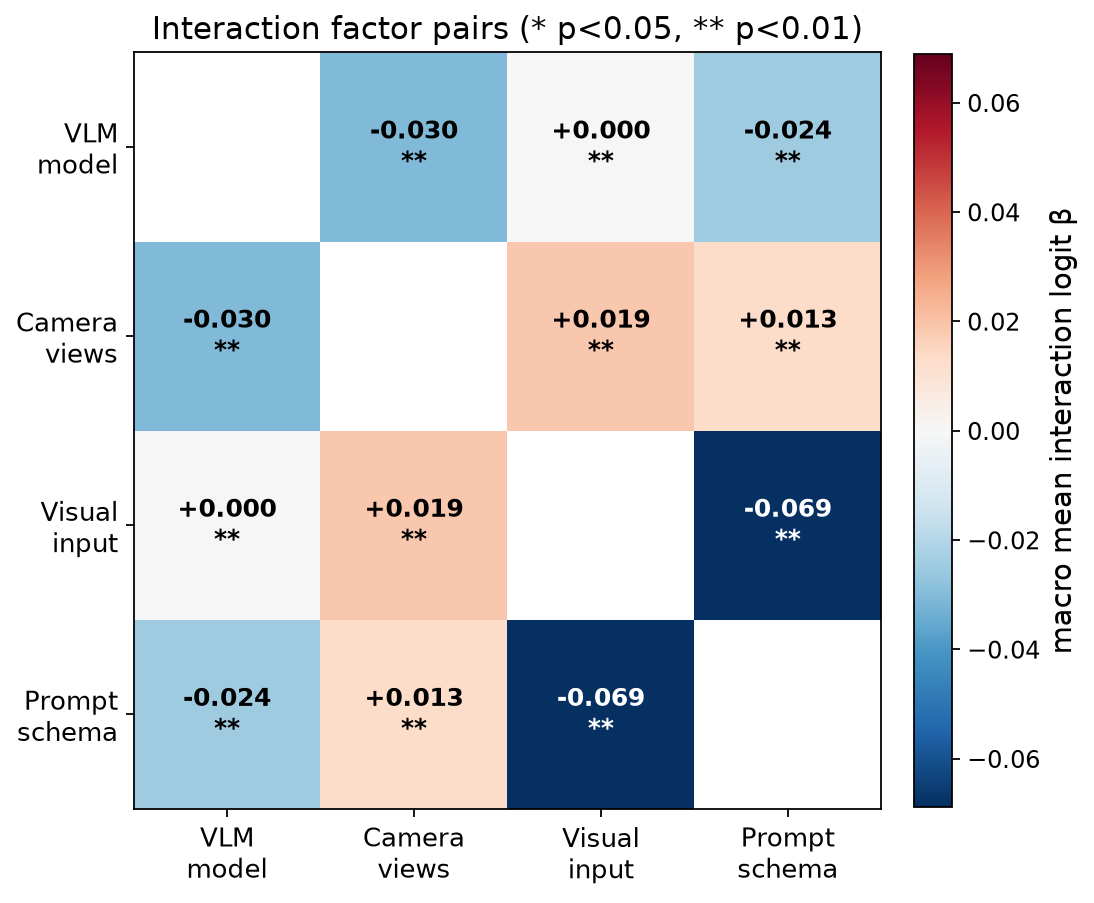}
\caption{\textbf{Interaction strength by factor pair.} Macro-mean interaction logit $\beta$ per factor pair (\texttt{*}/\texttt{**}: bootstrap $p{<}0.05$/$0.01$). Model-involving pairs dominate.}
\label{fig:stage1_inter_pairs}
\end{figure}

\begin{figure}[tbp]
\centering
\includegraphics[width=\linewidth]{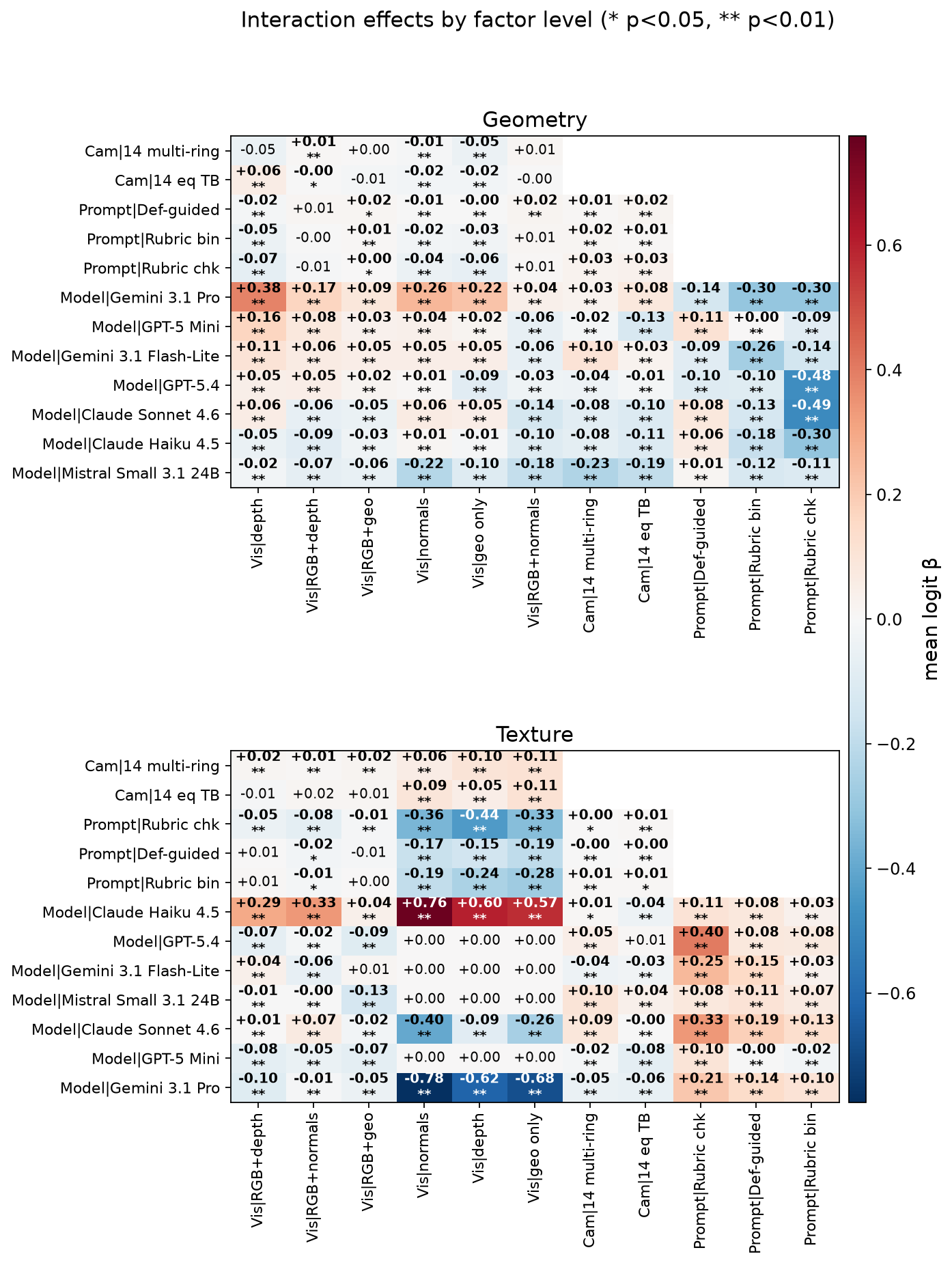}
\caption{\textbf{Level$\times$level interaction coefficients} for geometry (top) and texture (bottom): lower-triangle mean interaction logit $\beta$ (\texttt{*}/\texttt{**}: bootstrap $p{<}0.05$/$0.01$). Model-involving level pairs carry the strongest interactions.}
\label{fig:stage1_interaction}
\end{figure}

\emph{There is no universally best pipeline setting: the same visual or prompt change can help one model and hurt another.} Collectively, the six interaction families carry the largest aggregate partial pseudo-$R^2$ block (mean $1.0\times10^{-2}$) and are significant for 9/9 defects---because this block pools many factor pairs and parameters, no single interaction is as influential as model choice---so the effect of one factor often depends on another. Figure~\ref{fig:stage1_inter_pairs} summarizes interaction strength at the factor-pair level: the model$\times$visual-input pair is by far the strongest, with the visual$\times$prompt and model$\times$prompt pairs next, confirming that the same visual or prompt change can help one model and hurt another. Figure~\ref{fig:stage1_interaction} resolves these to the level$\times$level detail, separately for geometry and texture. 

\subsection{More views and extra channels add no value: a cheap six-view RGB design is near-optimal}

A practically important pattern emerges once the factors are read together: \emph{the cheapest design is also among the best}. Neither denser camera protocol earns its cost---the 14-view equatorial and 14-view multi-ring turntables do not improve on the compact six-view oblique turntable (best-vs-worst camera gap ${\sim}$0.02 macro MCC; Table~\ref{tab:effect_sizes})---and no visual channel beyond RGB helps on \emph{both} aspects: the top geometry designs use the six-view protocol with RGB or RGB augmented by a geometry/normals cue, and while a normals-only variant edges into the top geometry cluster it collapses on texture (Tex MCC 0.012, Table~\ref{tab:stage1_top}), so RGB is the only channel that is strong on both aspects and stripping color hurts (as the coefficient analysis above shows---an effect that may partly reflect alignment with the RGB-viewing human reference). Color (RGB) and a rubric-guided prompt are the only choices that clearly matter on both aspects, but interactions mean no single design is provably best for every model, so we select a \emph{well-performing default} empirically. Table~\ref{tab:stage1_top} lists the top designs by pooled macro MCC: \texttt{c004} (six oblique RGB views, rubric-guided checklist) ranks first for both geometry (0.150) and texture (0.082). The point-estimate gaps within this top cluster are small ($\sim$0.002 macro MCC on geometry) and confidence intervals overlap, so \texttt{c004} is \emph{not} uniquely separable from its neighbors. Asset-cluster bootstrap resampling ranks it first for geometry in 45\% of draws and for texture in 82\%, and top-three in 82\%/99\% (detailed in Appendix~\ref{subsec:selection}). We therefore carry \texttt{c004} forward as the concrete instance of a general recommendation---\emph{use the cheapest member of the near-optimal cluster: a compact six-view RGB turntable with a rubric-guided prompt}---rather than as a globally optimal design. This is not in tension with the judge interactions of the previous subsection: those interactions relocate each judge's optimum slightly \emph{within} this tied cluster, not outside it, so fixing one cheap member (\texttt{c004}) costs any single judge only a negligible, in-cluster gap while buying a controlled, reproducible design.

\begin{table}[tbp]
\caption{Top inference designs by pooled macro MCC vs.\ silver majority. \texttt{c004} (row A) is carried forward; the released configuration manifest provides exact reproducibility IDs.}
\label{tab:stage1_top}
\centering
\footnotesize
\begin{tabular}{llllrr}
\toprule
ID & Camera views & Visual input & Prompt schema & Geo MCC & Tex MCC \\
\midrule
\textbf{A (\texttt{c004})} & 6 oblique & RGB & Rubric checklist & \textbf{0.150} & \textbf{0.082} \\
B (\texttt{c001}) & 6 oblique & RGB & Compact binary & 0.147 & 0.051 \\
C (\texttt{c020}) & 6 oblique & RGB + geometry cue & Rubric checklist & 0.144 & 0.070 \\
D (\texttt{c010}) & 6 oblique & Normals & Definition-guided binary & 0.142 & 0.012 \\
E (\texttt{c023}) & 6 oblique & RGB + normals & Rubric-guided binary & 0.140 & 0.055 \\
\bottomrule
\end{tabular}
\end{table}

\subsection{The design conclusions transfer to a different generator}
\label{subsec:trellis_transfer}

All Stage~1 screening above uses assets from a single text-to-3D generator (Roblox Cube). To test whether the factorial conclusions are generator-specific, we replay the \emph{identical} 84-configuration grid and four proxy judges on 358 TRELLIS-generated assets that overlap the paper's prompt set and carry five-rater relabeled silver labels; as everywhere in this variant we restrict to the all-five-valid subset (assets no rater flagged as a critical failure). We render every asset following the same six-oblique RGB (and auxiliary-channel) composite protocol used in Stage~1, so the only deliberate change is the underlying mesh generator.

\begin{figure}[tbp]
\centering
\includegraphics[width=\linewidth]{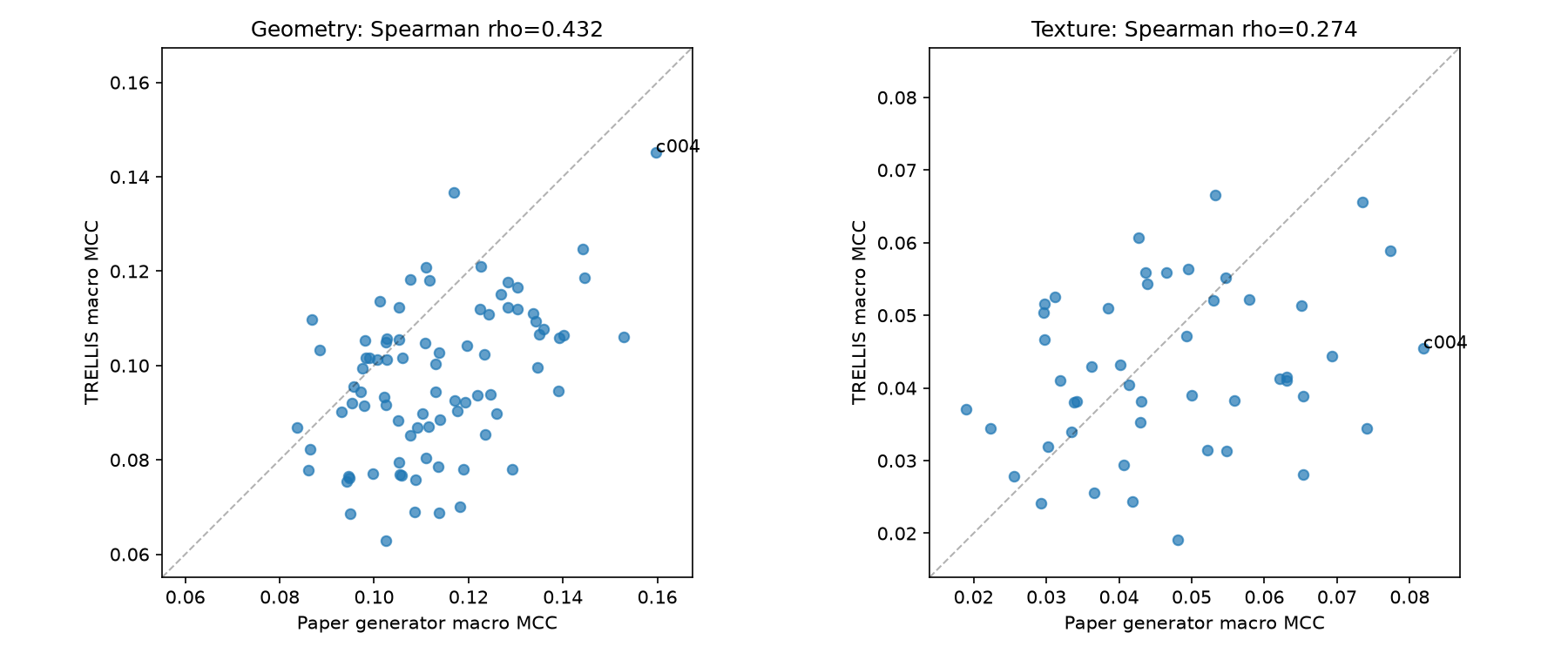}
\caption{\textbf{Cross-generator config-rank transfer.} Per-configuration macro MCC on the original Roblox Stage~1 sweep (1{,}100 assets, four proxy judges) vs.\ the TRELLIS replication (358 all-five-valid assets, same grid and judges). Points are the 84 design cells; \texttt{c004} is annotated. Geometry ranks transfer ($\rho{=}0.43$, $p{<}0.001$) while texture does not ($\rho{=}0.27$, n.s.); per-aspect values in Table~\ref{tab:trellis_transfer}.}
\label{fig:trellis_config_scatter}
\end{figure}

\begin{figure}[tbp]
\centering
\includegraphics[width=\linewidth]{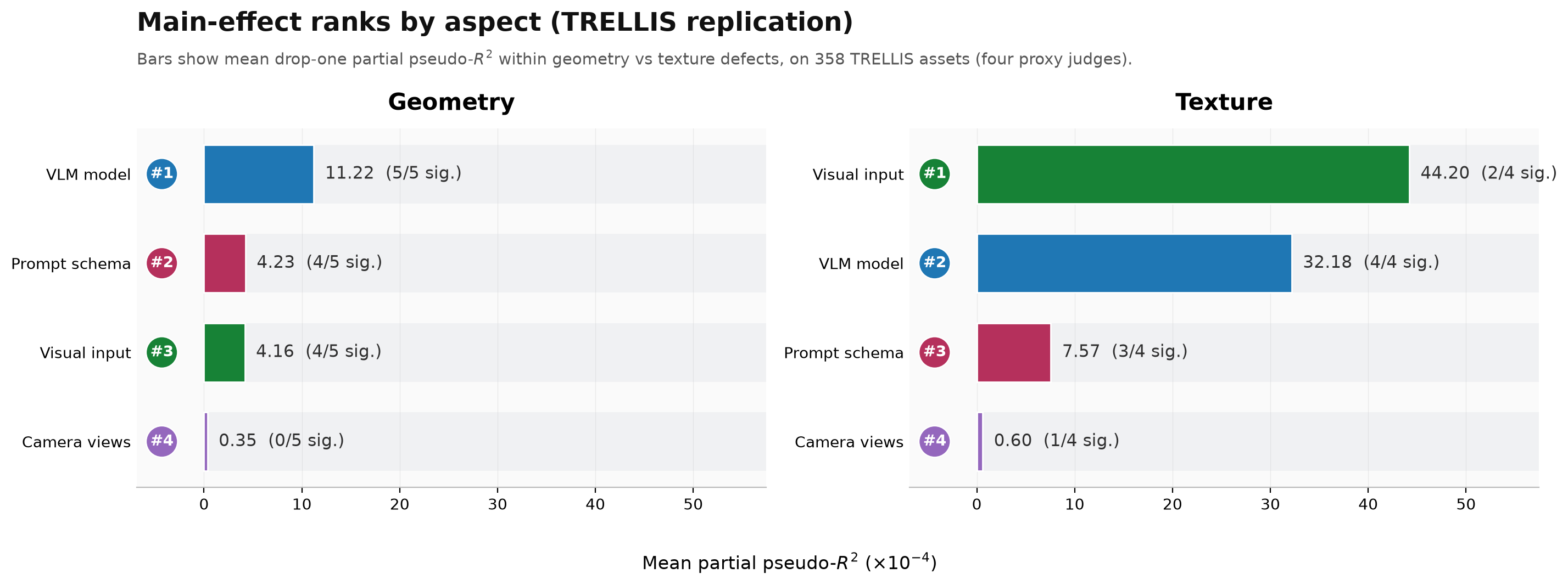}
\caption{\textbf{Factor-family importance on TRELLIS, by aspect} (TRELLIS analogue of Figure~\ref{fig:stage1_main_effect_rank}). Mean drop-one partial pseudo-$R^2$ per factor family within geometry vs.\ texture defects, on 358 all-five-valid TRELLIS assets (four proxy judges); ``$n/m$ sig.'' counts defects in the aspect with omnibus $p{<}0.05$. As on Roblox Cube, VLM model dominates geometry and camera-view protocol is least influential in both aspects; on texture, visual input overtakes VLM model as the leading factor.}
\label{fig:trellis_rank_aspect}
\end{figure}

\begin{figure}[tbp]
\centering
\includegraphics[width=\linewidth]{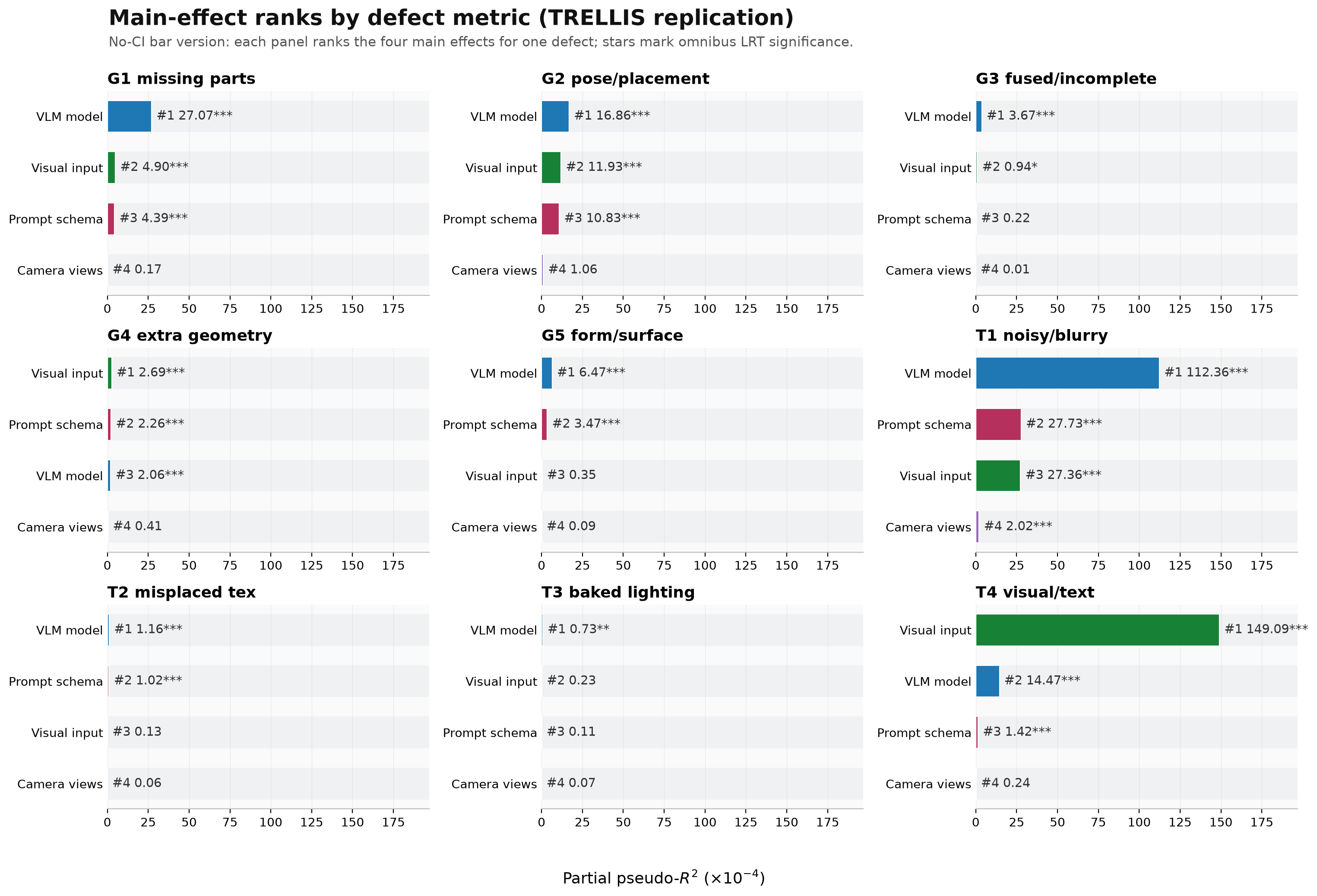}
\caption{\textbf{Factor-family importance on TRELLIS, by defect} (TRELLIS analogue of Figure~\ref{fig:stage1_factor_importance}). Each panel ranks the four main effects for one defect by drop-one partial pseudo-$R^2$ ($\times10^{-4}$); stars mark omnibus LRT significance ($p{<}0.05/0.01/0.001$). VLM model tops the high-signal geometry defects (missing parts, pose/placement); the largest texture signal is a visual-input effect on the texture-content-mismatch defect, while on the low-signal defects all main effects are near zero and rank order is not meaningful.}
\label{fig:trellis_rank_defect}
\end{figure}

Table~\ref{tab:trellis_transfer} and Figure~\ref{fig:trellis_config_scatter} summarize cross-generator transfer, and the picture differs sharply by aspect. For \emph{geometry}, the design conclusions transfer cleanly: per-configuration macro MCC ranks correlate between the original 1{,}100-asset Roblox sweep and the TRELLIS replication (Spearman $\rho{=}0.43$ over the 84 designs, $p{<}0.001$), the carried-forward champion \texttt{c004} remains the top-ranked geometry design on TRELLIS, and refitting the same drop-one factor model reproduces the two anchors that justify a fixed design---the VLM model dominates by a wide margin (mean partial pseudo-$R^2$ $1.1{\times}10^{-3}$, roughly $2.7\times$ the next factor) and the camera-view protocol is the least influential factor---with the visual-input/prompt-schema middle band swapping order (aspect-level family-rank $\rho{=}0.80$). \emph{Texture} transfers only weakly: the per-configuration rank correlation is low and not significant ($\rho{=}0.27$, $p{=}0.06$), \texttt{c004} falls to rank 20 of 84, and the factor structure reorders---visual input overtakes VLM model as the most influential texture factor (mean partial pseudo-$R^2$ $4.4{\times}10^{-3}$ vs.\ $3.2{\times}10^{-3}$), driven almost entirely by the texture-content-mismatch defect where the RGB channel is decisive (Figures~\ref{fig:trellis_rank_aspect}--\ref{fig:trellis_rank_defect}). This aspect split is consistent with the Cube screen, where texture is the lower-signal aspect throughout. Absolute MCC differs between generators (TRELLIS defect prevalence and mesh style differ from Cube), but the \emph{relative} geometry pipeline rankings---what we need to justify a fixed evaluation design---transfer. We therefore treat the geometry design findings, especially that model choice dominates and camera protocol matters least, as properties of the VLM judging pipeline rather than idiosyncrasies of one generator's output distribution, while flagging the texture rankings as generator-sensitive and lower-confidence.

\begin{table}[tbp]
\centering
\caption{TRELLIS cross-generator replication of Stage~1 config rankings (358 all-5-valid assets, same 84-config grid and four proxy judges).}
\label{tab:trellis_transfer}
\begin{tabular}{lrrrrr}
\toprule
Aspect & $\rho$ (config rank) & Top config (paper) & Top config (TRELLIS) & Top-5 overlap & \texttt{c004} rank (TRELLIS) \\
\midrule
Geometry & 0.432 & c004 & c004 & 2/5 & 1 \\
Texture & 0.274 & c004 & c030 & 2/5 & 20 \\
\bottomrule
\end{tabular}
\end{table}

\section{Extended VLM judge comparison and calibration under a fixed pipeline}
\label{sec:stage3golden}

We compare all 12 VLMs under a single fixed configuration. The configuration is selected by two complementary routes, and---reassuringly---both select \texttt{c004}: (1) a \emph{silver-only} route that selects the configuration on a 500-asset silver subset and then tests the extended model set on the disjoint 506-asset silver holdout; and (2) an \emph{expert-transfer} route that selects the configuration on the full silver set and then tests the extended model set against the higher-quality expert labels. The two routes use different label pools for selection but converge on the same design, which is why we report a single leaderboard against both references.

\subsection{The Gemini family leads; texture drops sharply from expert to silver}

\emph{The Gemini family tops nearly every column, and absolute agreement---texture especially---falls when the reference shifts from clean expert labels to the noisier silver holdout.} Table~\ref{tab:leaderboard_combined} reports macro MCC and macro F1 for both aspects on the expert-agreement cells and the 506-asset silver holdout. \textbf{Gemini~3.1~Pro} leads expert geometry (MCC 0.298$\pm$0.047 SE) and expert texture is led by \textbf{Gemini~3.1~Flash-Lite} (0.406), with Gemini~3.1~Pro close behind (0.386). The Gemini family occupies the top of every expert column and all but one silver column (GPT-5.4 edges ahead on silver-holdout geometry MCC); smaller and open models trail, and the two weakest judges---Qwen2.5-VL-7B and, despite its frontier billing, Claude~Haiku~4.5---sit near zero on the expert cells. Absolute scores drop from expert to silver---sharply for texture (best 0.406$\to$0.168)---while geometry holds a similar range (0.298$\to$0.226), consistent with noisier silver texture labels. Per-defect MCC (with bootstrap CIs) and F1 by model are shown in Appendix~\ref{subsec:perdefect} (Figures~\ref{fig:golden_heatmap}--\ref{fig:golden_defect_ci}). The leaderboard is an accuracy comparison under one fixed inference design; it does not isolate whether errors arise from visual perception, prompt following, or taxonomy understanding. The rank-1 model is stable whether we score the both-experts-agree cells or the either-expert union, though absolute MCC and the lower ranks shift (Appendix~\ref{subsec:expert_modes}).

\begin{table}[t]
\caption{Model-comparison leaderboard under the selected configuration (\texttt{c004}): macro MCC and macro F1 on the 129-asset expert split (agreement cells) and the 506-asset silver holdout, by aspect. Rows ordered by expert geometry MCC. Per-defect metrics are macro-averaged within each aspect.}
\label{tab:leaderboard_combined}
\centering
\scriptsize
\setlength{\tabcolsep}{4pt}
\begin{tabular}{l rrrr rrrr}
\toprule
 & \multicolumn{4}{c}{Expert (agreement cells)} & \multicolumn{4}{c}{Silver holdout ($n{=}506$)} \\
\cmidrule(lr){2-5}\cmidrule(lr){6-9}
 & \multicolumn{2}{c}{Geometry} & \multicolumn{2}{c}{Texture} & \multicolumn{2}{c}{Geometry} & \multicolumn{2}{c}{Texture} \\
\cmidrule(lr){2-3}\cmidrule(lr){4-5}\cmidrule(lr){6-7}\cmidrule(lr){8-9}
Model & MCC & F1 & MCC & F1 & MCC & F1 & MCC & F1 \\
\midrule
Gemini 3.1 Pro & \textbf{0.298} & 0.402 & 0.386 & \textbf{0.711} & 0.206 & 0.311 & \textbf{0.168} & \textbf{0.674} \\
Gemini 2.5 Pro & 0.282 & \textbf{0.415} & 0.280 & 0.614 & 0.185 & 0.307 & 0.108 & 0.603 \\
Gemini 3.1 Flash-Lite & 0.253 & 0.370 & \textbf{0.406} & 0.647 & 0.204 & 0.294 & 0.092 & 0.514 \\
GPT-5.4 & 0.217 & 0.356 & 0.218 & 0.494 & \textbf{0.226} & 0.302 & 0.127 & 0.430 \\
Claude Opus 4.7 & 0.180 & 0.341 & 0.301 & 0.552 & 0.193 & \textbf{0.314} & 0.138 & 0.484 \\
GPT-4o & 0.163 & 0.221 & 0.255 & 0.467 & 0.117 & 0.148 & 0.100 & 0.399 \\
Qwen 3.5 397B-A17B & 0.162 & 0.324 & 0.140 & 0.457 & 0.125 & 0.249 & 0.091 & 0.435 \\
Claude Sonnet 4.6 & 0.145 & 0.296 & 0.226 & 0.454 & 0.125 & 0.236 & 0.007 & 0.429 \\
GPT-5 Mini & 0.142 & 0.237 & 0.122 & 0.358 & 0.170 & 0.231 & 0.021 & 0.315 \\
Mistral Small 3.1 24B & 0.051 & 0.237 & 0.066 & 0.237 & 0.074 & 0.195 & 0.087 & 0.259 \\
Qwen2.5-VL-7B & 0.017 & 0.015 & 0.078 & 0.102 & 0.027 & 0.019 & 0.010 & 0.040 \\
Claude Haiku 4.5 & 0.014 & 0.079 & 0.029 & 0.157 & 0.082 & 0.124 & 0.016 & 0.225 \\
\bottomrule
\end{tabular}
\end{table}

\begin{figure}[tbp]
\centering
\begin{subfigure}{0.49\linewidth}\centering
\includegraphics[width=\linewidth]{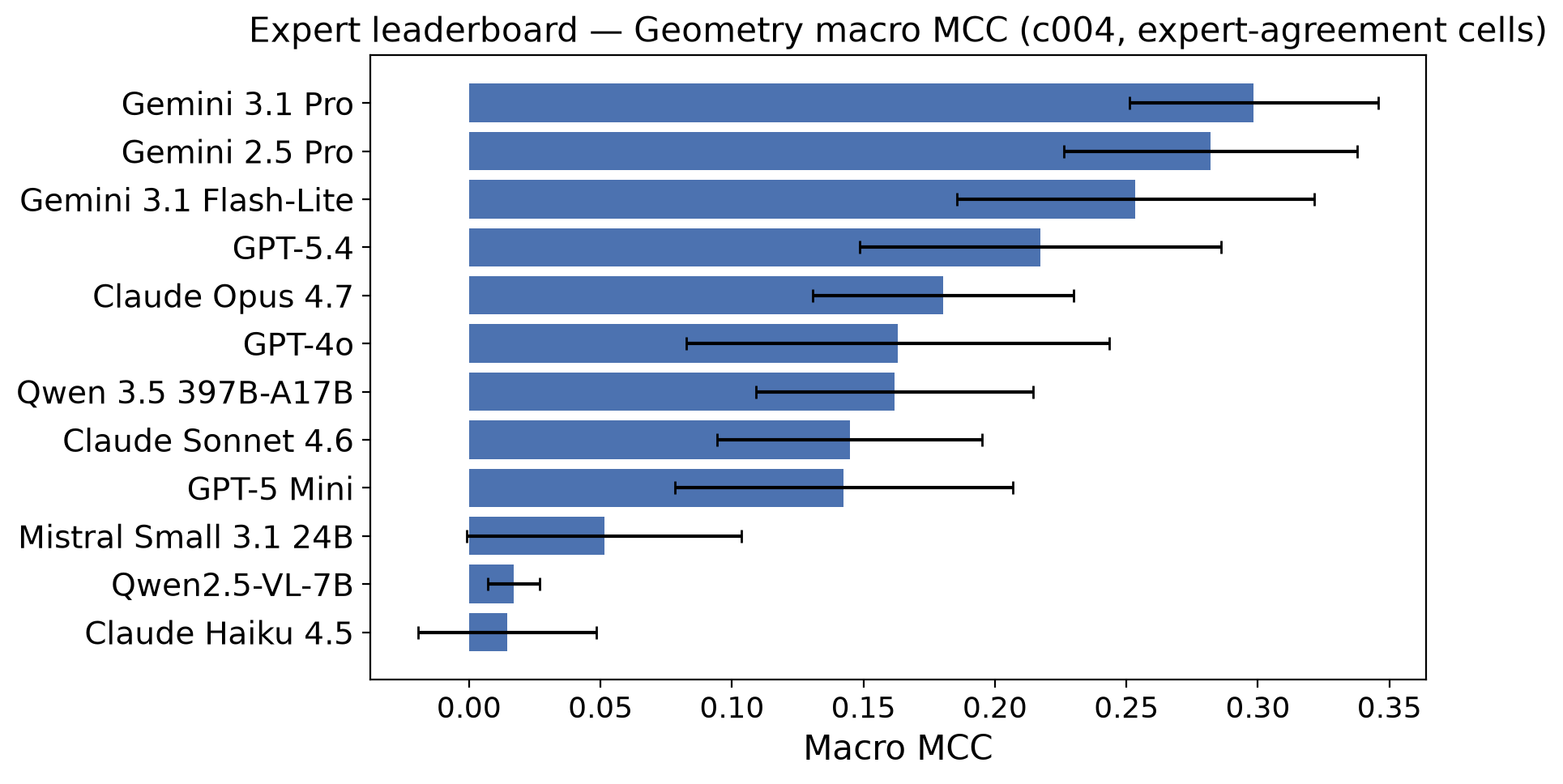}
\caption*{\scriptsize Geometry}
\end{subfigure}\hfill
\begin{subfigure}{0.49\linewidth}\centering
\includegraphics[width=\linewidth]{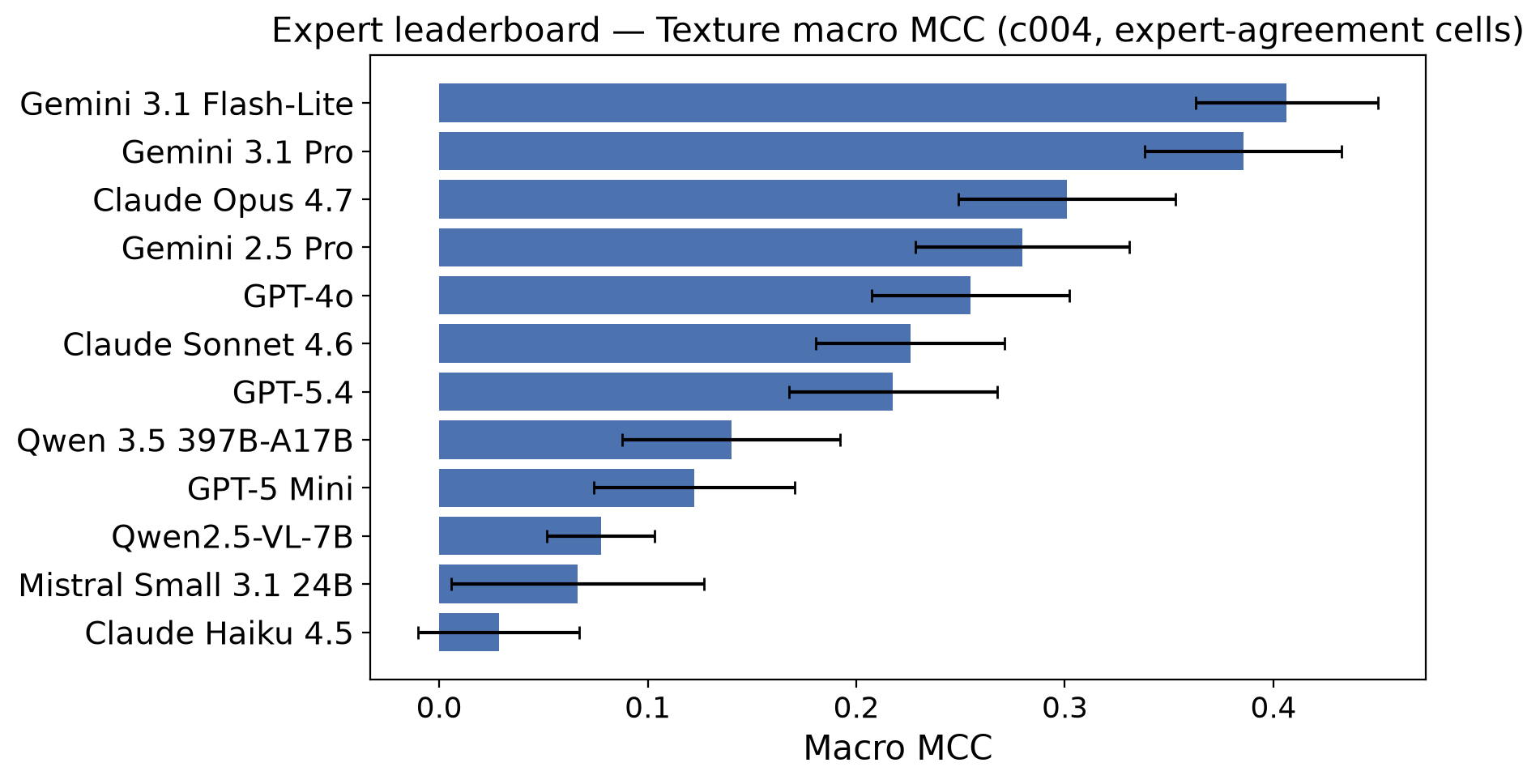}
\caption*{\scriptsize Texture}
\end{subfigure}
\caption{Expert-label macro MCC leaderboards ($\pm$1 SE).}
\label{fig:stage3_golden}
\end{figure}

We additionally include \textbf{GPT-6 Astra}, the newest frontier judge (released after the study was frozen), run at the fixed \texttt{c004} design on the full 1{,}006-asset silver pool to check that the carried-forward design and model ordering survive the latest generation (Table~\ref{tab:c004_frontier}, silver spot-check; not run on the expert-agreement cells). At a high level, the newest model posts the strongest point estimate in every column yet stays within the established band: it separably leads on geometry (its interval clears Gemini~3.1~Pro), while on aspect-averaged and texture MCC its interval overlaps Gemini~3.1~Pro (a statistical tie), and the wider ordering---and the trained-human ceiling of Section~\ref{sec:human}---is unchanged. A stronger judge therefore improves absolute agreement without dislodging either the fixed \texttt{c004} design or the model-choice-dominates conclusion, reinforcing that the pipeline findings are stable across model vintage.

\begin{table}[tbp]
\centering
\small
\caption{\textbf{Latest-frontier judge spot-check at the fixed \texttt{c004} design.} Macro MCC (\texttt{parse\_ok} only) on the 1{,}006-asset relabeled five-vote silver pool for nine judges run at \texttt{c004} (6-oblique RGB + Rubric-Guided Checklist), including the latest \textbf{GPT-6 Astra}. Brackets are asset-cluster bootstrap 95\% CIs (1{,}000 resamples of \texttt{object\_id}); best per column in bold. This roster is the set of models run at \texttt{c004} on silver and differs from the expert-set leaderboard of Table~\ref{tab:leaderboard_combined} (InternVL excluded).}
\label{tab:c004_frontier}
\begin{tabular}{lccc}
\toprule
Judge & Geometry MCC & Texture MCC & Avg MCC \\
\midrule
GPT-6 Astra & \textbf{0.301} \scriptsize[0.253, 0.347] & \textbf{0.169} \scriptsize[0.136, 0.202] & \textbf{0.235} \scriptsize[0.205, 0.265] \\
Gemini 3.1 Pro & 0.215 \scriptsize[0.185, 0.244] & 0.168 \scriptsize[0.138, 0.201] & 0.192 \scriptsize[0.170, 0.214] \\
Gemini 3.1 Flash-Lite & 0.217 \scriptsize[0.175, 0.256] & 0.117 \scriptsize[0.084, 0.149] & 0.167 \scriptsize[0.139, 0.193] \\
GPT-5.4 & 0.220 \scriptsize[0.181, 0.256] & 0.113 \scriptsize[0.081, 0.144] & 0.167 \scriptsize[0.142, 0.192] \\
Qwen 3.5 397B-A17B & 0.144 \scriptsize[0.108, 0.182] & 0.100 \scriptsize[0.070, 0.134] & 0.122 \scriptsize[0.097, 0.148] \\
GPT-5 Mini & 0.179 \scriptsize[0.138, 0.219] & 0.045 \scriptsize[0.013, 0.075] & 0.112 \scriptsize[0.087, 0.138] \\
Claude Sonnet 4.6 & 0.144 \scriptsize[0.091, 0.196] & 0.047 \scriptsize[0.019, 0.075] & 0.096 \scriptsize[0.067, 0.123] \\
Mistral Small 3.1 24B & 0.065 \scriptsize[0.029, 0.101] & 0.068 \scriptsize[0.038, 0.099] & 0.066 \scriptsize[0.043, 0.088] \\
Claude Haiku 4.5 & 0.067 \scriptsize[0.022, 0.114] & 0.001 \scriptsize[-0.032, 0.036] & 0.034 \scriptsize[0.004, 0.064] \\
\bottomrule
\end{tabular}
\end{table}

\subsection{Over- and under-calling by defect}
\begin{figure}[tbp]
\centering
\includegraphics[width=\linewidth]{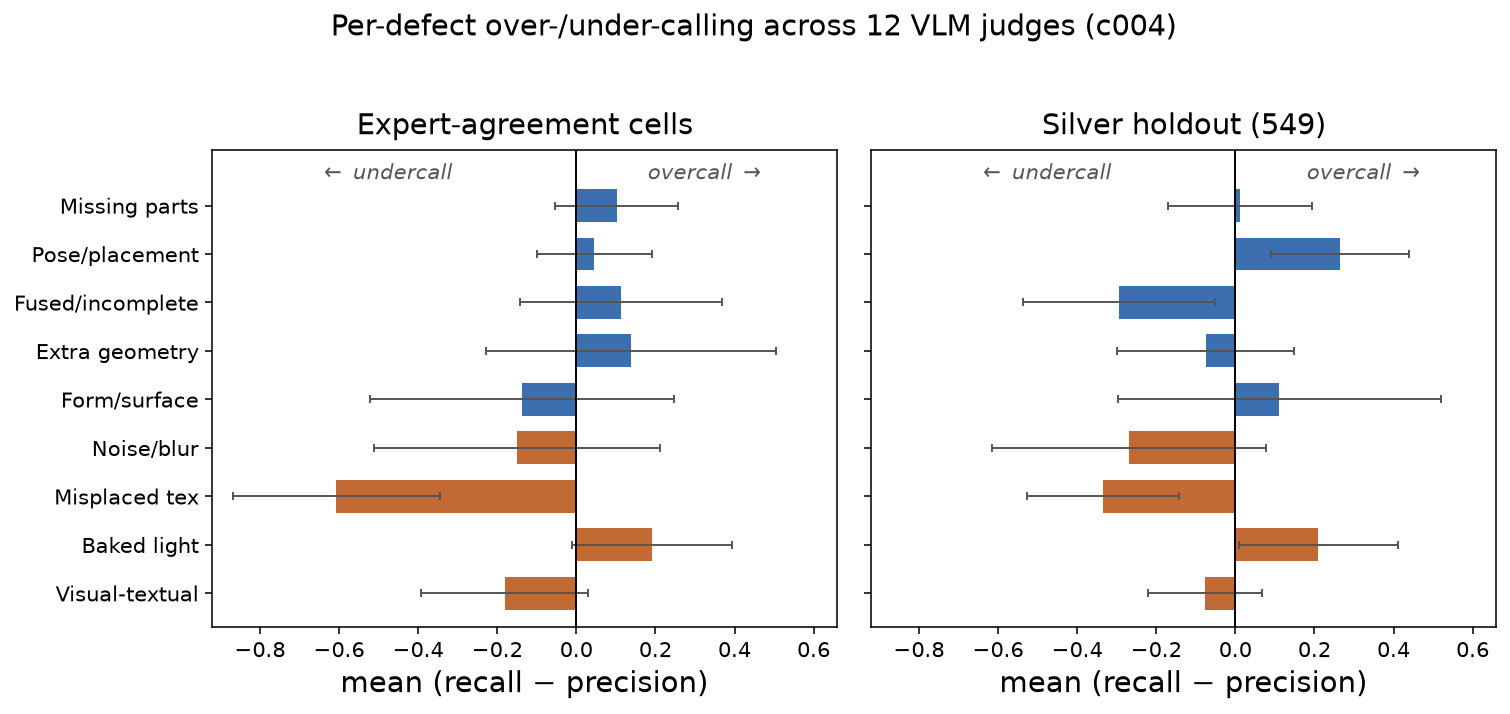}
\caption{Per-defect over-/under-calling for the 12 VLM judges under \texttt{c004}: mean recall$-$precision across models (whiskers $\pm$1 SD). Right of zero = overcall (false-positive-heavy); left = undercall (false-negative-heavy). Expert-agreement cells (left) and 506-asset silver holdout (right).}
\label{fig:calling}
\end{figure}

\begin{figure}[tbp]
\centering
\includegraphics[width=0.86\linewidth]{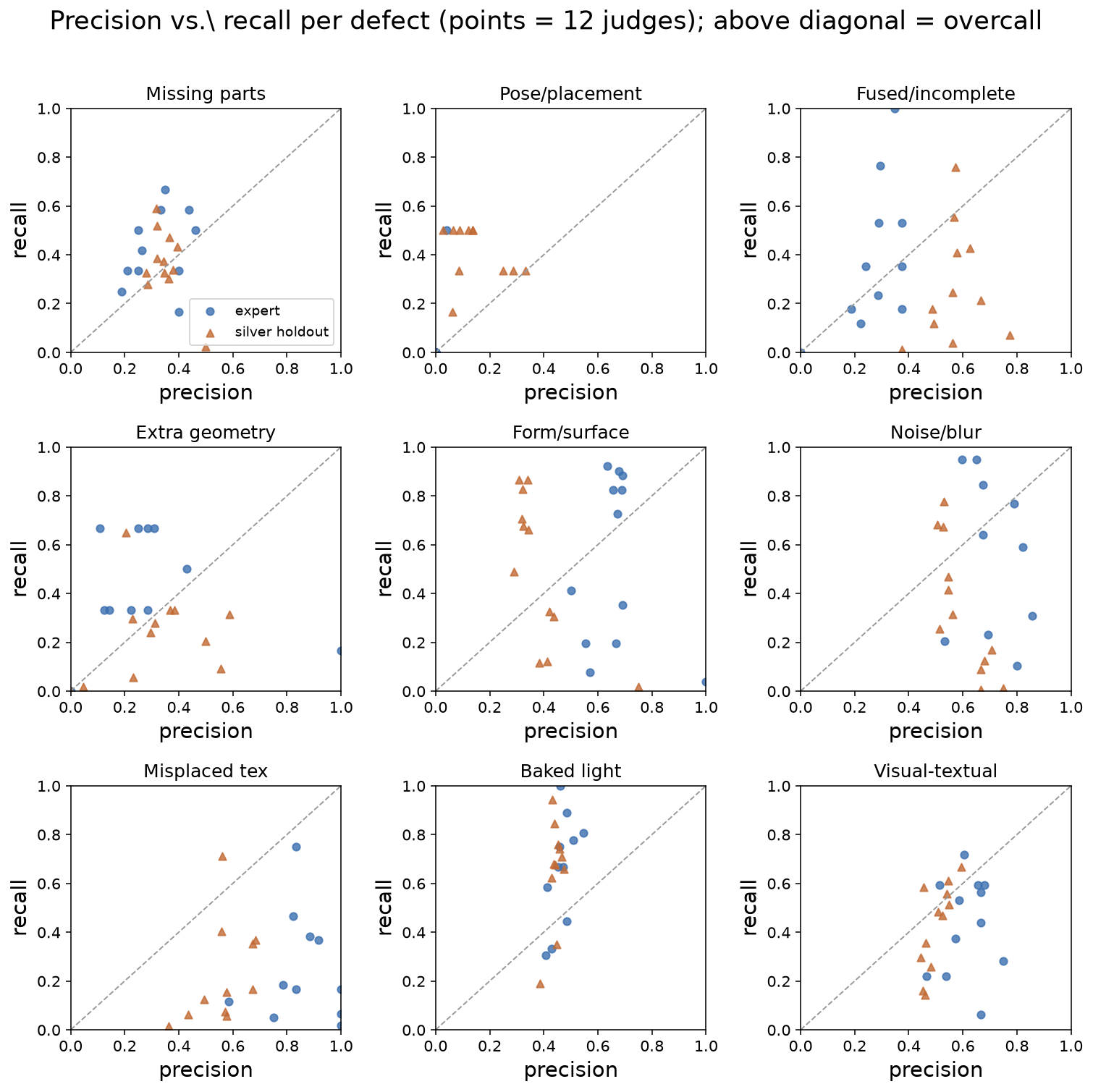}
\caption{Precision vs.\ recall per defect; each point is one of the 12 judges. Points above the diagonal overcall the defect (recall${>}$precision), below undercall it. Circles = expert-agreement cells, triangles = silver holdout.}
\label{fig:calling_scatter}
\end{figure}

Beyond aggregate accuracy, we ask whether judges systematically over- or under-call each defect by comparing per-defect recall against precision, averaged across the 12 models (Figure~\ref{fig:calling}): a positive recall$-$precision gap means false positives dominate (the model \emph{overcalls} the defect), a negative gap means false negatives dominate (it \emph{undercalls}). VLM judges tend to \emph{overcall missing parts} on both references and baked lighting/shadow on the expert cells---recall exceeds precision and most models are false-positive-heavy---while pervasive texture defects such as misplaced or overlapping texture are strongly \emph{undercalled} on both references (high precision, low recall); the per-model precision--recall structure is shown in Figure~\ref{fig:calling_scatter}. Because the released predictions are binary (no model-emitted probabilities), we report this false-positive/false-negative asymmetry rather than probabilistic calibration curves or Brier scores. The model-averaged per-defect breakdown is in Table~\ref{tab:per_defect_calling}.

\begin{table}[t]
\caption{Per-defect over-/under-calling by the 12 VLM judges under \texttt{c004}, averaged across models. \emph{Prec} and \emph{Rec} are mean precision and recall; \emph{Rec$-$Prec} is the mean signed gap (positive $\Rightarrow$ false positives dominate, i.e.\ the judges \emph{overcall} the defect; negative $\Rightarrow$ \emph{undercall}); \emph{FP share} is $\mathrm{FP}/(\mathrm{FP}{+}\mathrm{FN})$ ($>0.5$ = overcall). Because released predictions are binary, this false-positive/false-negative asymmetry stands in for probabilistic calibration.}
\label{tab:per_defect_calling}
\centering
\footnotesize
\begin{tabular}{llrrrrr}
\toprule
Aspect & Defect & Split & Prec & Rec & Rec$-$Prec & FP share \\
\midrule
Geometry & Extra geometry & expert & 0.29 & 0.42 & +0.14 & 0.65 \\
 &  & silver & 0.35 & 0.26 & -0.09 & 0.42 \\
 & Form or surface quality & expert & 0.67 & 0.53 & -0.14 & 0.45 \\
 &  & silver & 0.31 & 0.50 & +0.19 & 0.74 \\
 & Fused or incomplete parts & expert & 0.27 & 0.39 & +0.11 & 0.54 \\
 &  & silver & 0.39 & 0.27 & -0.12 & 0.33 \\
 & Missing parts & expert & 0.32 & 0.42 & +0.10 & 0.60 \\
 &  & silver & 0.14 & 0.35 & +0.21 & 0.78 \\
 & Pose or placement mismatch & expert & 0.00 & 0.05 & +0.05 & 0.74 \\
 &  & silver & 0.10 & 0.24 & +0.14 & 0.77 \\
\midrule
Texture & Baked lighting or shadow & expert & 0.47 & 0.66 & +0.19 & 0.69 \\
 &  & silver & 0.69 & 0.60 & -0.09 & 0.34 \\
 & Visual-textual mismatch & expert & 0.61 & 0.43 & -0.18 & 0.34 \\
 &  & silver & 0.29 & 0.43 & +0.15 & 0.64 \\
 & Misplaced or overlapping texture & expert & 0.86 & 0.25 & -0.61 & 0.08 \\
 &  & silver & 0.56 & 0.19 & -0.37 & 0.13 \\
 & Noise, blur, or grain & expert & 0.71 & 0.56 & -0.15 & 0.41 \\
 &  & silver & 0.89 & 0.31 & -0.58 & 0.08 \\
\bottomrule
\end{tabular}
\end{table}

\subsection{Severe defects are easier to catch: recall rises against the \emph{Major} target}
\label{sec:severity}
\emph{Human majorities mark a subset of each defect as \emph{Major} (Severe); scored against that severe target, the same binary VLM calls recover far more of the positives---recall rises---even though they cannot tell Major from minor, so precision collapses.} Beyond the binary present/absent target used everywhere else, the labeling interface records a per-defect severity toggle (Figure~\ref{fig:ui}b). We derive a per-defect \emph{severe-majority} target with the same five-vote majority rule as the binary labels---a defect is severe for an asset when at least three of its five labelers marked it \emph{Major}---over the eight severity-eligible defects (Noisy/Blurry/Graininess Texture has no severity toggle and is excluded). We then score each judge's \emph{unchanged} binary defect call against (a) the binary presence target and (b) the severe target, holding the 506-asset silver holdout, the 12 judges, and the eight defects fixed, so only the target changes (Table~\ref{tab:severity_vlm}). The hypothesis is that severe defects are more conspicuous and should therefore be caught more often (higher recall). This is borne out: averaged over the 12 judges, recall rises from 0.355 (binary) to 0.428 (severe) overall, and on both aspects (geometry 0.324$\to$0.420, texture 0.408$\to$0.441)---a judge's binary flag fires on roughly nine-tenths\footnote{Recall is averaged over judges and defects; the strongest judges exceed this, the weakest pull the mean down.} of the blatant, human-flagged \emph{Major} defects. The flip side is that precision falls sharply (0.354$\to$0.102 overall) and MCC/F1 drop with it: because the released predictions are a single binary present/absent call with no severity dimension, every correctly flagged \emph{minor} defect becomes a false positive against the severe target. The takeaway is diagnostic rather than a leaderboard claim---VLM judges reliably surface the defects humans consider serious, but a binary present/absent protocol cannot by itself grade severity, motivating an explicit severity rubric for future judge designs.

\begin{table}[t]
\caption{VLM binary defect calls scored against two majority-vote human targets on the 506-asset silver holdout, averaged over the 12 judges and macro-averaged over the eight severity-eligible defects (the Noisy/Blurry/Graininess texture defect has no severity option and is excluded from both targets here). \emph{Binary} = defect present; \emph{Severe} = marked Major by majority vote. Recall rises sharply against the severe target while precision falls: a conspicuous, human-flagged \emph{Major} defect is almost always caught, but the binary call cannot tell Major from minor.}
\label{tab:severity_vlm}
\centering
\small
\begin{tabular}{ll rrrr}
\toprule
Aspect & Target & MCC & F1 & Precision & Recall \\
\midrule
\multirow{2}{*}{Geometry} & Binary & 0.144 & 0.228 & 0.258 & 0.324 \\
 & Severe & 0.137 & 0.145 & 0.114 & 0.420 \\
\multirow{2}{*}{Texture} & Binary & 0.081 & 0.401 & 0.513 & 0.408 \\
 & Severe & 0.044 & 0.122 & 0.081 & 0.441 \\
\midrule
\multirow{2}{*}{Overall} & Binary & 0.121 & 0.293 & 0.354 & 0.355 \\
 & Severe & 0.102 & 0.136 & 0.102 & 0.428 \\
\bottomrule
\end{tabular}
\end{table}

\subsection{Baseline calibration and modality ablations: the rendered views carry most of the signal, the prompt a smaller lift}
\label{sec:baselines}

To calibrate what these MCC values mean, we benchmark the full pipeline against simple baselines scored with the same macro-MCC machinery. \textbf{Label-only} baselines use only the label distribution: \textbf{always-negative} predicts ``no defect'' everywhere; \textbf{majority-class} predicts each defect's most common label; both yield macro MCC 0.0 by construction. \textbf{Prevalence-matched random} samples predictions from the per-defect base rate, giving macro MCC $\approx 0$. \textbf{Modality ablations} on the selected config (\texttt{c004}) with Gemini~3.1~Pro: \textbf{prompt-only} (text rubric, no image) and \textbf{image-only} (multi-view RGB, no prompt line). \textbf{CLIP} is a local zero-shot heuristic (ViT-B-32) with per-defect contrastive templates and thresholds calibrated on the silver screening split.

Tables~\ref{tab:baselines_expert} and~\ref{tab:baselines_silver} report macro MCC with asset-cluster bootstrap 95\% confidence intervals. On expert-agreement geometry, image-only retains most of the full-pipeline signal (0.248 vs.\ 0.298 MCC) while text-only is much weaker (0.096) and CLIP is near zero; the same ordering holds on the silver holdout (full pipeline 0.206 geometry / 0.168 texture; image-only 0.197 / 0.202; prompt-only 0.054 / 0.021; CLIP near zero). On silver texture the image-only ablation slightly exceeds the full pipeline (0.202 vs.\ 0.168), consistent with the low, noisy texture ceiling where the added prompt line does not help. This confirms the judges rely primarily on the rendered evidence, that the prompt schema contributes a smaller but real increment, and that all trained VLMs clear the label-only and CLIP baselines.

\begin{table}[t]
\caption{Simple baselines on expert-agreement cells (539 geometry / 338 texture unique cells). Macro MCC with 95\% asset-cluster bootstrap CI.}
\label{tab:baselines_expert}
\centering
\footnotesize
\begin{tabular}{lrr}
\toprule
Baseline & Geometry & Texture \\
\midrule
Full pipeline & 0.298\ci{0.203}{0.391} & 0.386\ci{0.299}{0.477} \\
Image-only & 0.248\ci{0.139}{0.360} & 0.364\ci{0.265}{0.458} \\
Prompt-only & 0.096\ci{-0.010}{0.213} & 0.034\ci{-0.031}{0.100} \\
CLIP zero-shot & 0.029\ci{-0.050}{0.108} & 0.047\ci{-0.074}{0.133} \\
Prevalence-matched random & 0.000\ci{-0.006}{0.006} & -0.000\ci{-0.007}{0.007} \\
Always-negative & 0.000\ci{0.000}{0.000} & 0.000\ci{0.000}{0.000} \\
Majority-class & 0.000\ci{0.000}{0.000} & 0.000\ci{0.000}{0.000} \\
\bottomrule
\end{tabular}
\end{table}

\begin{table}[t]
\caption{Simple baselines on the 506-asset silver validation holdout (relabel-5 silver majority labels). Macro MCC with 95\% asset-cluster bootstrap CI.}
\label{tab:baselines_silver}
\centering
\footnotesize
\begin{tabular}{lrr}
\toprule
Baseline & Geometry & Texture \\
\midrule
Full pipeline & 0.206\ci{0.162}{0.246} & 0.168\ci{0.128}{0.210} \\
Image-only & 0.197\ci{0.156}{0.240} & 0.202\ci{0.163}{0.247} \\
Prompt-only & 0.054\ci{-0.012}{0.129} & 0.021\ci{-0.030}{0.065} \\
CLIP zero-shot & -0.011\ci{-0.054}{0.033} & 0.011\ci{-0.023}{0.047} \\
Prevalence-matched random & 0.000\ci{-0.003}{0.003} & 0.000\ci{-0.003}{0.003} \\
Always-negative & 0.000\ci{0.000}{0.000} & 0.000\ci{0.000}{0.000} \\
Majority-class & 0.000\ci{0.000}{0.000} & 0.000\ci{0.000}{0.000} \\
\bottomrule
\end{tabular}
\end{table}

\begin{figure}[tbp]
\centering
\begin{subfigure}{0.49\linewidth}\centering
\includegraphics[width=\linewidth]{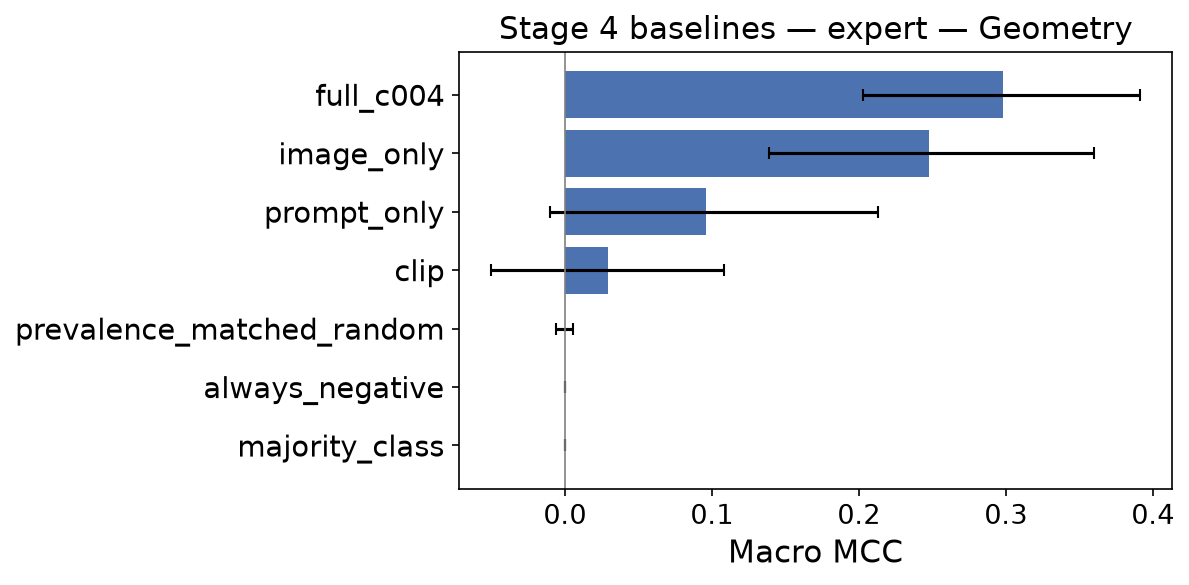}
\caption*{\scriptsize Geometry}
\end{subfigure}\hfill
\begin{subfigure}{0.49\linewidth}\centering
\includegraphics[width=\linewidth]{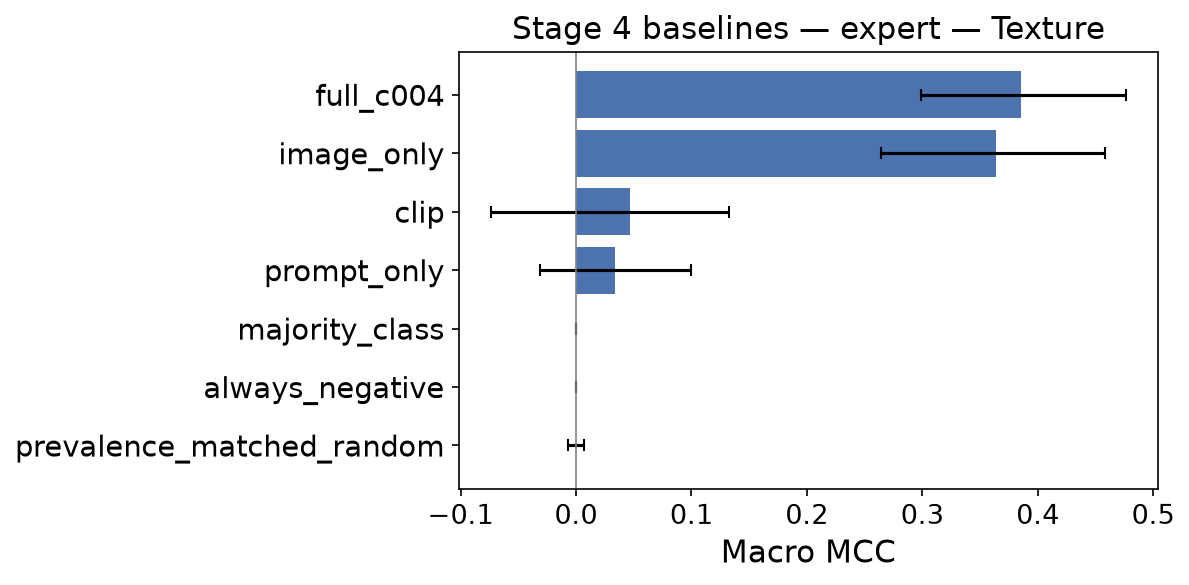}
\caption*{\scriptsize Texture}
\end{subfigure}
\caption{Simple baseline macro MCC on expert-agreement cells ($\pm$95\% bootstrap CI).}
\label{fig:baselines_expert}
\end{figure}

\begin{figure}[tbp]
\centering
\begin{subfigure}{0.49\linewidth}\centering
\includegraphics[width=\linewidth]{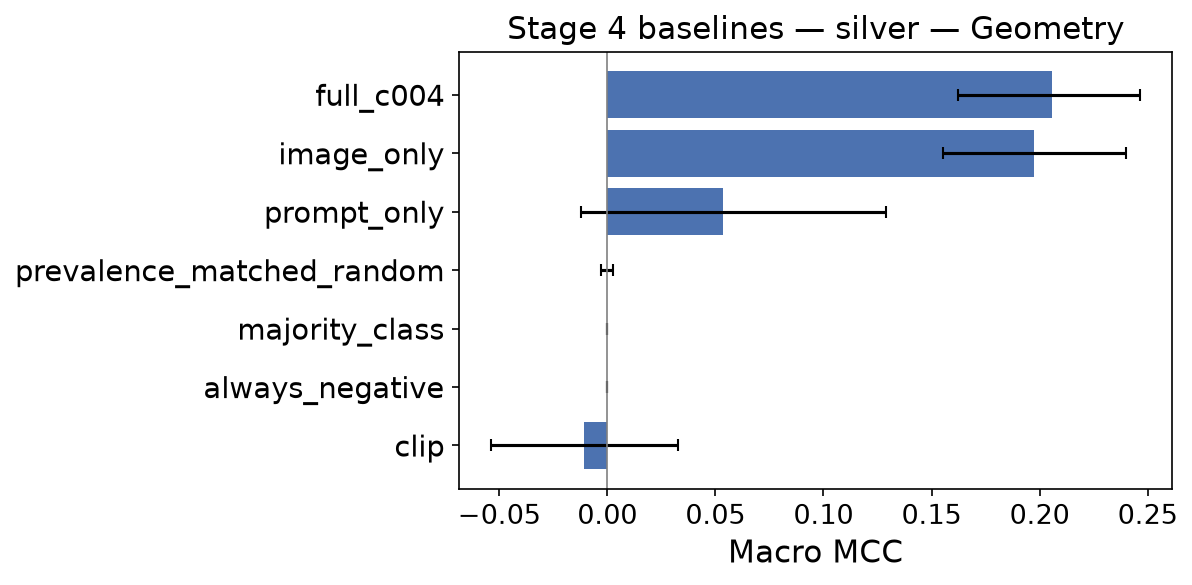}
\caption*{\scriptsize Geometry}
\end{subfigure}\hfill
\begin{subfigure}{0.49\linewidth}\centering
\includegraphics[width=\linewidth]{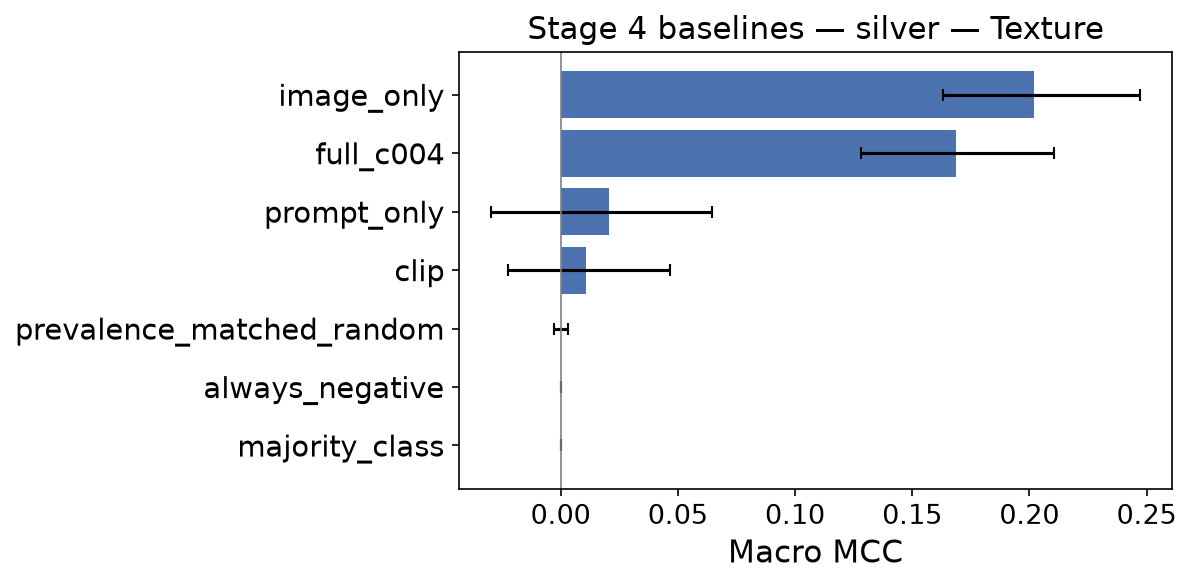}
\caption*{\scriptsize Texture}
\end{subfigure}
\caption{Simple baseline macro MCC on the 506-asset silver holdout ($\pm$95\% bootstrap CI).}
\label{fig:baselines_silver}
\end{figure}

\subsection{Generator-comparison verdicts depend on the judge}
\label{sec:generator_agreement}

\begin{figure}[tbp]
\centering
\includegraphics[width=\linewidth]{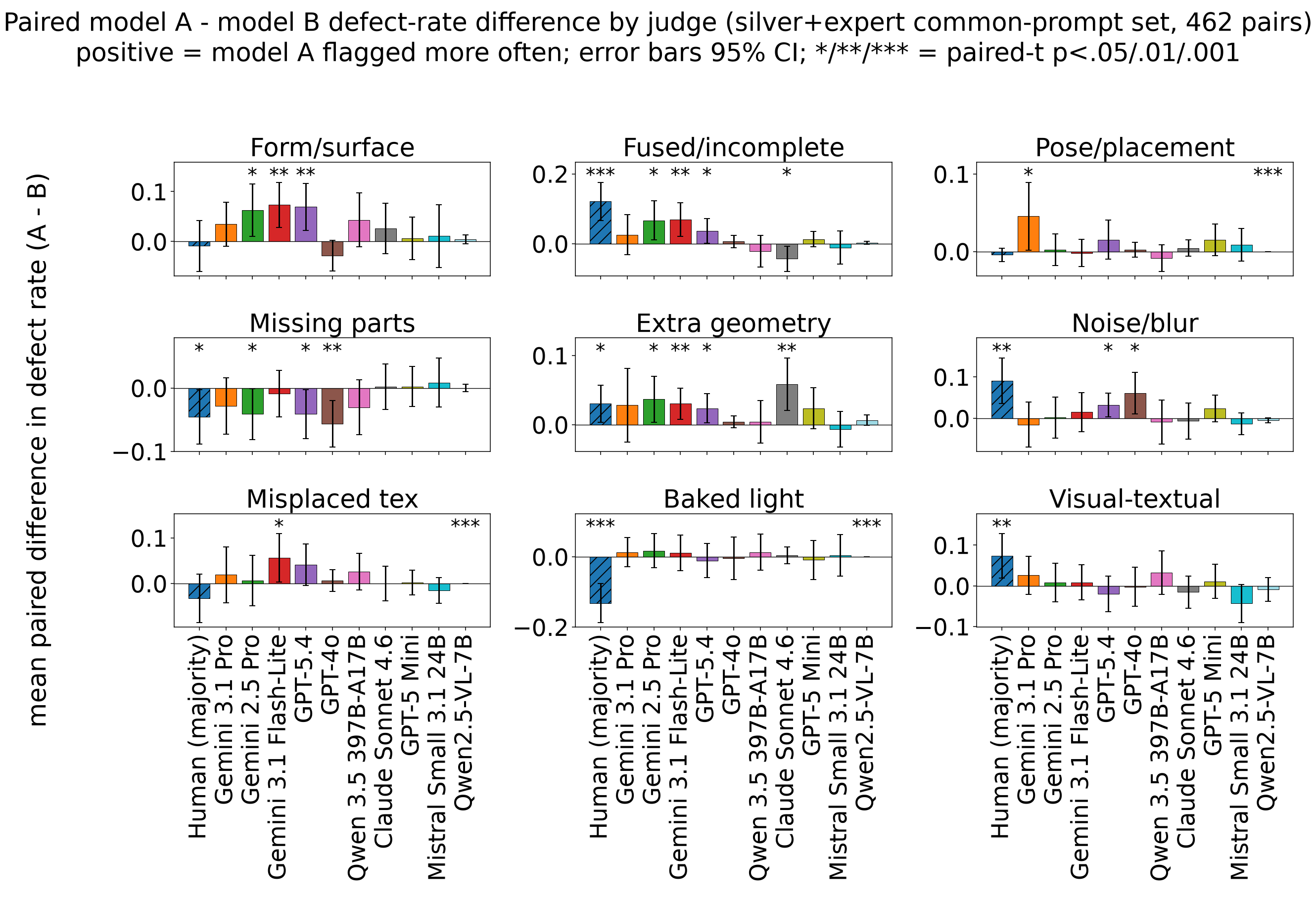}
\caption{\textbf{Paired generator comparison by judge} on the pooled silver$+$expert common-prompt set (462 prompts present in both arms; human reference is crowd majority on silver and expert majority on the expert assets). Each panel is a defect; each bar is a judge---human majority vote (hatched), then the 11 VLMs scored under \texttt{c004} on both splits, in leaderboard (Table~\ref{tab:leaderboard_combined}) order. Bar height is the mean paired defect-rate difference $\bar\delta=\mathrm{rate}_A-\mathrm{rate}_B$ (positive $=$ model~A flagged more), error bars are 95\% CIs, and \texttt{*}/\texttt{**}/\texttt{***} mark paired-$t$ $p{<}0.05/0.01/0.001$. One judge appears on only one side of the pool and is omitted: Claude~Opus~4.7, which is scored on the expert set but not on the silver grid. Judges frequently disagree in magnitude, significance, and occasionally sign.}
\label{fig:paired_compare}
\end{figure}

A common downstream use of a defect judge is \emph{model comparison}: deciding which generator produces fewer defects. We test whether judges agree on this task using a matched-pair design. Restricting to the \textbf{common-prompt set}---prompts for which \emph{both} generator arms (model~A, model~B) produced an asset---removes prompt-composition confounds, so any measured A$-$B difference reflects the generators, not the prompt mix. Because the silver and expert assets both carry human majority-vote labels, we \textbf{pool} the two splits into a single analysis over their combined common-prompt set (462 paired prompts). For each judge (the human majority vote and each of the 11 VLMs scored under \texttt{c004} on \emph{both} splits, ordered as in the Table~\ref{tab:leaderboard_combined} leaderboard) and each defect, we compute the per-prompt difference in defect rate $\delta = \mathrm{rate}_A - \mathrm{rate}_B$ and run a paired $t$-test ($H_0{:}\ \bar\delta{=}0$); Figure~\ref{fig:paired_compare} reports $\bar\delta$ with 95\% CIs and significance per defect$\times$judge.

By the human reference, the two generators differ significantly on \textbf{6 of 9} defects. \textbf{Judge performance varies substantially on this comparison task.} For each judge, we count how many of these six human-significant gaps point in the same direction---that is, identify the same generator as worse---regardless of the judge's own significance. Stronger judges agree on most defects (5/6 for GPT-5.4, GPT-5~Mini, GPT-4o, Gemini~2.5~Pro, and Gemini~3.1~Flash-Lite; 4/6 for Gemini~3.1~Pro), whereas weaker judges are near-random or opposite (0/6 for Mistral~Small, 1/6 for Claude~Sonnet~4.6, and 2/6 for Qwen2.5-VL-7B). Under a stricter criterion requiring the judge's gap to be both significant ($p{<}0.05$) and in the human direction, performance drops sharply: GPT-5.4 confirms 4/6, Gemini~2.5~Pro 3/6, and Gemini~3.1~Flash-Lite and GPT-4o 2/6, while several directionally consistent but under-powered judges fall to 0/6. Weak judges also report spurious gaps where humans find none; for example, Qwen2.5-VL-7B flags 3 of 9 defects as significant. Thus, generator rankings depend materially on the judge: frontier judges are usually directionally consistent on large, prevalent defects but remain unreliable on subtle and texture defects, while weaker judges are unsuitable for model ranking. The pipeline effect therefore propagates downstream evaluation use cases and decisions.

\subsection{Judge rankings transfer across human references on geometry, but not texture}
\label{sec:stage3silver}

\begin{figure}[tbp]
\centering
\begin{subfigure}{\linewidth}\centering\includegraphics[width=\linewidth]{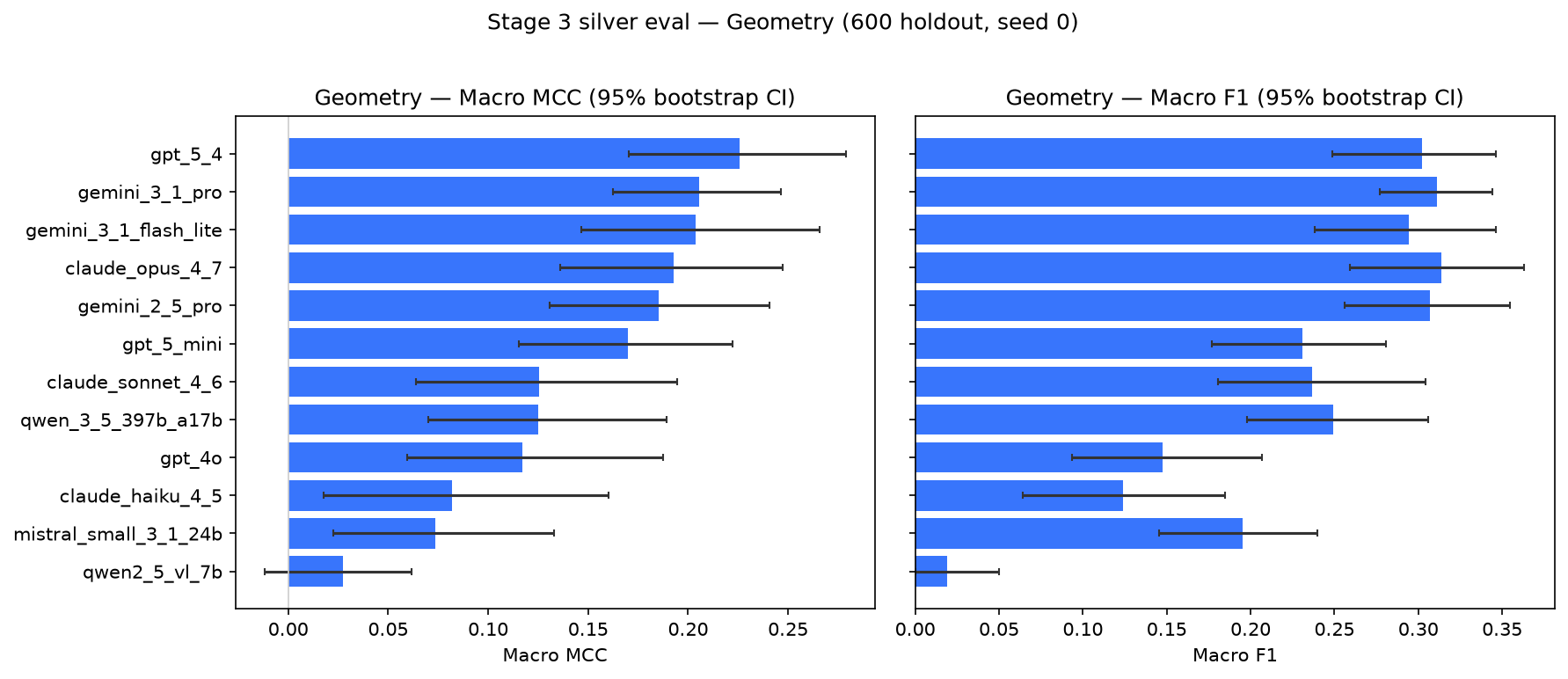}\caption*{\scriptsize Geometry}\end{subfigure}

\vspace{4pt}
\begin{subfigure}{\linewidth}\centering\includegraphics[width=\linewidth]{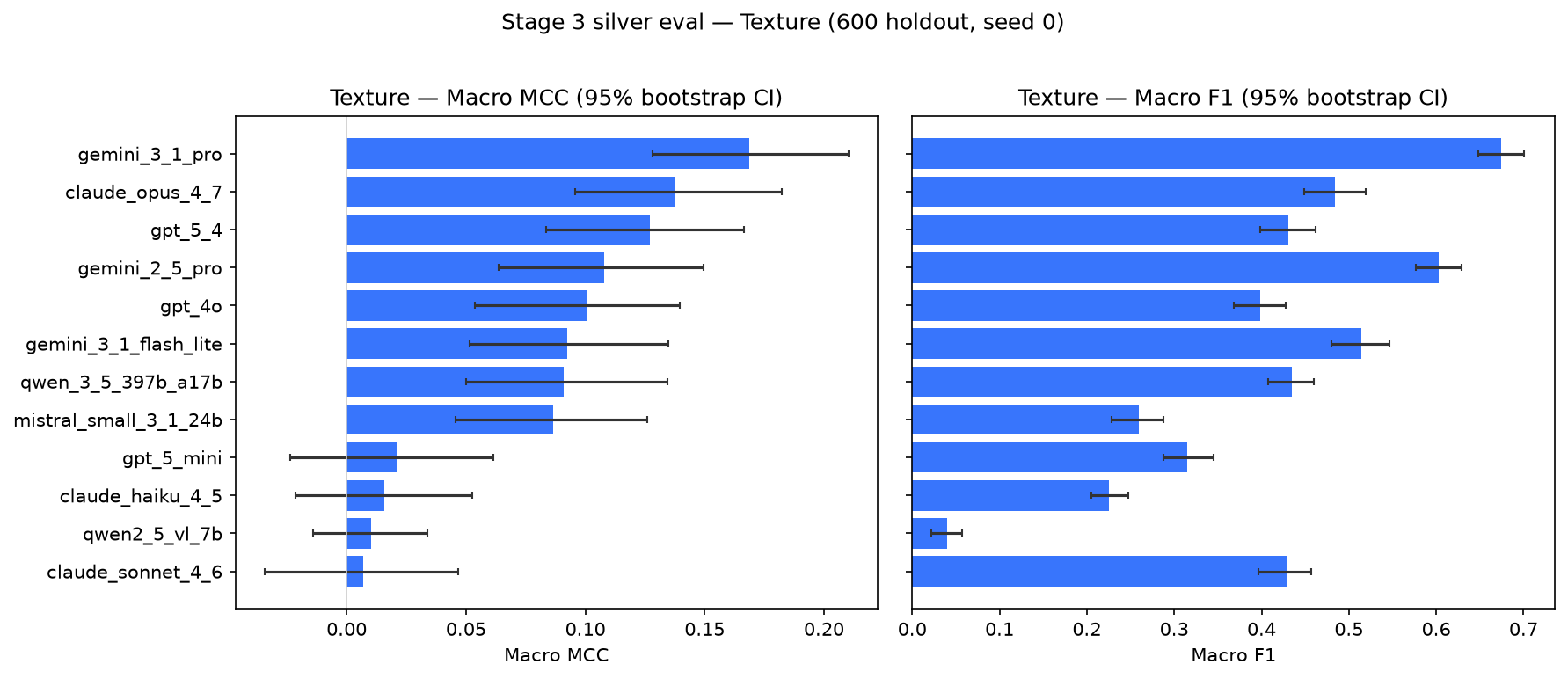}\caption*{\scriptsize Texture}\end{subfigure}
\caption{Silver-holdout macro MCC with 95\% asset-cluster confidence intervals for geometry and texture.}
\label{fig:silver_macro}
\end{figure}

\begin{figure}[tbp]
\centering
\begin{subfigure}{\linewidth}\centering\includegraphics[width=\linewidth]{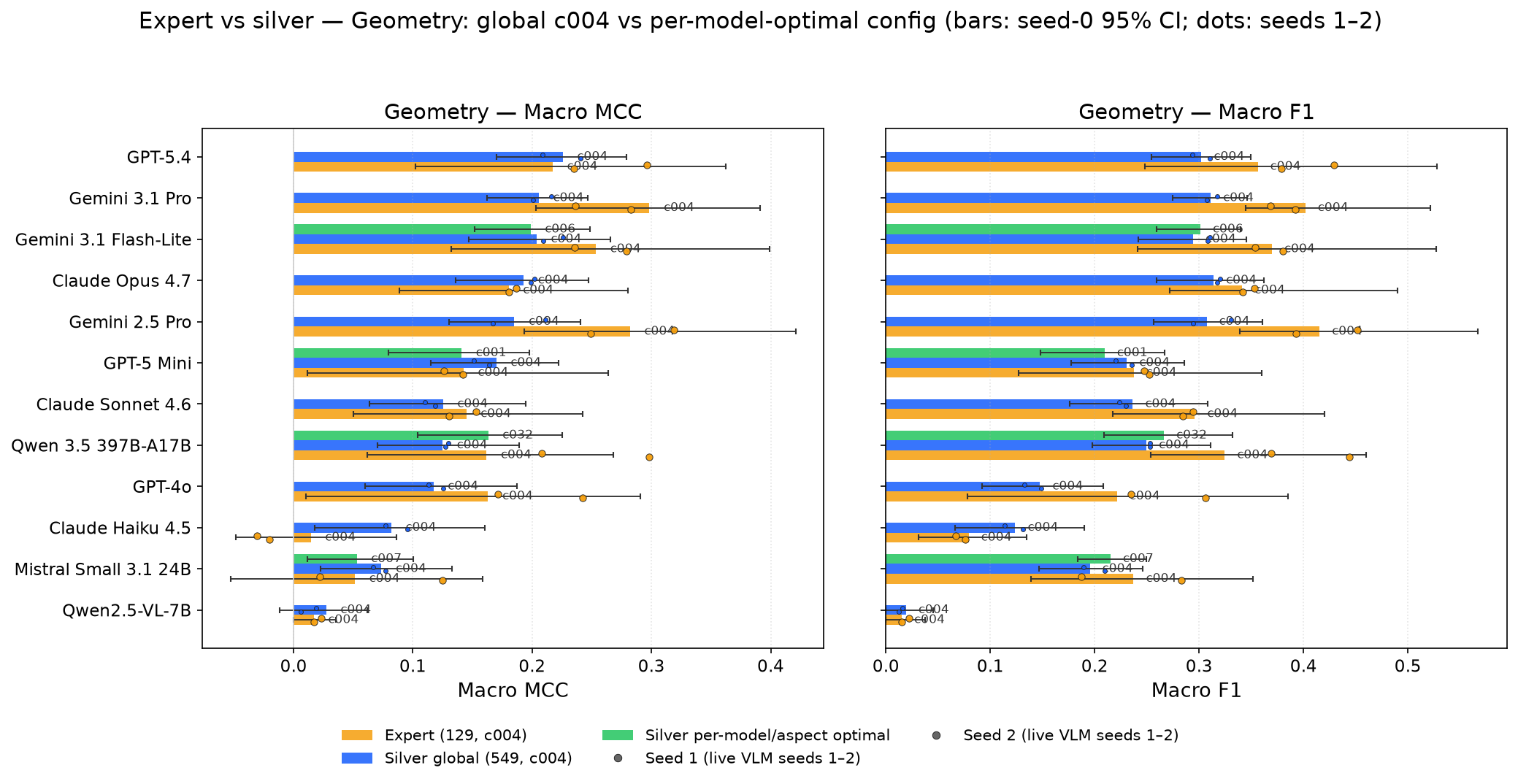}\caption*{\scriptsize Geometry}\end{subfigure}

\vspace{4pt}
\begin{subfigure}{\linewidth}\centering\includegraphics[width=\linewidth]{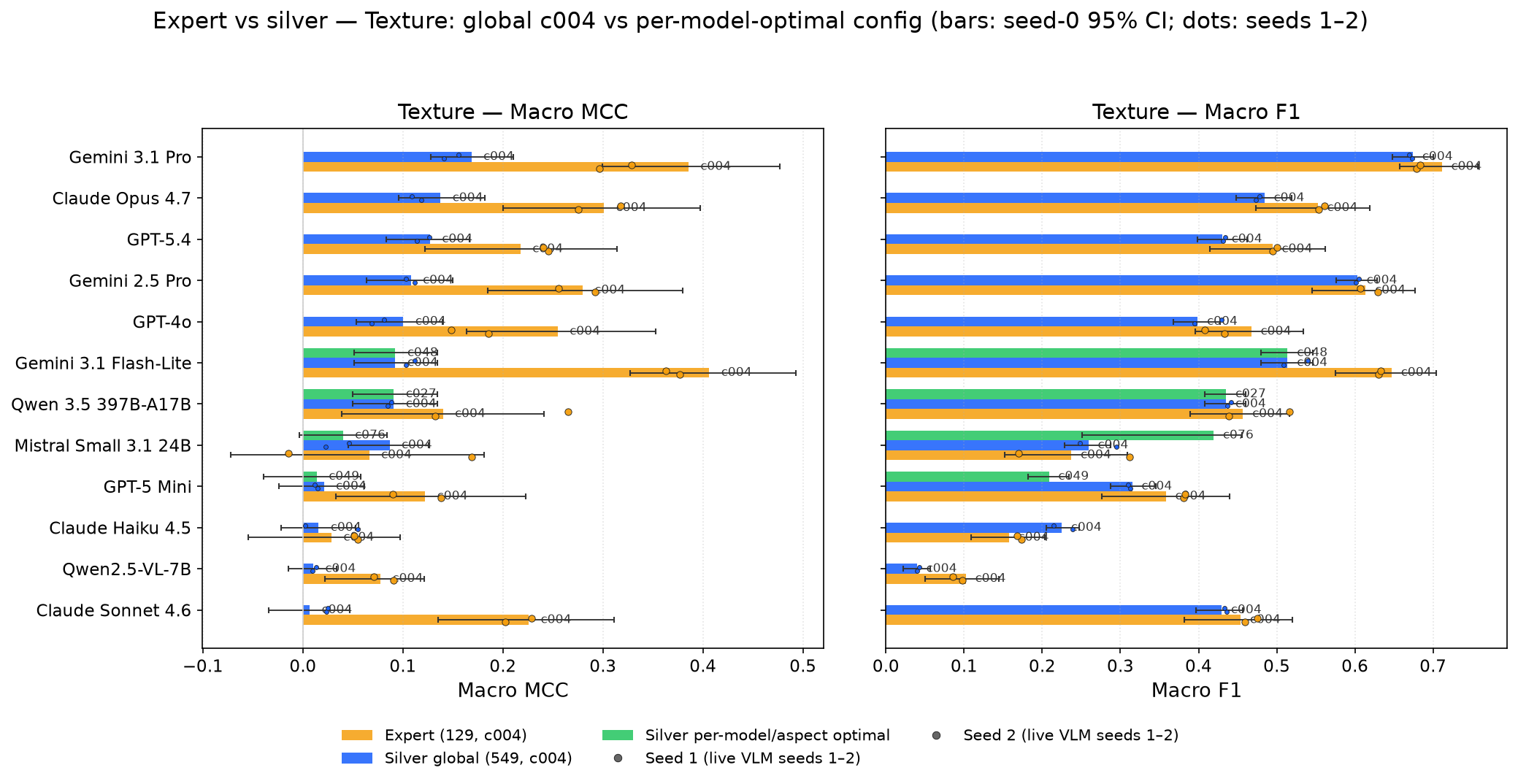}\caption*{\scriptsize Texture}\end{subfigure}
\caption{Expert-label vs.\ silver-holdout macro MCC per model. Bars show seed-0 point estimates with 95\% bootstrap CIs; dots show independent reruns for seeds 1--2 on expert and silver holdout arms.}
\label{fig:transfer}
\end{figure}

\begin{figure}[tbp]
\centering
\includegraphics[width=0.8\linewidth]{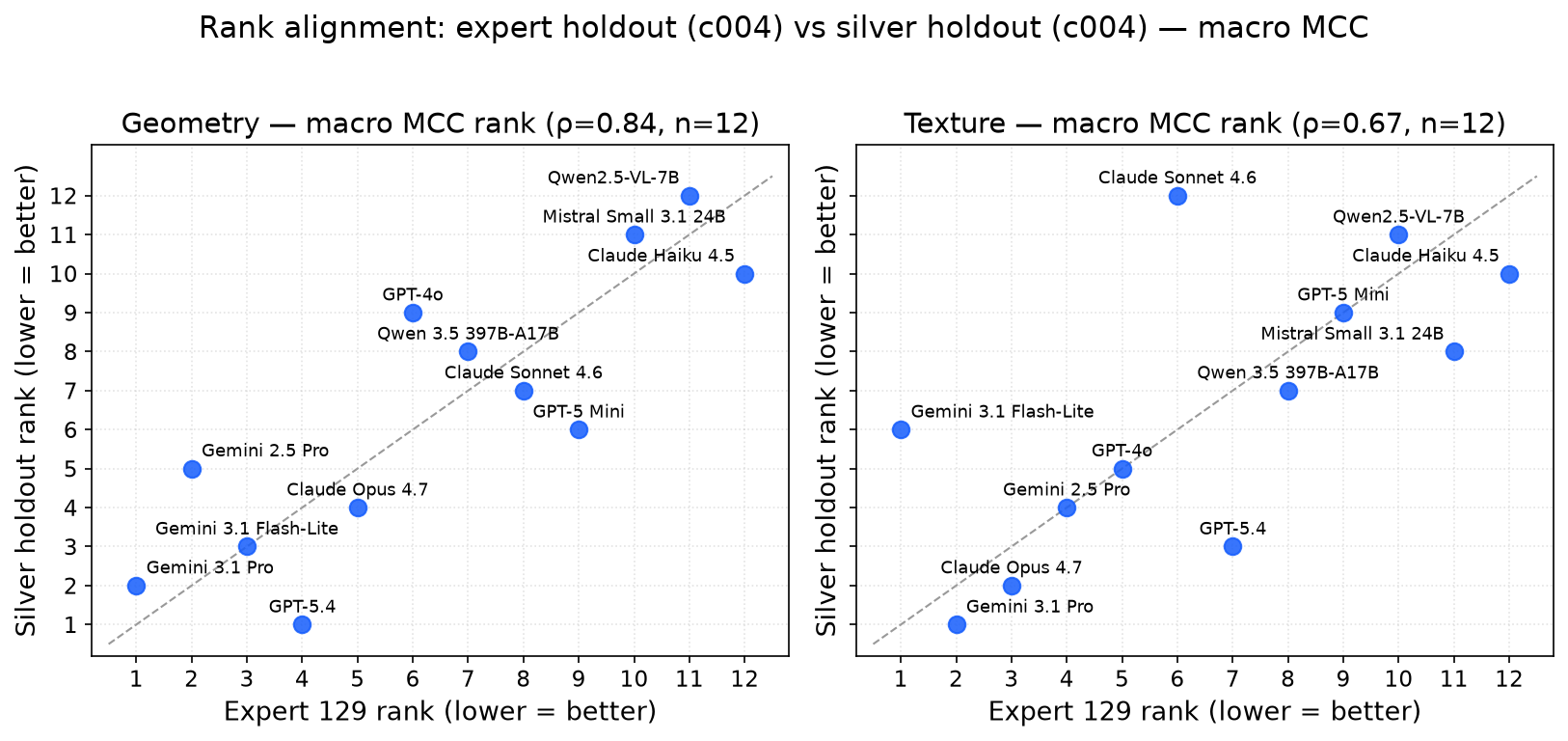}
\caption{Rank alignment: expert-label macro MCC rank vs.\ silver-holdout rank.}
\label{fig:rank_align}
\end{figure}

An evaluation pipeline depends not only on the judge model but also on the human reference used to measure it. We therefore ask whether the \emph{ranking of judges} is stable across two independently collected human reference systems: embedded 3D-artist experts (expert) and a larger pool of trained vendor annotators (silver). A stable ranking suggests the measured pipeline quality reflects the judges themselves, whereas a reordered leaderboard indicates sensitivity to the underlying human reference. The expert split provides cleaner labels but limited sample size, while the silver holdout provides broader coverage at higher label noise. To evaluate ranking robustness without reusing the data employed for screening and pipeline selection, we use a silver holdout of 506 assets disjoint from the 500-asset selection set, score judges against the silver majority vote, and compute asset-cluster bootstrap confidence intervals (Figure~\ref{fig:silver_macro}).
Figure~\ref{fig:transfer} compares expert-label and silver-holdout macro MCC for each judge. Geometry rankings are moderately preserved across the two references (Figure~\ref{fig:rank_align}), indicating that judge ordering is relatively robust to the choice of human reference. Texture rankings, however, differ substantially, and absolute MCC drops on the silver holdout (best 0.406$\rightarrow$0.168). This instability reflects the evaluation task itself rather than label quantity alone. Texture defects are considerably harder for humans to assess, exhibiting substantially lower inter-annotator agreement than geometry under both labeling systems (Table~\ref{tab:perdef}). Moreover, the two human reference systems disagree with each other on the same assets: on the 129 expert assets that also carry silver annotations, silver and expert majority votes agree on 78.6\% of geometry labels (Cohen's $\kappa=0.44$) but only 57.6\% of texture labels ($\kappa=0.16$), while experts consistently identify texture defects more frequently than the crowd (Table~\ref{tab:silver_expert_agree}). Thus, the two reference systems differ in both annotation noise and labeling threshold.

These results show that judge rankings can inherit the biases of the reference labeling system. When human agreement is high, as for geometry, pipeline conclusions transfer across reference populations. When the underlying task is difficult and human agreement is low, as for texture, different human populations induce different judge leaderboards. Consequently, judge rankings should be interpreted relative to the reference system on which they are measured, particularly for challenging evaluation tasks. Separately, Gemini~3.1~Flash-Lite remains competitive with frontier models on geometry under both references, suggesting that inexpensive models remain useful for pipeline selection even when final reporting relies on stronger judges. Our analysis is descriptive rather than prescriptive: it evaluates the robustness of judge rankings across the two human reference systems available in this benchmark rather than making claims about all future generators or labeler populations.

 \begin{table}[tbp]
  \centering
  \caption{Between-system human agreement on the 129 expert assets, all of which also carry silver crowd labels. Chance-corrected agreement
  (Cohen's $\kappa$) between the silver-majority and expert-majority verdicts is far higher on geometry than texture, mirroring the judge-ranking
  transfer gap and paralleling the within-system $\kappa$ of Table~\ref{tab:perdef}.}
  \label{tab:silver_expert_agree}
  \begin{tabular}{lccc}
  \toprule
  Aspect & \#Defects & Agreement & $\kappa_{\mathrm{silver}\leftrightarrow\mathrm{exp}}$ \\
  \midrule
  Geometry & 5 & 78.6\% & 0.44 \\
  Texture  & 4 & 57.6\% & 0.16 \\
  \bottomrule
  \end{tabular}
  \end{table}

\section{Positioning VLM judges relative to trained human labelers}
\label{sec:human}

Model scores are only meaningful relative to human-labeler variability. We therefore report two human-reference protocols that score a held-out human labeler and the VLMs against the same target labels.

\paragraph{Expert-labeler reference.}
The expert split has two labelers per asset. We randomly hold out one expert as the \emph{test labeler} for each asset and use the other expert as ground truth; VLMs are evaluated against the same ground-truth expert. This \emph{single held-out expert} target differs from the \emph{expert-agreement} target of the leaderboard (Table~\ref{tab:leaderboard_combined}): it scores every asset against one expert rather than restricting to cells where both experts agree, so absolute MCC is higher and the top VLM can differ. The held-out expert achieves macro MCC 0.519 on geometry and 0.312 on texture. The best VLMs on this single-expert protocol reach 0.403 geometry (Gemini~2.5~Pro) and 0.206 texture (Gemini~3.1~Flash-Lite), below the held-out expert on both aspects.

\paragraph{Silver leave-one-labeler-out reference.}
On the 506-asset silver validation holdout---each carrying exactly five crowd labels under the relabel-5 pass---we hold out one random labeler and use the majority vote of the remaining four as ground truth. The held-out crowd labeler achieves 0.316 geometry and 0.289 texture macro MCC. The best VLMs on the same protocol reach 0.259 geometry (Gemini~3.1~Flash-Lite) and 0.219 texture (Gemini~3.1~Pro), below the held-out labeler on both aspects. The uniform five-vote coverage lifts the human texture reference relative to the earlier pooled labels (0.289 vs.\ 0.215), consistent with the relabel-5 pass improving texture label quality.

\begin{table}[tbp]
\caption{Human vs.\ best VLM macro MCC under matched target-label protocols. Expert split: one expert held out as the test labeler, with the other expert as ground truth. Silver LOO: one crowd labeler held out, with the remaining-labeler majority as ground truth.}
\label{tab:human}
\centering
\footnotesize
\begin{tabular}{lrr}
\toprule
Evaluator & Geometry & Texture \\
\midrule
\multicolumn{3}{l}{\textit{Expert-labeler reference}} \\
\quad Human expert & 0.519 & 0.312 \\
\quad Best VLM & 0.403 (Gemini 2.5 Pro) & 0.206 (Gemini 3.1 Flash-Lite) \\
\midrule
\multicolumn{3}{l}{\textit{Silver leave-one-labeler-out ($n{=}506$, five votes)}} \\
\quad Human (LOO) & 0.316 & 0.289 \\
\quad Best VLM & 0.259 (Gemini 3.1 Flash-Lite) & 0.219 (Gemini 3.1 Pro) \\
\bottomrule
\end{tabular}
\end{table}

\begin{figure}[tbp]
\centering
\begin{subfigure}{\linewidth}\centering\includegraphics[width=\linewidth]{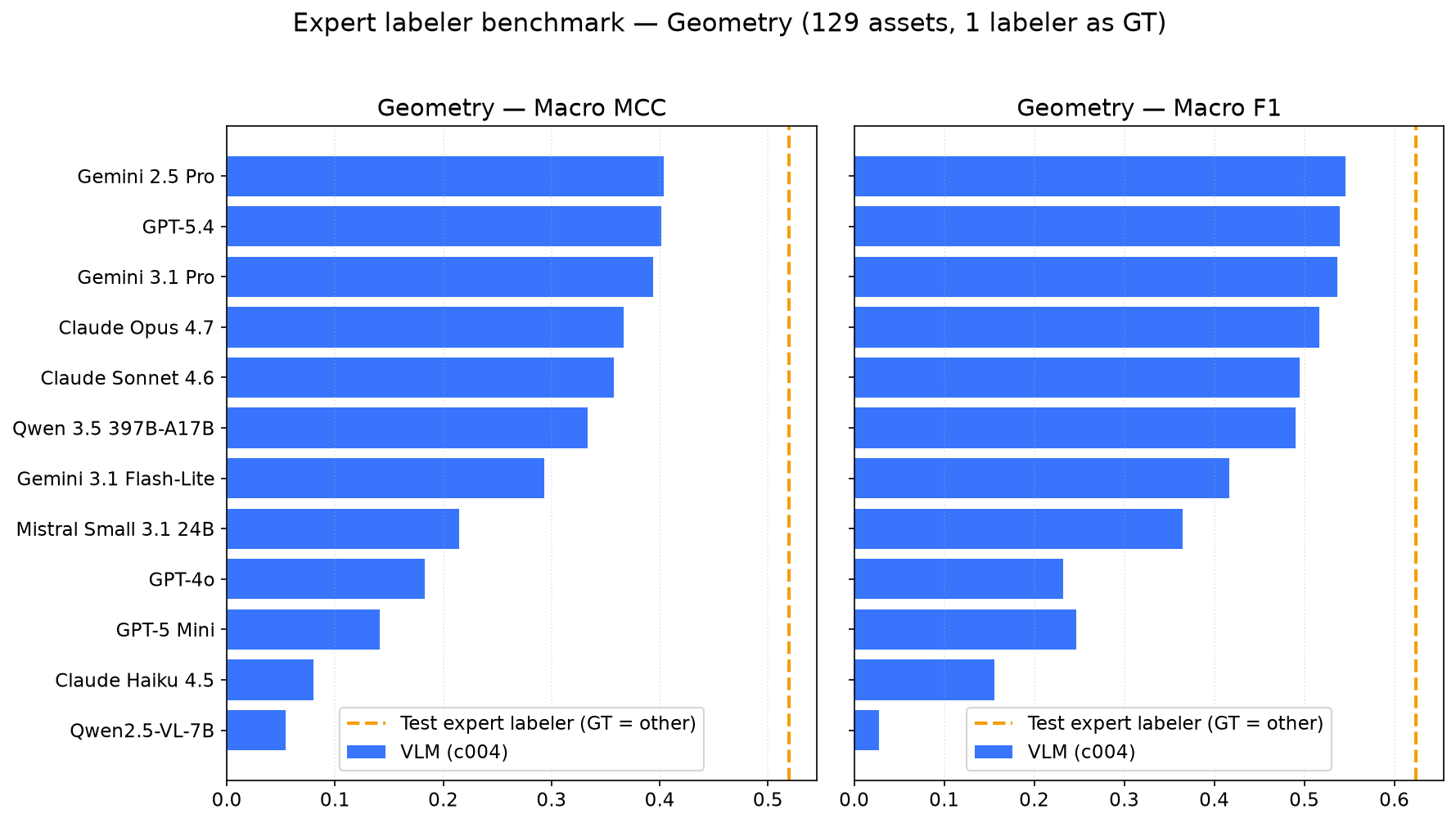}\caption*{\scriptsize Geometry}\end{subfigure}

\vspace{4pt}
\begin{subfigure}{\linewidth}\centering\includegraphics[width=\linewidth]{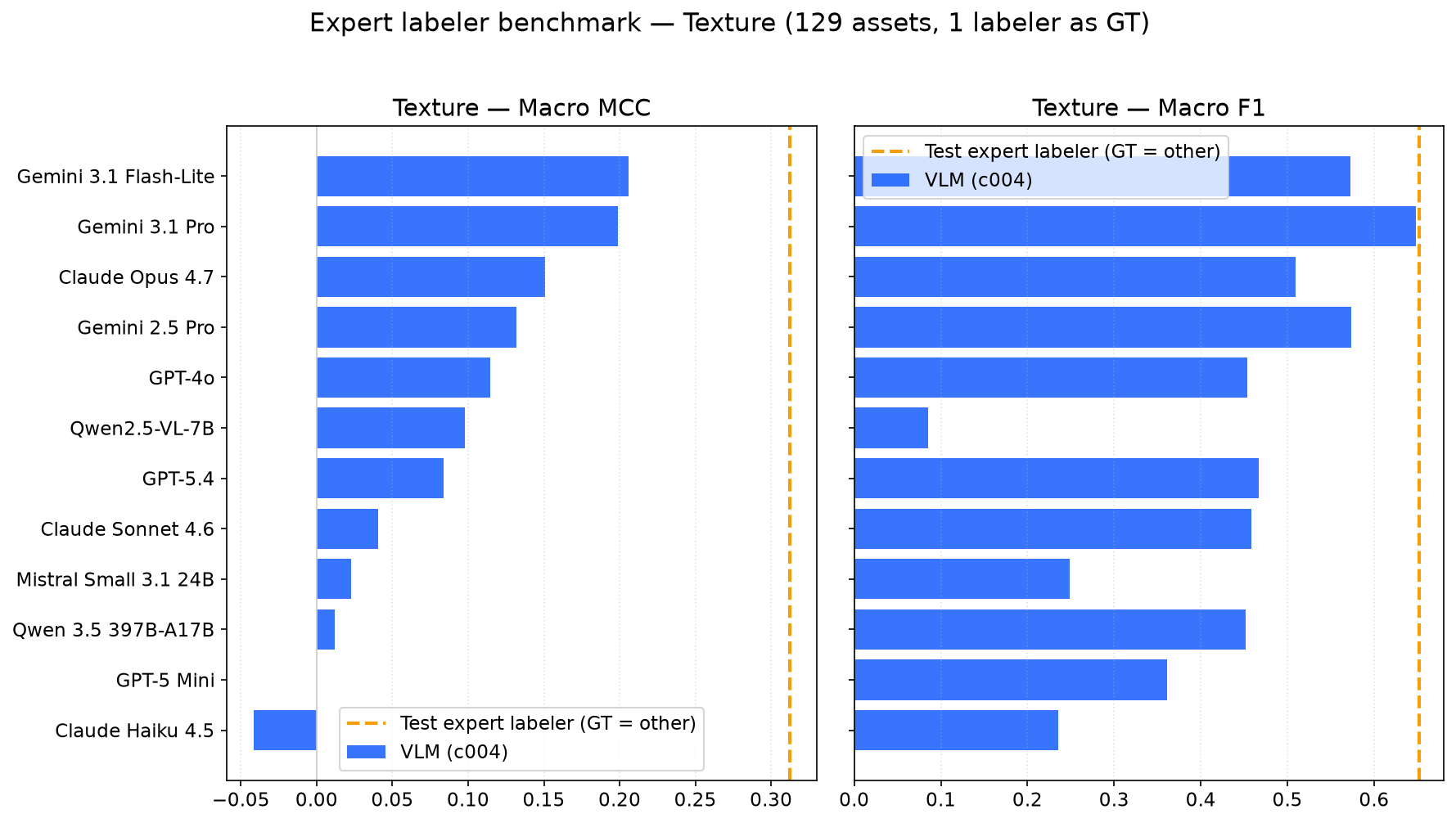}\caption*{\scriptsize Texture}\end{subfigure}
\caption{Expert-labeler benchmark: one expert labeler vs.\ VLM judges, all scored against the other expert labeler.}
\label{fig:golden_human}
\end{figure}

\begin{figure}[tbp]
\centering
\begin{subfigure}{\linewidth}\centering\includegraphics[width=\linewidth]{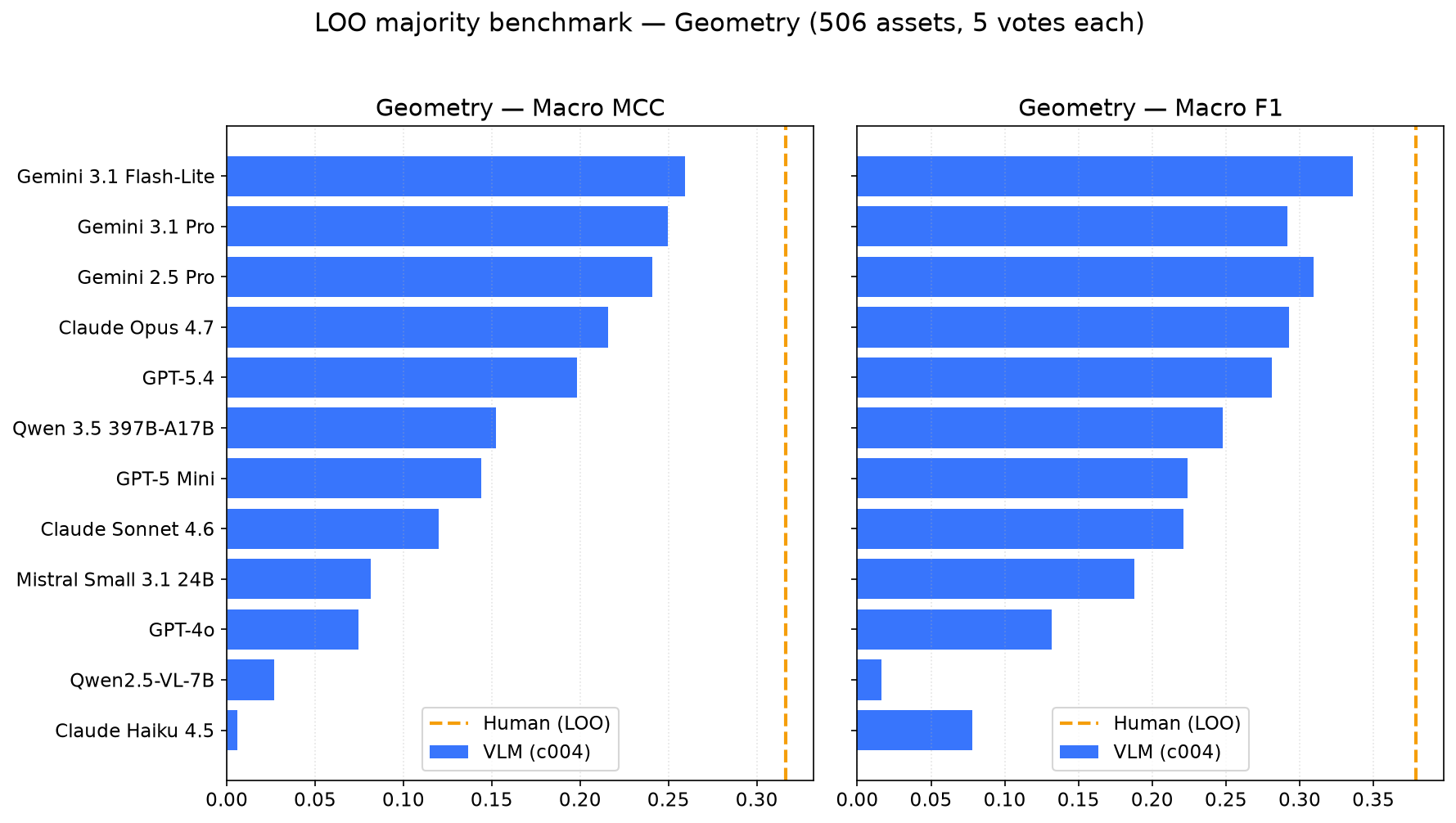}\caption*{\scriptsize Geometry}\end{subfigure}

\vspace{4pt}
\begin{subfigure}{\linewidth}\centering\includegraphics[width=\linewidth]{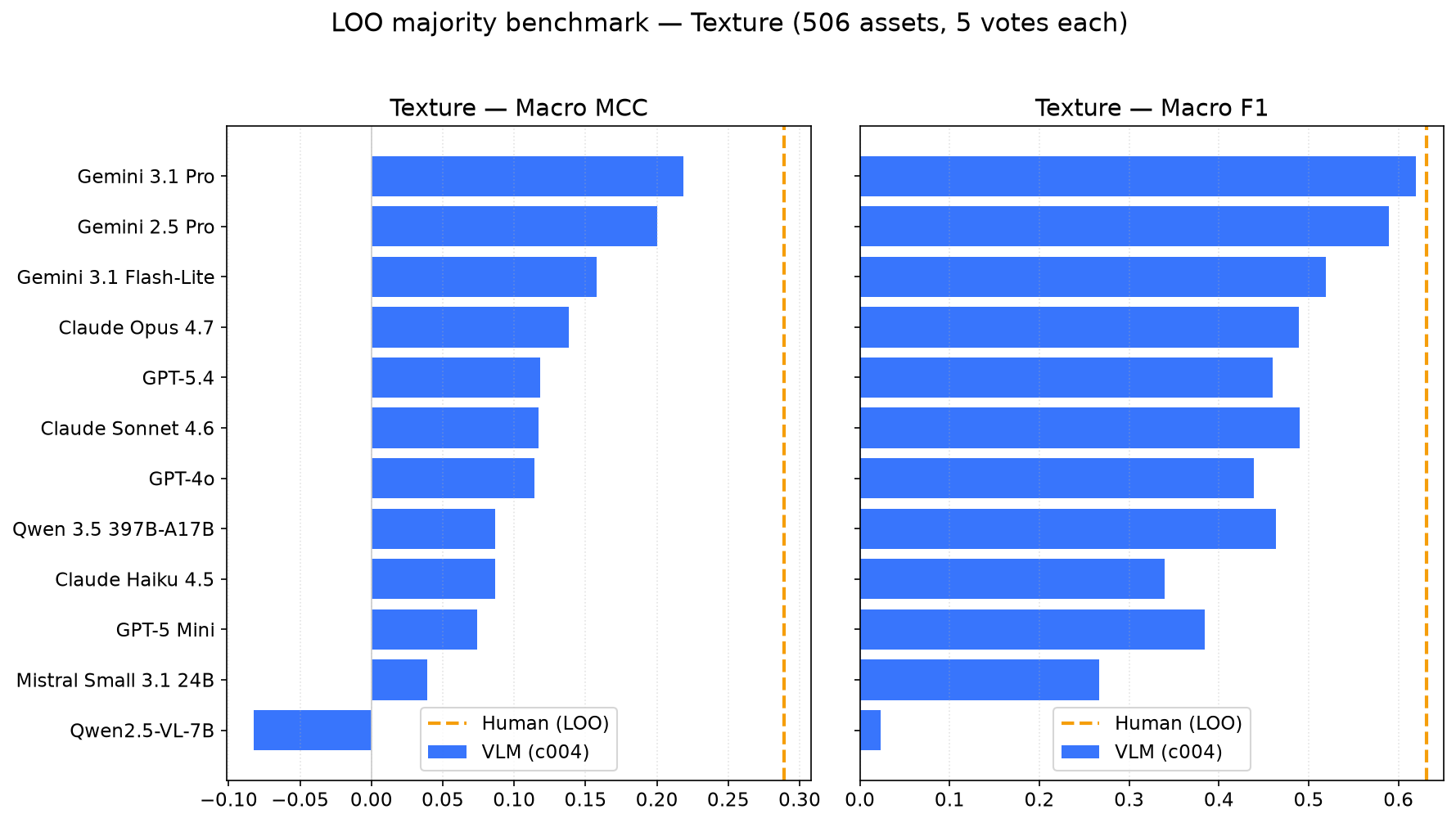}\caption*{\scriptsize Texture}\end{subfigure}
\caption{Silver leave-one-labeler-out benchmark: one crowd labeler vs.\ VLM judges, all scored against the majority of the remaining labelers.}
\label{fig:loo_human}
\end{figure}

\begin{figure}[t]
\centering
\includegraphics[width=0.82\linewidth]{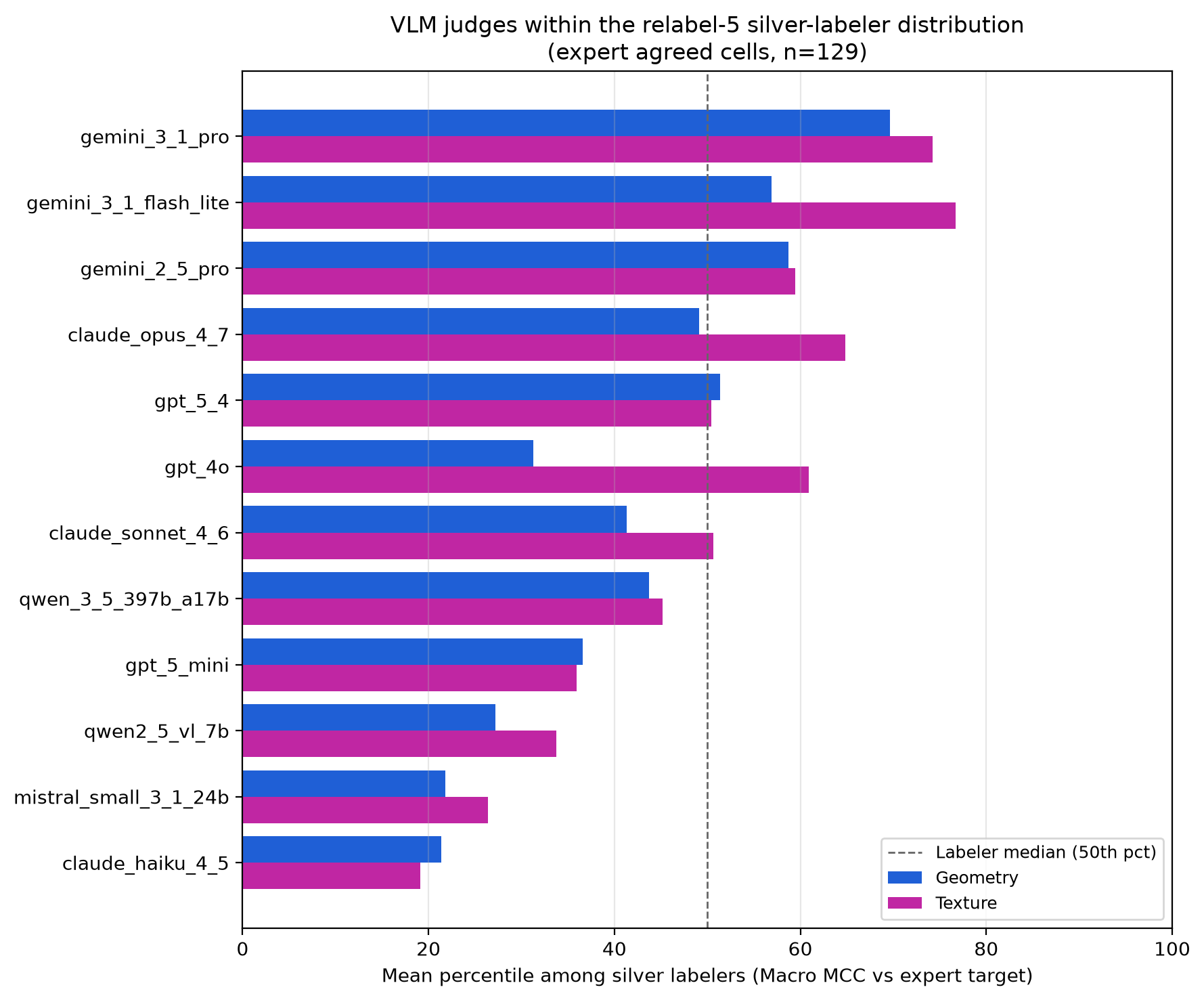}
\caption{VLM judges within the relabel-5 silver-labeler distribution on the expert set. Bars show each judge's mean percentile among individual silver labelers (macro MCC vs.\ the expert agreed-cell target, $n{=}129$ assets, 42 labelers / 32 anchors), separately for geometry and texture; the dashed line marks the labeler median. The strongest judges reach the upper third of the labeler distribution but none reach the top, and the texture gap present under the earlier crowd labels is closed after relabeling.}
\label{fig:silver_pct}
\end{figure}

\paragraph{Silver labeler percentiles in the expert set.}
As a third human reference, we place each VLM against the distribution of individual silver labelers who labeled the expert assets, using the relabel-5 crowd votes. For each anchor labeler (one who labeled $\geq 10$ expert assets) we score every VLM and every silver labeler against the expert target on that labeler's shared agreed cells, then record each VLM's percentile within the labeler distribution, averaged over anchors (Figure~\ref{fig:silver_pct}). The strongest VLMs sit near the 70th percentile of the silver-labeler distribution on geometry and the 74th--77th percentile on texture, while mid-tier models fall near or below the labeler median; no VLM reaches the top of the labeler distribution on either aspect. VLM judges are therefore competitive with, but do not surpass, individual trained labelers, consistent with the leave-one-out and expert-labeler benchmarks. Notably, under the relabel-5 labels the texture gap closes: whereas the earlier crowd labels placed VLMs below the labeler distribution on texture, the strongest judges now exceed the median labeler on texture as well, reflecting reduced texture-label noise after relabeling, so the residual human advantage is now concentrated on geometry.

\section{Limitations}
\label{sec:limitations}

\paragraph{Label noise and texture ceiling.}
Texture macro MCC is uniformly low on silver labels and sensitive to expert-to-silver shift. Crowd-label agreement is lower for texture defects (0.75--0.79 mean) than geometry (0.79--0.97), so VLMs may be partially capped by reference-label noise rather than perception limits alone.

\paragraph{Untested inference designs and future work.}
The screening grid varies camera-view protocol, visual input, and prompt schema, but holds the delivery format fixed: every judgment is a single stitched multi-view grid in one call. Separate per-view images, turntable video, and multi-turn interactive inspection are untested here and are plausibly stronger, as are reasoning-augmented prompting (chain-of-thought or rationale-augmented judgments~\citep{wei2022chain,sprague2024cot}) and per-defect targeted prompts rather than one shared schema. Results are descriptive over the observed 84-design grid rather than universal over all VLM-judge pipelines. Section~\ref{subsec:trellis_transfer} partially addresses generator scope with a 489-asset TRELLIS replication; broader inference designs and additional generators remain future work.

\paragraph{Model and generator coverage.}
The benchmark covers a fixed generator pool and 12 VLM families; transfer to new generators or future models is not guaranteed.

\paragraph{Unresolved error attribution.}
The benchmark measures end-to-end agreement with human labels. It does not determine whether a wrong prediction is caused by insufficient rendered evidence, a model perception error, a prompt/schema failure, or ambiguity in the human label. Controlled ablations with human-visible render audits would be needed to separate these causes. A related confound is that the human reference labels were produced from the textured RGB rendering---the same modality shown to the RGB judge---so the strong RGB effect may partly reflect alignment between the judge's input and the humans' viewing condition rather than color being intrinsically necessary for defect perception. Isolating this would require a human-side ablation, collecting reference labels from annotators shown only depth, normal, or untextured-geometry views, which we leave to future work.

\paragraph{Human-reference sample size.}
The expert split has 129 assets and two expert labels per asset. This is sufficient to expose large gaps between VLMs and held-out experts, but not to estimate fine-grained expert-labeler variability across many annotators. The silver leave-one-labeler-out analysis provides a complementary crowd-label reference but uses a different labeler population.

\paragraph{Compute and cost.}
The 84-design screening phase scores ${\sim}$5.4M defect cells (1.2M model calls on all 1{,}006 silver assets); the expert and silver model-comparison runs add 2{,}838 and 12{,}078 model calls respectively (Table~\ref{tab:accounting}). A per-phase breakdown of these counts is in Appendix~\ref{app:cost}.

\section{Conclusion}
\label{sec:conclusion}

We presented a methodology for rigorously studying automated VLM judges as \emph{evaluation pipelines} rather than isolated models, with 3D-DefectBench serving as its first large-scale instantiation for fine-grained 3D generation evaluation. The central premise is that the reliability of an automated judge depends not only on the underlying VLM, but also on how evidence is constructed, how the evaluation task is specified, and how human reference labels are collected and interpreted. Understanding automated evaluation therefore requires analyzing the complete measurement system rather than benchmarking judge models in isolation.

Our study demonstrates three key principles. First, \textbf{evaluation pipelines should be analyzed systematically}. Although VLM capability is the dominant determinant of agreement with human labels, camera protocol, visual input, and prompt schema all contribute measurable effects and interact with the underlying model. Second, \textbf{pipeline design should be studied and optimized through controlled experimentation}. A balanced factorial design, followed by cost-aware staged validation, identifies a compact six-view RGB protocol with a rubric-guided prompt as a strong cost--performance trade-off while showing that denser rendering protocols and additional geometric channels provide limited benefit within the studied design space (though the RGB advantage may partly reflect alignment with the RGB-based human reference). Third, \textbf{automated judges should be evaluated against carefully characterized human labeling systems, rather than against human labels treated as unquestioned ground truth}. Human label quality depends on annotator expertise, rubric interpretation, and task difficulty, so measured VLM agreement can reflect weaknesses in the reference labels as well as in the judge. Our tiered design---trained silver labelers, expert rubric authors, and multiple independent annotations per asset---allows us to quantify human-label reliability, compare VLMs across reference regimes, and study how reference quality influences apparent judge performance.

Beyond the specific findings for 3D generation, we believe the broader contribution is methodological. Many GenAI evaluation systems rely on automated judges operating over constructed inputs and imperfect human references. Rather than viewing these systems simply as foundation models applied to an evaluation task, we argue that they should be treated as \emph{measurement systems} whose reliability emerges from the interaction between model capability, input construction, task specification, statistical evaluation, and human reference labels. Understanding such systems requires more than reporting a leaderboard: it requires systematically studying which design choices matter, how they interact, how conclusions transfer across settings, and how the resulting measurements relate to human judgment. We hope this work helps move automated evaluation from \emph{model-centric benchmarking} toward rigorous \emph{measurement-system analysis}.

We release 3D-DefectBench, including labels, prompts, model predictions, Croissant metadata, and reproducible analysis code, on HuggingFace at \url{https://huggingface.co/datasets/zzhao0500/3D-DefectBench}.
\FloatBarrier
\section*{Acknowledgments}
\addcontentsline{toc}{section}{Acknowledgments}
We gratefully acknowledge our colleagues at Roblox---Kiran Bhat, Tinghui Zhou, and Daniel Chin---for their support of this work. We thank Kiran Bhat and Tinghui Zhou for sharing their insights on developing and using this evaluation rubric system for model development, and Daniel Chin for his contributions to guiding the annotation resources and practice.

\bibliographystyle{plainnat}
\bibliography{references}

\appendix

\section{Taxonomy column mapping}
\label{app:taxonomy}
The nine defect categories are: form/surface quality, fused or incomplete parts, pose/placement mismatch, missing parts, extra geometry, noise/blur/grain, misplaced or overlapping texture, baked lighting/shadow artifacts, and visual-textual mismatch. The prompt-conditioned categories are missing parts, pose/placement mismatch, and visual-textual mismatch. Released tables include stable machine-readable column names; Table~\ref{tab:taxonomy_mapping} maps them to the reader-facing labels used throughout.

\begin{table}[htbp]
\caption{Mapping from released column names to reader-facing defect labels.}
\label{tab:taxonomy_mapping}
\centering
\footnotesize
\begin{tabular}{ll}
\toprule
Released column & Reader-facing label \\
\midrule
\texttt{q\_form\_surface} & Form or surface quality \\
\texttt{q\_fused\_incomplete} & Fused or incomplete parts \\
\texttt{q\_pose\_placement} & Pose or placement mismatch \\
\texttt{q\_missing\_parts} & Missing parts \\
\texttt{q\_extra\_geometry} & Extra geometry \\
\texttt{t\_noisy\_blurry\_grainy} & Noise, blur, or grain \\
\texttt{t\_misplaced\_overlapping} & Misplaced or overlapping texture \\
\texttt{t\_baked\_lighting\_shadow} & Baked lighting or shadow \\
\texttt{t\_incorrect\_visual\_textual} & Visual-textual mismatch \\
\bottomrule
\end{tabular}
\end{table}

\section{Full 84 inference-design grid}
\label{app:config_grid}
Configurations are the Cartesian product of 3 camera-view protocols, 7 visual inputs, and 4 prompt schemas (84 total). Released configuration IDs are deterministic identifiers used only for reproducibility. The selected evaluation configuration corresponds to released ID \texttt{c004}; Table~\ref{tab:config_mapping} gives the human-readable mapping and Table~\ref{tab:config_grid_full} lists the full grid.

\begin{table}[htbp]
\caption{Human-readable mapping for the selected inference design.}
\label{tab:config_mapping}
\centering
\footnotesize
\begin{tabular}{lll}
\toprule
Design component & Released value & Meaning \\
\midrule
Camera protocol & \texttt{6\_oblique\_tb} & Six oblique turntable views around the asset \\
Visual input & \texttt{RGB} & Standard color renders only \\
Prompt schema & Rubric-Guided Checklist & Per-defect checklist with rubric definitions \\
\bottomrule
\end{tabular}
\end{table}

\begin{longtable}{llll}
\caption{Full 84-configuration inference-design grid. Configuration IDs are deterministic reproducibility keys.}\\
\label{tab:config_grid_full}\\
\toprule
ID & Camera protocol & Visual input & Prompt schema \\
\midrule
\endfirsthead
\multicolumn{4}{c}{\tablename\ \thetable{} -- continued from previous page} \\
\toprule
ID & Camera protocol & Visual input & Prompt schema \\
\midrule
\endhead
\midrule
\multicolumn{4}{r}{Continued on next page} \\
\endfoot
\bottomrule
\endlastfoot
\texttt{c001} & 6 oblique & RGB & Compact Binary \\
\texttt{c002} & 6 oblique & RGB & Definition-Guided Binary \\
\texttt{c003} & 6 oblique & RGB & Rubric-Guided Binary \\
\texttt{c004} & 6 oblique & RGB & Rubric-Guided Checklist \\
\texttt{c005} & 6 oblique & Geometry-only & Compact Binary \\
\texttt{c006} & 6 oblique & Geometry-only & Definition-Guided Binary \\
\texttt{c007} & 6 oblique & Geometry-only & Rubric-Guided Binary \\
\texttt{c008} & 6 oblique & Geometry-only & Rubric-Guided Checklist \\
\texttt{c009} & 6 oblique & Normals & Compact Binary \\
\texttt{c010} & 6 oblique & Normals & Definition-Guided Binary \\
\texttt{c011} & 6 oblique & Normals & Rubric-Guided Binary \\
\texttt{c012} & 6 oblique & Normals & Rubric-Guided Checklist \\
\texttt{c013} & 6 oblique & Depth & Compact Binary \\
\texttt{c014} & 6 oblique & Depth & Definition-Guided Binary \\
\texttt{c015} & 6 oblique & Depth & Rubric-Guided Binary \\
\texttt{c016} & 6 oblique & Depth & Rubric-Guided Checklist \\
\texttt{c017} & 6 oblique & RGB + geometry & Compact Binary \\
\texttt{c018} & 6 oblique & RGB + geometry & Definition-Guided Binary \\
\texttt{c019} & 6 oblique & RGB + geometry & Rubric-Guided Binary \\
\texttt{c020} & 6 oblique & RGB + geometry & Rubric-Guided Checklist \\
\texttt{c021} & 6 oblique & RGB + normals & Compact Binary \\
\texttt{c022} & 6 oblique & RGB + normals & Definition-Guided Binary \\
\texttt{c023} & 6 oblique & RGB + normals & Rubric-Guided Binary \\
\texttt{c024} & 6 oblique & RGB + normals & Rubric-Guided Checklist \\
\texttt{c025} & 6 oblique & RGB + depth & Compact Binary \\
\texttt{c026} & 6 oblique & RGB + depth & Definition-Guided Binary \\
\texttt{c027} & 6 oblique & RGB + depth & Rubric-Guided Binary \\
\texttt{c028} & 6 oblique & RGB + depth & Rubric-Guided Checklist \\
\texttt{c029} & 14\_eq\_tb & RGB & Compact Binary \\
\texttt{c030} & 14\_eq\_tb & RGB & Definition-Guided Binary \\
\texttt{c031} & 14\_eq\_tb & RGB & Rubric-Guided Binary \\
\texttt{c032} & 14\_eq\_tb & RGB & Rubric-Guided Checklist \\
\texttt{c033} & 14\_eq\_tb & Geometry-only & Compact Binary \\
\texttt{c034} & 14\_eq\_tb & Geometry-only & Definition-Guided Binary \\
\texttt{c035} & 14\_eq\_tb & Geometry-only & Rubric-Guided Binary \\
\texttt{c036} & 14\_eq\_tb & Geometry-only & Rubric-Guided Checklist \\
\texttt{c037} & 14\_eq\_tb & Normals & Compact Binary \\
\texttt{c038} & 14\_eq\_tb & Normals & Definition-Guided Binary \\
\texttt{c039} & 14\_eq\_tb & Normals & Rubric-Guided Binary \\
\texttt{c040} & 14\_eq\_tb & Normals & Rubric-Guided Checklist \\
\texttt{c041} & 14\_eq\_tb & Depth & Compact Binary \\
\texttt{c042} & 14\_eq\_tb & Depth & Definition-Guided Binary \\
\texttt{c043} & 14\_eq\_tb & Depth & Rubric-Guided Binary \\
\texttt{c044} & 14\_eq\_tb & Depth & Rubric-Guided Checklist \\
\texttt{c045} & 14\_eq\_tb & RGB + geometry & Compact Binary \\
\texttt{c046} & 14\_eq\_tb & RGB + geometry & Definition-Guided Binary \\
\texttt{c047} & 14\_eq\_tb & RGB + geometry & Rubric-Guided Binary \\
\texttt{c048} & 14\_eq\_tb & RGB + geometry & Rubric-Guided Checklist \\
\texttt{c049} & 14\_eq\_tb & RGB + normals & Compact Binary \\
\texttt{c050} & 14\_eq\_tb & RGB + normals & Definition-Guided Binary \\
\texttt{c051} & 14\_eq\_tb & RGB + normals & Rubric-Guided Binary \\
\texttt{c052} & 14\_eq\_tb & RGB + normals & Rubric-Guided Checklist \\
\texttt{c053} & 14\_eq\_tb & RGB + depth & Compact Binary \\
\texttt{c054} & 14\_eq\_tb & RGB + depth & Definition-Guided Binary \\
\texttt{c055} & 14\_eq\_tb & RGB + depth & Rubric-Guided Binary \\
\texttt{c056} & 14\_eq\_tb & RGB + depth & Rubric-Guided Checklist \\
\texttt{c057} & 14\_multi\_ring\_tb & RGB & Compact Binary \\
\texttt{c058} & 14\_multi\_ring\_tb & RGB & Definition-Guided Binary \\
\texttt{c059} & 14\_multi\_ring\_tb & RGB & Rubric-Guided Binary \\
\texttt{c060} & 14\_multi\_ring\_tb & RGB & Rubric-Guided Checklist \\
\texttt{c061} & 14\_multi\_ring\_tb & Geometry-only & Compact Binary \\
\texttt{c062} & 14\_multi\_ring\_tb & Geometry-only & Definition-Guided Binary \\
\texttt{c063} & 14\_multi\_ring\_tb & Geometry-only & Rubric-Guided Binary \\
\texttt{c064} & 14\_multi\_ring\_tb & Geometry-only & Rubric-Guided Checklist \\
\texttt{c065} & 14\_multi\_ring\_tb & Normals & Compact Binary \\
\texttt{c066} & 14\_multi\_ring\_tb & Normals & Definition-Guided Binary \\
\texttt{c067} & 14\_multi\_ring\_tb & Normals & Rubric-Guided Binary \\
\texttt{c068} & 14\_multi\_ring\_tb & Normals & Rubric-Guided Checklist \\
\texttt{c069} & 14\_multi\_ring\_tb & Depth & Compact Binary \\
\texttt{c070} & 14\_multi\_ring\_tb & Depth & Definition-Guided Binary \\
\texttt{c071} & 14\_multi\_ring\_tb & Depth & Rubric-Guided Binary \\
\texttt{c072} & 14\_multi\_ring\_tb & Depth & Rubric-Guided Checklist \\
\texttt{c073} & 14\_multi\_ring\_tb & RGB + geometry & Compact Binary \\
\texttt{c074} & 14\_multi\_ring\_tb & RGB + geometry & Definition-Guided Binary \\
\texttt{c075} & 14\_multi\_ring\_tb & RGB + geometry & Rubric-Guided Binary \\
\texttt{c076} & 14\_multi\_ring\_tb & RGB + geometry & Rubric-Guided Checklist \\
\texttt{c077} & 14\_multi\_ring\_tb & RGB + normals & Compact Binary \\
\texttt{c078} & 14\_multi\_ring\_tb & RGB + normals & Definition-Guided Binary \\
\texttt{c079} & 14\_multi\_ring\_tb & RGB + normals & Rubric-Guided Binary \\
\texttt{c080} & 14\_multi\_ring\_tb & RGB + normals & Rubric-Guided Checklist \\
\texttt{c081} & 14\_multi\_ring\_tb & RGB + depth & Compact Binary \\
\texttt{c082} & 14\_multi\_ring\_tb & RGB + depth & Definition-Guided Binary \\
\texttt{c083} & 14\_multi\_ring\_tb & RGB + depth & Rubric-Guided Binary \\
\texttt{c084} & 14\_multi\_ring\_tb & RGB + depth & Rubric-Guided Checklist \\
\end{longtable}

\section{Full selected prompt text}
\label{app:prompts}
For reference, we reproduce the full text of the selected geometry prompt (Rubric-Guided Checklist) below; the texture prompt is structurally identical with texture-specific categories. The other three geometry schemas are this prompt with progressively less rubric detail and a single-line output format (Table~\ref{tab:prompt_schemas}).

{\footnotesize
\begin{verbatim}
You are a 3D model quality evaluator. Your task is to assess ONLY the geometry quality
of a generated 3D asset by comparing the provided text prompt against a multi-view PNG
image of the asset.

The PNG image contains a grid of renderings from different viewpoints. Carefully inspect
all views before making a judgment. If a geometry defect is visible in any view, even if
only partially visible, mark that defect as present.

Evaluation scope:
- Evaluate geometry only.
- Ignore texture, color, material, lighting, shadow, transparency, surface pattern,
  decals, logos, and text readability unless they directly reveal a geometry issue.
- Use the prompt to determine required objects, parts, quantity, pose, placement, and
  structural expectations.
- Do not over-penalize fantastical or stylized objects if they still satisfy the prompt
  and are structurally coherent.
- Do not assume hidden defects that are not visually supported.
- If the asset is clearly a completely different object category from the prompt, mark
  the relevant prompt-adherence geometry defects.
- If the prompt is too ambiguous to determine the expected object, make the best visual
  judgment based only on clearly observable defects.

Rate the model's geometry using the following five defect categories. For each category,
output 1 if the defect is present and 0 if the defect is clearly not present.

1. Incorrect Prompt Quantity / Missing Parts

Mark 1 if the number of objects or major parts does not match the prompt or common
expectation for the object. This includes:
- Required objects or parts are missing.
- Major anatomy, facial features, wheels, limbs, blades, handles, cockpit, doors,
  windows, or other essential components are absent.
- The prompt specifies a quantity and the generated count is wrong.
- Extra separate objects appear when the prompt asks for one integrated object.
- Prompt-specific components are missing even if the overall object category is
  recognizable.

Examples:
- A prompt for "robot with a shield" produces a robot without the shield.
- A bus is missing a required wheel.
- A windmill has the wrong number of blades.
- A humanoid statue is missing arms.
- A monster truck is missing its cockpit and only produces the chassis.
- A turtle is missing its mouth and nose.
- A hotel courtesy shuttle is missing many required prompt elements.
- A mobile pet-grooming van is missing key van features such as cockpit, doors, and
  windows.
- A prompt for "three cubes" produces only two cubes.

Do not mark this category for minor simplification unless an important part is missing
or the prompt clearly requires the detail.

2. Incorrect Prompt Pose / Placement

Mark 1 if the object's pose, orientation, action, or spatial relation does not match the
prompt. This includes:
- The requested pose or action is wrong, such as standing vs sitting.
- A part is oriented in the wrong direction.
- Parts are attached in the wrong place.
- Object-to-object placement is wrong, such as an object under another object when the
  prompt says it should be on top.
- Components are visibly misplaced, such as wheels too high, sails facing the wrong
  direction, or structural parts clipping into the wrong area.

Examples:
- A bear is on all fours when it should be standing.
- A boat's sail faces the wrong direction.
- A car wheel is positioned too high and clips through the car body.
- An apple appears under a plate when the prompt asks for an apple on a plate.
- A ship's tail fin or sail is positioned incorrectly.

3. Fused or Incomplete Parts

Mark 1 if geometry is improperly formed, disconnected, merged, broken, or structurally
incomplete. This includes:
- Separate parts that should remain distinct are fused, melted, or over-connected.
- Parts that should connect are floating, detached, or disconnected.
- Surfaces contain holes, gaps, missing sections, or unsealed geometry.
- Components are truncated, unfinished, or only partially formed.
- Internal structure is incomplete enough to make the object structurally incorrect.

Examples:
- A dog's legs are fused together.
- A frog's front arms and rear legs merge into one continuous mesh near the torso.
- A hand is fused with the object it is holding.
- A monitor is not connected to its stand.
- Harp strings are disconnected from the frame.
- A chest has a visible hole in the main body, and the lid and base are fused.
- A mountain mesh has visible holes.
- A fairy's legs are unintentionally merged together.
- A sailboat has an unintended fused connection between the sail and the boat body, and
  part of the sail is missing.
- A tennis racket has net strings that are fused together or inconsistently separated.

Do not mark this category for intended artistic simplification unless the form is
visibly broken, fused, disconnected, or incomplete.

4. Unrecognizable Extra Geometry

Mark 1 if there are geometric elements that should not exist and cannot be interpreted
as meaningful parts of the prompt. This includes:
- Random blobs, protrusions, floating pieces, internal junk geometry, or loose
  structures.
- Extra appendages or shapes not implied by the object.
- Nonsensical interior geometry.
- Geometry that makes the model confusing even if the main object is recognizable.

Examples:
- A car's exterior shell is recognizable, but the interior cabin contains dense, blobby,
  nonsensical geometry that does not correspond to seats, dashboard, or other correct
  interior components.
- A half-human, half-dog model contains unrecognizable extra geometry on the back of the
  shoe.
- A sailboat has floating unrecognizable geometry beside the hull.
- A cargo ship has floating extra geometry that should not be part of the model.
- A Skibidi toilet contains unrecognizable extra geometry inside the toilet.
- A burger cart has extra geometry on top of the burger that does not correspond to a
  meaningful component.

Do not mark this category for legitimate decorative or stylized elements that plausibly
belong to the object.

5. Form and Surface Quality Issues

Mark 1 if the object's shape, proportions, symmetry, surface, or structural fidelity is
poor. This includes:
- Warped, lumpy, jagged, collapsed, noisy, or uneven geometry.
- Incorrect proportions, asymmetric parts that should be symmetric, or distorted
  anatomy.
- Important structural details are overly smoothed, rounded, softened, or missing.
- Surfaces lack physical plausibility, sharpness, thickness, or defining form.
- The object is recognizable but the geometric form is low quality or lacks expected
  structural coherence.

Examples:
- A wreath or ornaments are overly smooth and lose expected detail.
- A fire engine lacks defined details between the window and grille, and the siren
  lights are warped.
- A star has uneven surface quality and lacks symmetry.
- Mansion stairs have warped or uneven edges, and pillars are too round or lack detail.
- A hamster's whisker geometry is uneven and lacks symmetry.
- A dog's head has sharp edges that should not be present on the cheek or nose.
- A turtle has uneven bumpiness on the shell and lacks sharpness/detail around the eyes.
- A stack of money has an incorrect square shape instead of a rectangular shape.
- A Roman stone building has blobby pillars and warped pediment surfaces.
- A flower has uneven petals or an asymmetric center.

Output format:
Return exactly the following JSON object and nothing else:

{
    "geometry_checklist": {
        "q_missing_parts": {
            "label": 0 or 1,
            "evidence": "short evidence or not observed"
        },
        "q_pose_placement": {
            "label": 0 or 1,
            "evidence": "short evidence or not observed"
        },
        "q_fused_incomplete": {
            "label": 0 or 1,
            "evidence": "short evidence or not observed"
        },
        "q_extra_geometry": {
            "label": 0 or 1,
            "evidence": "short evidence or not observed"
        },
        "q_form_surface": {
            "label": 0 or 1,
            "evidence": "short evidence or not observed"
        }
    },
    "geometry_rating": [q_missing_parts, q_pose_placement, q_fused_incomplete,
  q_extra_geometry, q_form_surface]
}

Each label must be 0 or 1.
Keep each evidence field concise.
Do not include any text outside the JSON.
\end{verbatim}

}

\section{Factor effect sizes}
\label{app:effect_sizes}

\begin{figure}[tbp]
\centering
\begin{subfigure}{\linewidth}\centering\includegraphics[width=0.85\linewidth]{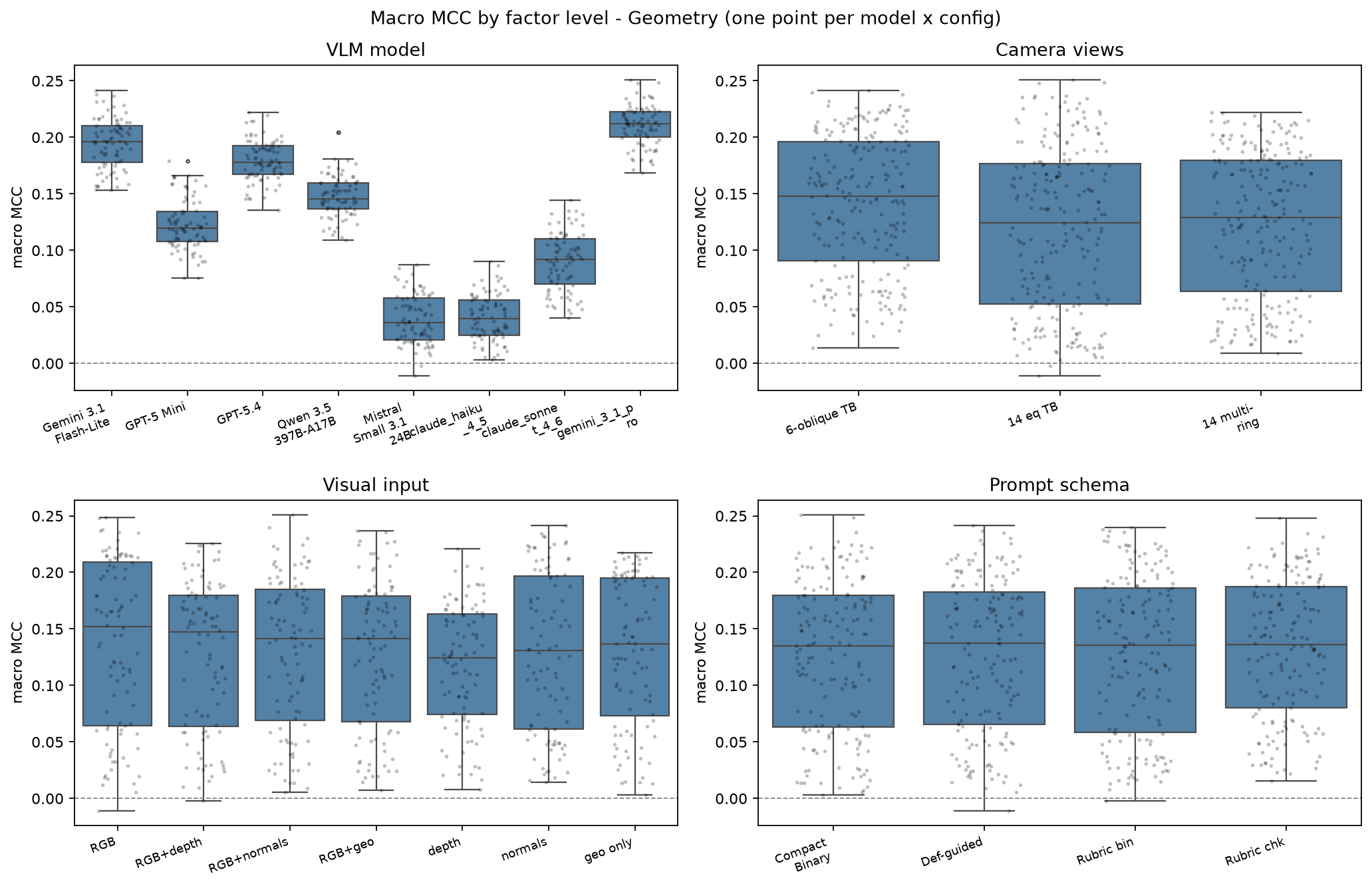}\caption*{\scriptsize Geometry}\end{subfigure}

\vspace{4pt}
\begin{subfigure}{\linewidth}\centering\includegraphics[width=0.85\linewidth]{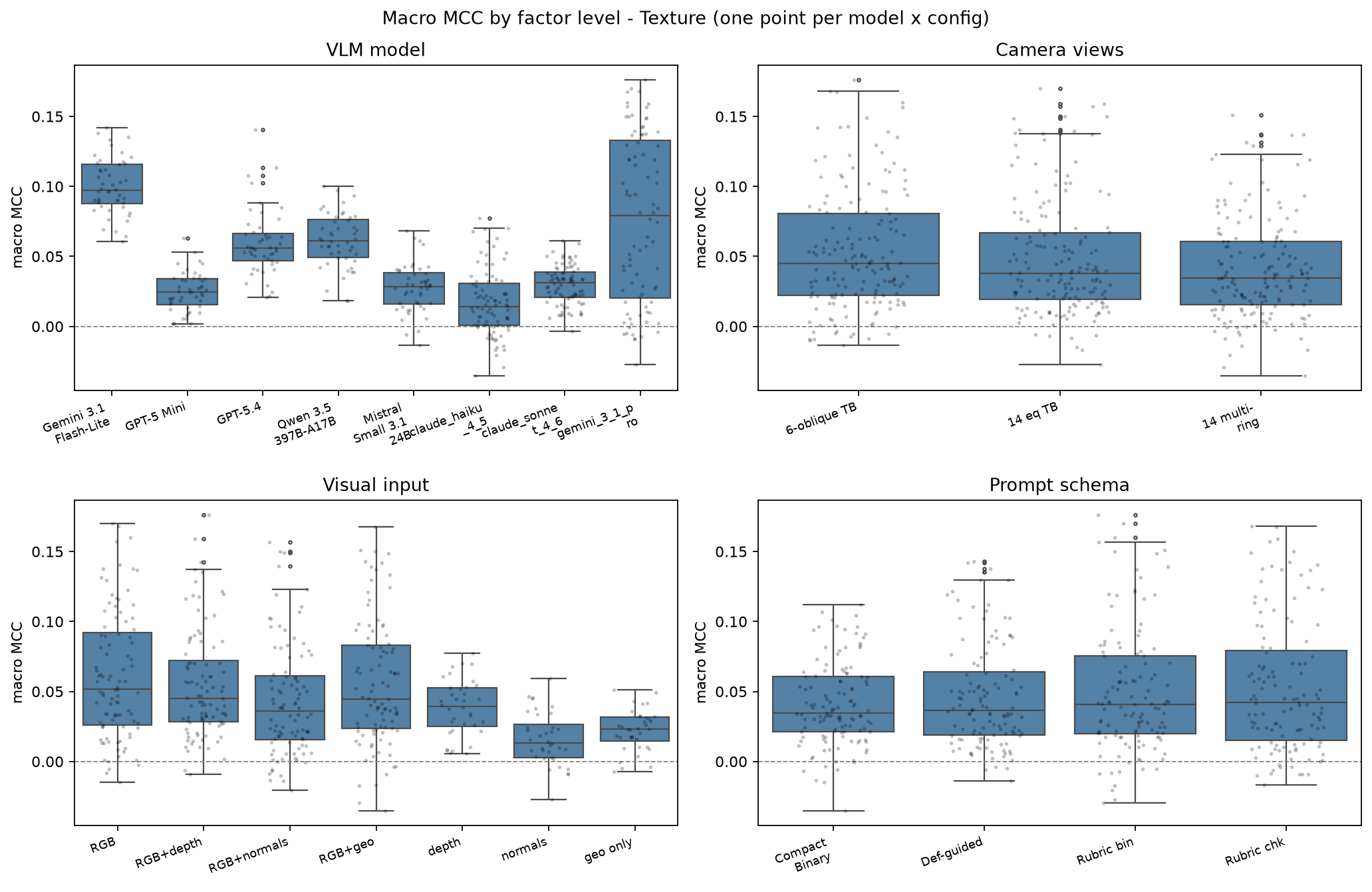}\caption*{\scriptsize Texture}\end{subfigure}
\caption{Macro MCC by factor level (one point per model$\times$config) in the factor-analysis phase.}
\label{fig:stage1_boxplots}
\end{figure}

Figure~\ref{fig:stage1_boxplots} shows the distribution of macro MCC (one point per model$\times$config) across the levels of each factor, separately for geometry and texture---the raw signal underlying the fitted coefficients. Table~\ref{tab:effect_sizes} translates the small partial pseudo-$R^2$ values ($\sim$$10^{-4}$--$10^{-3}$) into the best-vs-worst level gap in pooled macro MCC: the VLM-model factor moves geometry macro MCC by $\sim$0.17, an order of magnitude larger than the prompt-schema gap ($\sim$0.01), matching the pseudo-$R^2$ ordering.

\begin{table}[tbp]
\caption{Best-vs-worst level gaps in pooled screening macro MCC per factor (silver labels).}
\label{tab:effect_sizes}
\centering
\footnotesize
\begin{tabular}{llrlr}
\toprule
Aspect & Factor & Best MCC & Worst level & $\Delta$ MCC \\
\midrule
Geometry & VLM model & 0.207 & Mistral Small 3.1 24B & 0.169 \\
Geometry & Visual input & 0.132 & depth & 0.022 \\
Geometry & Camera protocol & 0.132 & 14 equatorial & 0.019 \\
Geometry & Prompt schema & 0.126 & Definition-Guided Binary & 0.009 \\
Texture & VLM model & 0.121 & Claude Haiku 4.5 & 0.117 \\
Texture & Prompt schema & 0.065 & Compact Binary & 0.023 \\
Texture & Visual input & 0.060 & RGB+normals & 0.018 \\
Texture & Camera protocol & 0.060 & 14 multi-ring & 0.014 \\
\bottomrule
\end{tabular}
\end{table}

\section{Statistical methodology}
\label{subsec:stats}
\paragraph{Model and importance.} We fit Eq.~\eqref{eq:factor} (defined at the head of Section~\ref{sec:stage1}) for each of the nine defects with $L_2$-regularized logistic regression (scikit-learn, \texttt{C}${=}10^6$, \texttt{lbfgs}, \texttt{tol}${=}10^{-7}$), i.e.\ effectively unregularized so the fit approximates the maximum-likelihood solution. Writing $\ell_{\text{full}}$ for the fitted log-likelihood and $\ell_{0}$ for the intercept-only log-likelihood, the McFadden pseudo-$R^2$ is $R^2_{\text{McF}} = 1 - \ell_{\text{full}}/\ell_{0}$. For factor family $f$ we refit without that family's design block ($\ell_{-f}$) and report the \emph{partial} pseudo-$R^2$, $\Delta R^2_f = R^2_{\text{full}} - R^2_{-f}$, as the family's importance, together with an omnibus likelihood-ratio test $\mathrm{LR}_f = 2(\ell_{\text{full}} - \ell_{-f})$, referred to a $\chi^2_{\mathrm{df}_f}$ distribution with $\mathrm{df}_f$ equal to the number of coefficients dropped with family $f$.

\paragraph{Multiple testing.} Across the 45 omnibus tests (9 defects $\times$ 5 factor families) we report Benjamini--Hochberg FDR-adjusted $p$-values alongside raw $p$-values. At $q{<}0.05$ the significant-defect counts are unchanged from the raw counts (VLM model 9/9, visual input 7/9, prompt schema 8/9, camera protocol 6/9, interactions 9/9), so the family ordering is robust to FDR control.

\paragraph{Clustering and repeated measures.} Each asset contributes many correlated rows to a defect's regression---one per (model, config) cell in which it was scored---so the rows are not independent. The omnibus LR $\chi^2$ tests pool these rows as if independent and therefore yield \emph{anti-conservative} $p$-values; we accordingly use them only to \emph{rank} factor families by importance rather than as calibrated significance statements, and we verify that the ranking survives FDR control. Every interval we rely on for inference---macro MCC per model and per configuration, bootstrap coefficient CIs, and the configuration/model \emph{difference} CIs (Table~\ref{tab:config_paired})---instead uses an asset-cluster bootstrap that resamples whole assets with replacement, so all defect cells from one mesh move together and within-asset correlation is carried into the interval. This is the conservative interval underlying all ranking and separation claims in the paper.

\section{Configuration-selection stability and paired tests}
\label{subsec:selection}
To keep configuration selection disjoint from the 506-asset validation holdout of Section~\ref{sec:stage3golden}, we perform it on the complementary 500-asset configuration-selection split (the two are mutually exclusive and exhaustive over the 1{,}006 valid assets). On that split \texttt{c004} tops the pooled silver-screening ranking (geometry macro MCC 0.156; nearest rivals \texttt{c006} 0.152, \texttt{c008} 0.151, \texttt{c020} 0.150). The point-estimate gap to the next configuration is tiny ($\sim$0.004 macro MCC) and the paired bootstrap intervals below overlap zero, so \texttt{c004} is not separable from the top cluster on point estimates alone. We therefore assess selection stability with an asset-cluster bootstrap (2{,}000 resamples of the 500-asset selection split; for each resample we recompute pooled per-config macro MCC and record the rank-1 and top-3 configurations). Table~\ref{tab:selection_stability} reports the resulting winner frequencies: \texttt{c004} is rank-1 in 37\% (geometry) and 74\% (texture) of resamples and top-3 in 76\%/96\%. The high top-3 frequency together with the small point-estimate gaps supports treating \texttt{c004} as a pragmatic default that is reliably near-optimal, rather than a uniquely optimal design. Selection uses noisy silver labels, so we do not claim global optimality.

\begin{table}[tbp]
\caption{Bootstrap configuration-selection stability (2{,}000 asset-cluster resamples). Rank-1 and top-3 frequency for the leading configurations.}
\label{tab:selection_stability}
\centering
\footnotesize
\begin{tabular}{llrr}
\toprule
Aspect & Config & Rank-1 freq. & Top-3 freq. \\
\midrule
Geometry & \texttt{c004} & 0.37 & 0.76 \\
Geometry & \texttt{c006} & 0.21 & 0.53 \\
Geometry & \texttt{c008} & 0.13 & 0.44 \\
Geometry & \texttt{c020} & 0.11 & 0.39 \\
\midrule
Texture & \texttt{c004} & 0.74 & 0.96 \\
Texture & \texttt{c032} & 0.19 & 0.82 \\
Texture & \texttt{c028} & 0.01 & 0.26 \\
Texture & \texttt{c076} & 0.02 & 0.19 \\
\bottomrule
\end{tabular}
\end{table}

\paragraph{Is \texttt{c004} separably better than its rivals?} Rank frequencies show \emph{which} configuration wins a resample but not whether the winning margin is statistically real. We therefore test \texttt{c004} directly against its four nearest rivals with a paired asset-cluster bootstrap of the pooled macro-MCC \emph{difference} (Table~\ref{tab:config_paired}). The result reinforces the ``near-optimal default, not uniquely optimal'' reading. On geometry, \texttt{c004} is statistically \emph{indistinguishable} from its three closest rivals---\texttt{c001} (95\% difference CI $[-0.012,\,0.025]$ spans zero), \texttt{c020} $[-0.013,\,0.024]$, and \texttt{c003} $[-0.002,\,0.027]$---and separably beats only \texttt{c032} $[+0.005,\,0.041]$. On texture the ordering is sharper: \texttt{c004} separably beats \texttt{c001}, \texttt{c003}, and \texttt{c020} (all difference CIs exclude zero) and is statistically tied only with \texttt{c032} $[-0.009,\,0.021]$. No single design is uniquely best: on both aspects \texttt{c004} sits inside a small cluster of statistically tied top configurations, which is why we treat it as a pragmatic default rather than an optimum and do not tune downstream conclusions to it. Because these are eight pre-specified comparisons against a single anchor we report the raw difference CIs without multiplicity correction; a Bonferroni adjustment ($\alpha{=}0.05/8$) would only widen the intervals and strengthen the ``tied cluster'' conclusion.

\begin{table}[h]
\caption{Paired asset-cluster bootstrap comparison of the carried-forward \texttt{c004} against its nearest screening rivals (pooled macro MCC on the 500-asset configuration-selection split, 2{,}000 resamples). $\Delta$ is \texttt{c004} minus the rival; a 95\% difference CI excluding zero marks \texttt{c004} as separably better. We read the difference CIs---not the bootstrap tail probabilities---as the inferential object, and apply no multiplicity correction across the eight comparisons.}
\label{tab:config_paired}
\centering
\footnotesize
\begin{tabular}{llrlc}
\toprule
Aspect & Comparison & $\Delta$ macro MCC & 95\% CI & Separable \\
\midrule
Geometry & \texttt{c004} vs.\ \texttt{c032} & 0.024 & [0.005, 0.041] & \ding{51} \\
Geometry & \texttt{c004} vs.\ \texttt{c003} & 0.012 & [-0.002, 0.027] & \ding{55} \\
Geometry & \texttt{c004} vs.\ \texttt{c001} & 0.007 & [-0.012, 0.025] & \ding{55} \\
Geometry & \texttt{c004} vs.\ \texttt{c020} & 0.006 & [-0.013, 0.024] & \ding{55} \\
\midrule
Texture & \texttt{c004} vs.\ \texttt{c001} & 0.032 & [0.015, 0.048] & \ding{51} \\
Texture & \texttt{c004} vs.\ \texttt{c020} & 0.020 & [0.003, 0.035] & \ding{51} \\
Texture & \texttt{c004} vs.\ \texttt{c003} & 0.017 & [0.002, 0.031] & \ding{51} \\
Texture & \texttt{c004} vs.\ \texttt{c032} & 0.006 & [-0.009, 0.021] & \ding{55} \\
\bottomrule
\end{tabular}
\end{table}

\paragraph{Why selection uses silver rather than expert labels.} Configuration selection is performed on silver labels because only the carried-forward \texttt{c004} design was ever scored against the 129-asset expert set; the screening judges were not run on the expert assets under the other 83 designs, so re-ranking configurations directly on expert labels would require new inference that this study does not perform. The expert-transfer route (Section~\ref{sec:stage3golden}) validates the choice indirectly---selecting on the full silver set and evaluating on the higher-quality expert labels independently returns \texttt{c004}---and the paired tests above bound how much any other top-cluster configuration could have shifted the selection.

\begin{figure}[tbp]
\centering
\includegraphics[width=\linewidth]{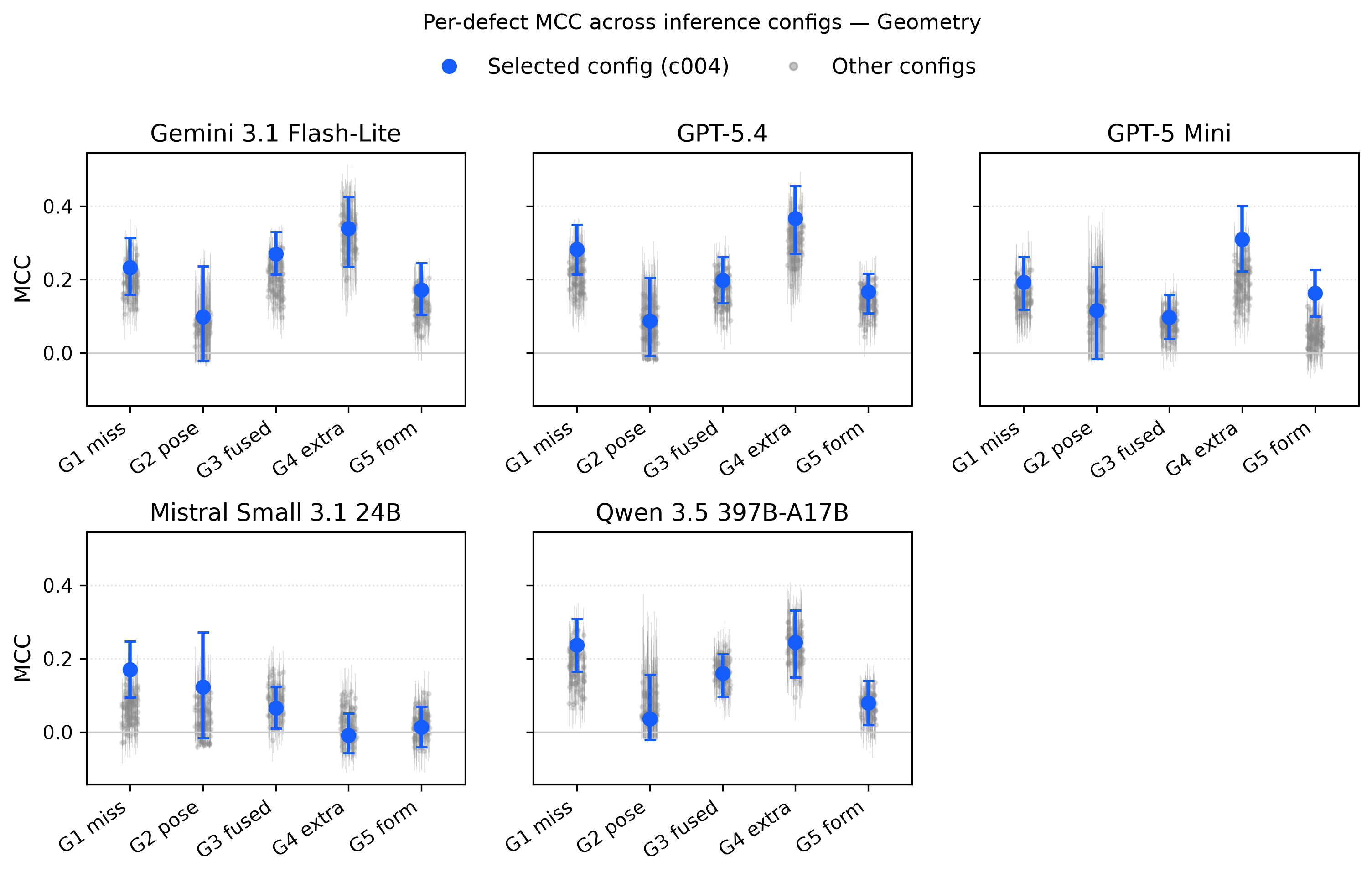}
\caption{Per-defect geometry MCC by configuration and screening model (gray: all 84 configs; blue: \texttt{c004}). Error bars are 95\% asset-cluster bootstrap CIs.}
\label{fig:config_mcc_geometry}
\end{figure}

\begin{figure}[tbp]
\centering
\includegraphics[width=\linewidth]{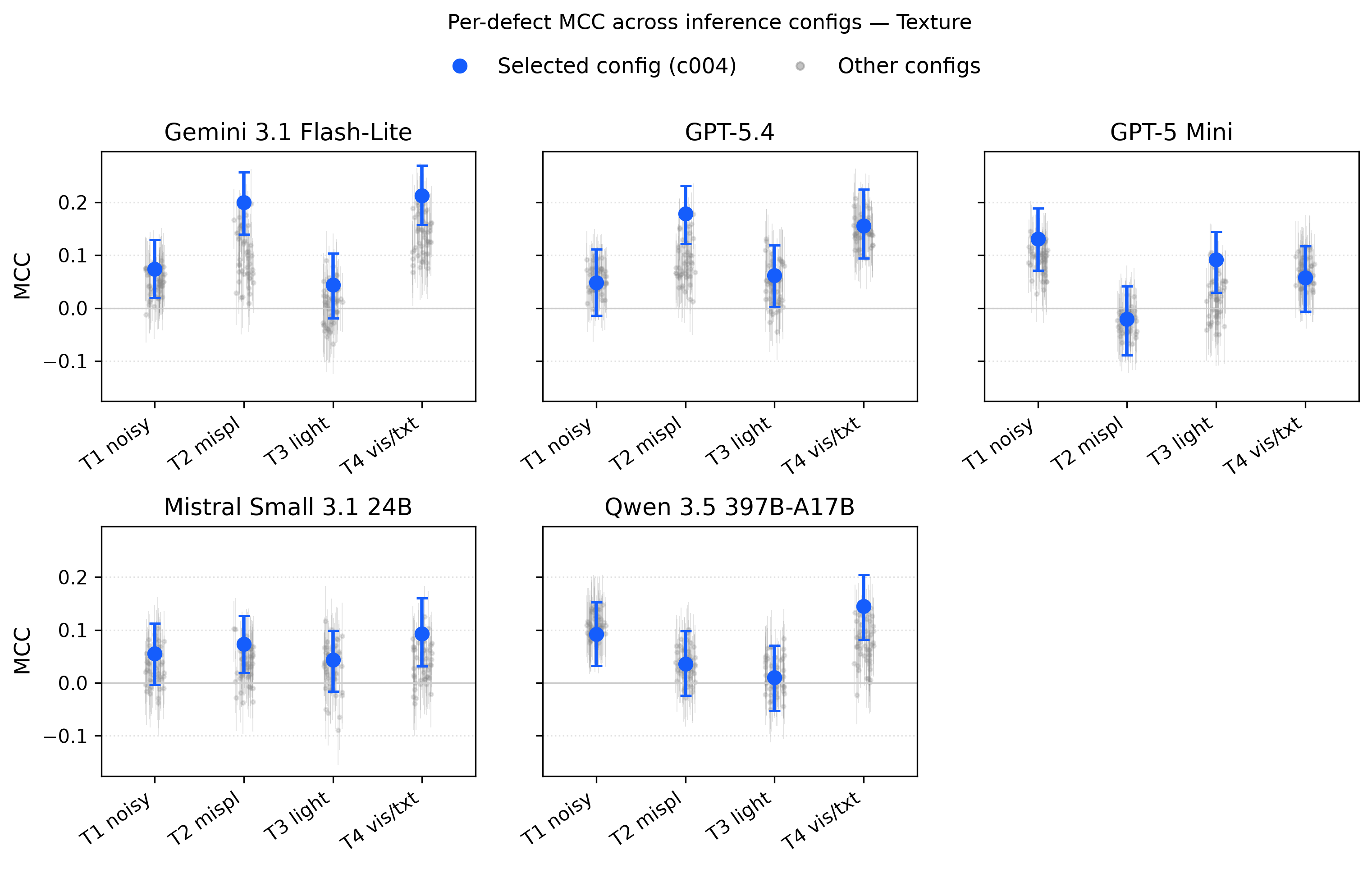}
\caption{Per-defect texture MCC by configuration and screening model (gray: all texture-eligible configs; blue: \texttt{c004}). Error bars are 95\% asset-cluster bootstrap CIs.}
\label{fig:config_mcc_texture}
\end{figure}

\paragraph{Per-defect view of configuration spread.} Figures~\ref{fig:config_mcc_geometry} and~\ref{fig:config_mcc_texture} show, for each screening model, the per-defect MCC (with bootstrap CIs) of every configuration in gray and the carried-forward \texttt{c004} in blue. \texttt{c004} sits at or near the top of the configuration spread on most geometry defects across models, while on texture the configurations are tightly bunched and largely overlapping---consistent with the small selection margins and the low texture ceiling.

\section{Per-defect model-comparison metrics}
\label{subsec:perdefect}
\begin{figure}[tbp]
\centering
\begin{subfigure}{0.49\linewidth}\centering
\includegraphics[width=\linewidth]{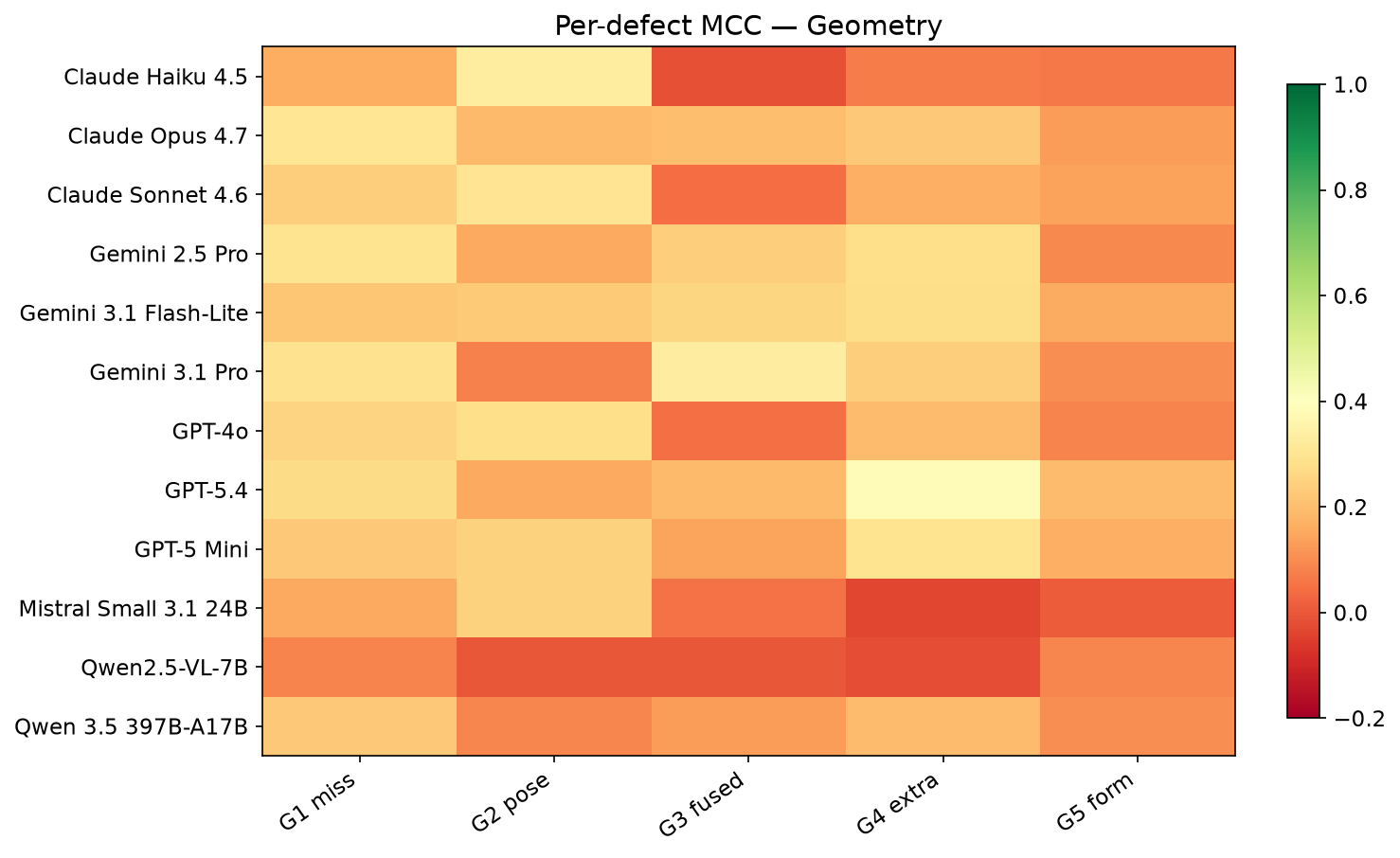}
\caption*{\scriptsize Geometry}
\end{subfigure}\hfill
\begin{subfigure}{0.49\linewidth}\centering
\includegraphics[width=\linewidth]{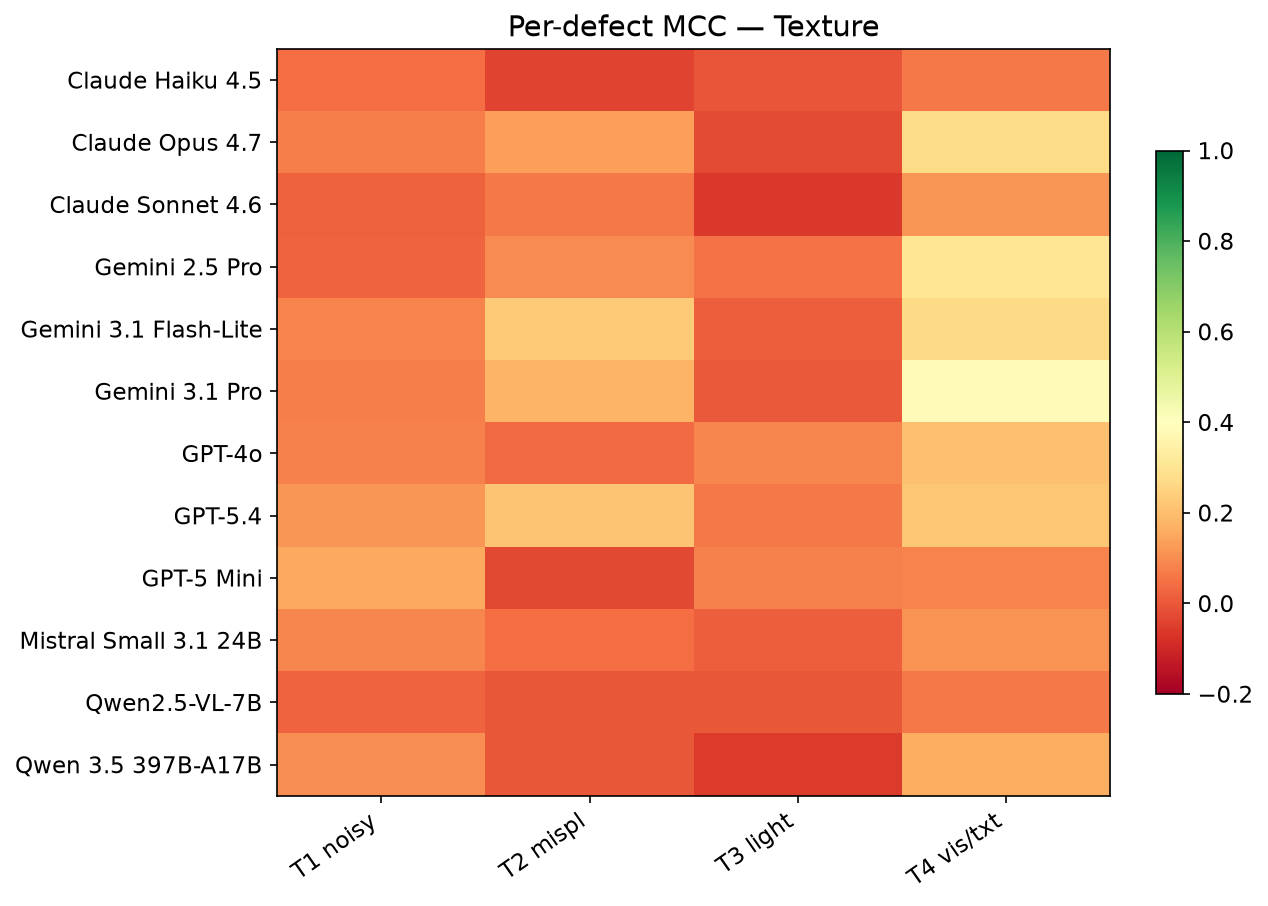}
\caption*{\scriptsize Texture}
\end{subfigure}
\caption{Per-defect MCC heatmaps (model $\times$ defect) on expert-agreement cells.}
\label{fig:golden_heatmap}
\end{figure}

\begin{figure}[tbp]
\centering
\begin{subfigure}{0.49\linewidth}\centering
\includegraphics[width=\linewidth]{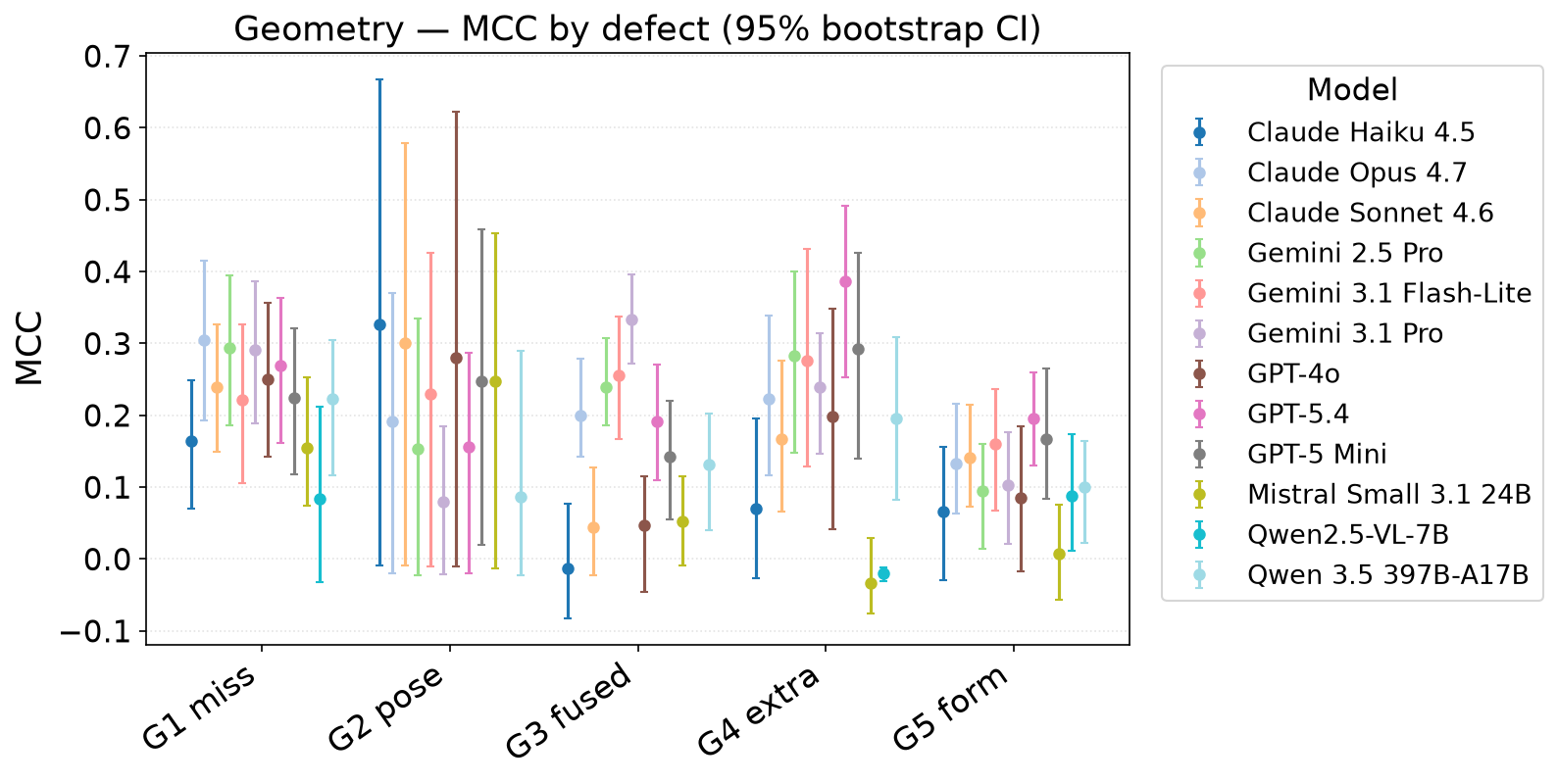}
\caption*{\scriptsize Geometry}
\end{subfigure}\hfill
\begin{subfigure}{0.49\linewidth}\centering
\includegraphics[width=\linewidth]{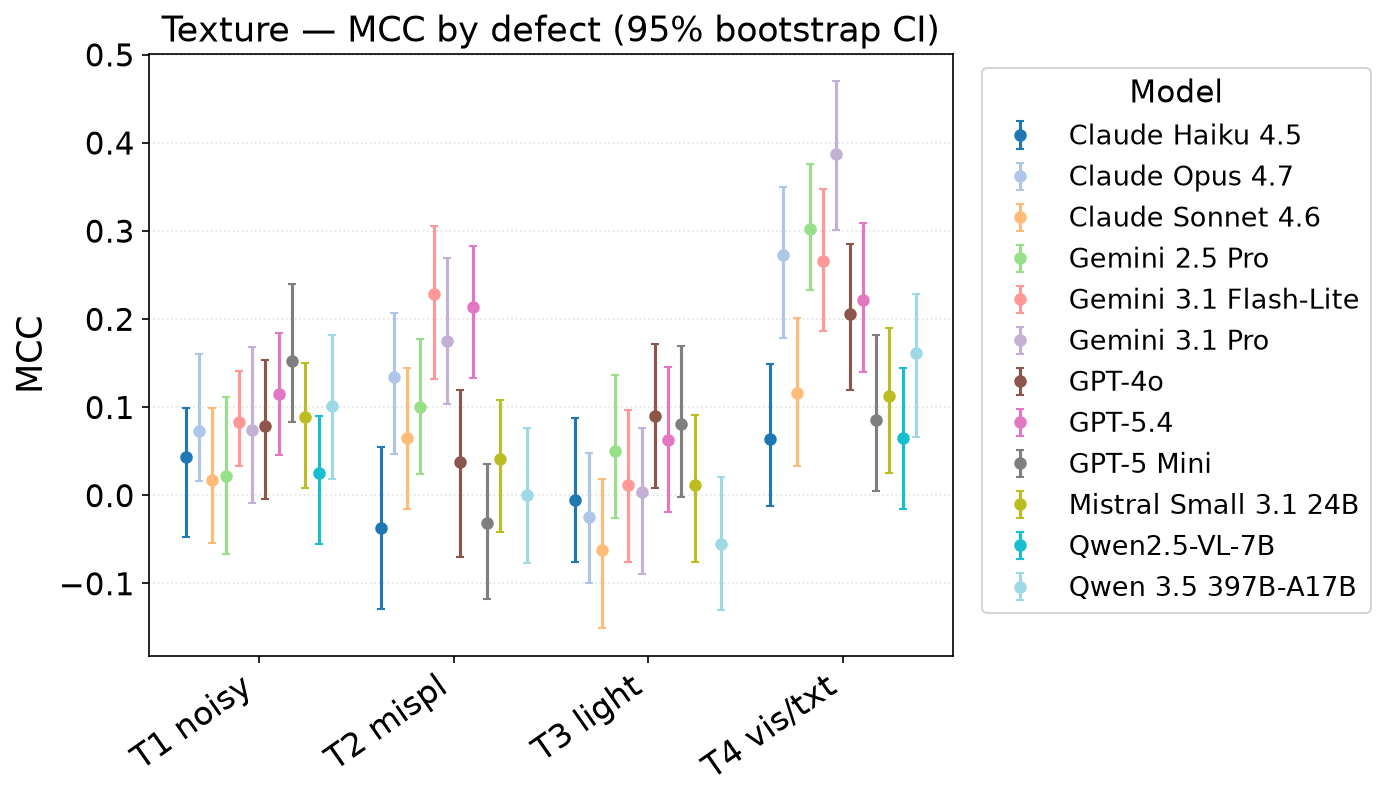}
\caption*{\scriptsize Texture}
\end{subfigure}
\caption{Per-defect MCC with 95\% asset-cluster bootstrap CI by model (expert-agreement cells).}
\label{fig:golden_defect_ci}
\end{figure}

Figure~\ref{fig:golden_heatmap} shows the per-defect MCC for every model on expert-agreement cells as a model$\times$defect matrix; Figure~\ref{fig:golden_defect_ci} shows the same per-defect MCC with asset-cluster bootstrap 95\% CIs. Per-defect F1 follows the same structure and is included in the released tables alongside MCC, precision, and recall.

\section{Expert label modes}
\label{subsec:expert_modes}
The expert split has exactly two expert labels per asset, so a disagreement cell (one pass, one fail) has no majority. We therefore construct the expert leaderboard from two complementary consensus targets, plus a per-expert diagnostic. \textbf{Agreement-only} (primary) keeps only the 877 unique defect cells where both experts agree (539 geometry, 338 texture); this isolates clean targets but excludes the 284 disagreement cells. \textbf{Either-expert union} marks a defect present if either expert flags it, covering the full grid of 1{,}161 unique cells (645 geometry, 516 texture); this is recall-oriented and gives full coverage. These two consensus targets are the ones reported in the main leaderboard (Table~\ref{tab:leaderboard_combined}). The \textbf{per-expert} diagnostic instead scores each VLM separately against expert~0 and expert~1 to expose sensitivity to the labeler choice; it is distinct from the \emph{single held-out expert} protocol used for the human-labeler comparison (Section~\ref{sec:human}, Table~\ref{tab:human}), which holds one expert out as the test labeler and scores both that expert and the VLMs against the other.

Table~\ref{tab:expert_modes} shows the top models under agreement-only and either-expert union. The rank-1 model is stable across modes (Gemini~3.1~Pro leads geometry, Gemini~3.1~Flash-Lite leads texture in both), but absolute MCC and the rank-2 ordering shift: the agreement filter changes the target, not just the sample size. Table~\ref{tab:expert_disagreement} breaks down the 284 disagreement cells by defect; Table~\ref{tab:expert_modes_full} lists the top-five models under each label mode.

\begin{table}[tbp]
\caption{Expert leaderboard top-3 by label mode (macro MCC, \texttt{c004}). Agreement-only uses both-experts-agree cells; union marks a defect if either expert flags it.}
\label{tab:expert_modes}
\centering
\footnotesize
\begin{tabular}{lll}
\toprule
Mode / aspect & Rank-1 & Rank-2 \\
\midrule
Agreement, geometry & Gemini 3.1 Pro (0.298) & Gemini 2.5 Pro (0.282) \\
Agreement, texture & Gemini 3.1 Flash-Lite (0.406) & Gemini 3.1 Pro (0.386) \\
Union, geometry & Gemini 3.1 Pro (0.273) & Gemini 3.1 Flash-Lite (0.199) \\
Union, texture & Gemini 3.1 Flash-Lite (0.281) & Gemini 3.1 Pro (0.235) \\
\bottomrule
\end{tabular}
\end{table}

\begin{table}[h]
\caption{Expert-expert disagreement cells by defect ($n{=}284$ of 1{,}161 unique cells).}
\label{tab:expert_disagreement}
\centering
\footnotesize
\begin{tabular}{llr}
\toprule
Aspect & Defect & Disagreement cells \\
\midrule
Geometry & Extra geometry & 8 \\
Geometry & Form or surface quality & 41 \\
Geometry & Fused or incomplete parts & 28 \\
Geometry & Missing parts & 25 \\
Geometry & Pose or placement mismatch & 4 \\
\midrule
Texture & Baked lighting or shadow & 47 \\
Texture & Visual-textual mismatch & 45 \\
Texture & Misplaced or overlapping texture & 43 \\
Texture & Noise, blur, or grain & 43 \\
\bottomrule
\end{tabular}
\end{table}

\begin{table}[h]
\caption{Expert leaderboard by label mode (macro MCC, top five models per mode; \texttt{c004}). Agreement-only uses both-experts-agree cells; union marks a defect if either expert flags it.}
\label{tab:expert_modes_full}
\centering
\scriptsize
\begin{tabular}{lllr}
\toprule
Label mode & Aspect & Model & Macro MCC \\
\midrule
Agreement-only & Geometry & Gemini 2.5 Pro & 0.247 \\
 & Geometry & Gemini 3.1 Pro & 0.229 \\
 & Geometry & Gemini 3.1 Flash-Lite & 0.221 \\
 & Geometry & Claude Opus 4.7 & 0.183 \\
 & Geometry & GPT-5.4 & 0.183 \\
\midrule
 & Texture & Gemini 3.1 Flash-Lite & 0.297 \\
 & Texture & Gemini 3.1 Pro & 0.290 \\
 & Texture & Claude Opus 4.7 & 0.230 \\
 & Texture & GPT-4o & 0.174 \\
 & Texture & Gemini 2.5 Pro & 0.172 \\
\midrule
Either-expert union & Geometry & Gemini 3.1 Pro & 0.273 \\
 & Geometry & Gemini 3.1 Flash-Lite & 0.199 \\
 & Geometry & Gemini 2.5 Pro & 0.175 \\
 & Geometry & GPT-5.4 & 0.172 \\
 & Geometry & Qwen 3.5 397B-A17B & 0.113 \\
\midrule
 & Texture & Gemini 3.1 Flash-Lite & 0.281 \\
 & Texture & Gemini 3.1 Pro & 0.235 \\
 & Texture & Gemini 2.5 Pro & 0.223 \\
 & Texture & Claude Opus 4.7 & 0.196 \\
 & Texture & GPT-4o & 0.182 \\
\bottomrule
\end{tabular}
\end{table}

\section{Cost and compute}
\label{app:cost}
Table~\ref{tab:accounting} summarizes API call and scored-cell counts by phase. Screening costs are dominated by the 84-design grid ($\sim$1.2M model calls on all 1{,}006 silver assets); final comparison costs are dominated by frontier model calls on multi-image inputs (2{,}838 expert calls + 12{,}078 silver holdout calls). We quantify cost as these API-call and scored-cell counts rather than as dollar figures, since per-call prices differ across providers and change over time.

\section{Reproducibility}
\label{app:repro}
Every results table and analysis figure in this paper is reproducible from the released labels and model predictions with no new model inference, except the TRELLIS cross-generator replication (Section~\ref{subsec:trellis_transfer}), which is reproduced by \texttt{experiments/stage1\_trellis/} in the post-submission analysis bundle. The public release includes the analysis code that recomputes the factor-importance estimates, the bootstrap configuration-selection stability and the paired \texttt{c004}-vs-neighbor comparisons (Appendix~\ref{subsec:selection}), the expert-consensus leaderboards, and the per-defect and baseline metrics reported here. The overview schematic (Figure~\ref{fig:overview}) is composed manually from the released per-defect prevalence statistics.

\paragraph{Rendering configuration.}
Views are rendered with \texttt{pyrender} on \texttt{trimesh} geometry in an offscreen buffer. Each mesh is recentered to the origin, scaled so its bounding sphere has unit radius, then placed at radius $2.0$; the camera distance is auto-fit to frame the object with a $1.10$ margin, capped to the maximum fit distance across the views of a protocol so all panels share one scale. The camera is a perspective camera with $\mathrm{yfov}{=}30^\circ$ and aspect ratio $1.0$. Scenes use flat ambient lighting only (\texttt{ambient\_light}${=}[1,1,1]$; no directional or image-based lighting); the background is black $(0,0,0)$, with meshes rendered to RGBA and composited over black. Outputs are 8-bit sRGB PNGs; no explicit gamma correction or tonemapping is applied. The six-view oblique protocol uses azimuths $\{45^\circ,135^\circ,225^\circ,315^\circ\}$ at $0^\circ$ elevation plus top ($+90^\circ$) and bottom ($-90^\circ$) views; the 14-view equatorial and 14-view multi-ring protocols add equatorial and multi-elevation rings respectively (exact angle lists in the released \texttt{factorial\_config.py}). Because these settings were fixed across all 84 designs, the study does not measure how judge agreement responds to background color or lighting; that sensitivity analysis requires new renders and is future work.

\begin{table}[h]
\caption{Rendering parameters, fixed across all inference designs.}
\label{tab:render_params}
\centering\footnotesize
\begin{tabular}{ll}
\toprule
Parameter & Value \\
\midrule
Renderer & \texttt{pyrender} + \texttt{trimesh}, offscreen \\
Per-view resolution & $512\times512$ (six-oblique grid $3\times2$, $1536\times1024$) \\
Camera & perspective, $\mathrm{yfov}{=}30^\circ$, aspect $1.0$ \\
Object framing & centered, unit-sphere normalized, radius $2.0$, $1.10$ fit margin \\
Lighting & flat ambient $[1,1,1]$; no directional/HDRI \\
Background & black $(0,0,0)$; RGBA composited on black \\
Color/encoding & 8-bit sRGB PNG; no explicit gamma/tonemap \\
\bottomrule
\end{tabular}
\end{table}

\paragraph{Depth and normal encodings.}
The depth channel is normalized \emph{per frame}: over pixels on the visible mesh (mask${>}0$) we take the min and max depth and render grey ${=}\ 255\,(1-(d-d_{\text{near}})/\text{span})$, i.e.\ inverted so the nearest surface is white and the farthest black, replicated across RGB. Because normalization is per frame rather than against a fixed near/far plane, depth greys are not comparable in absolute scale across views or assets. The normal channel is in view (camera) space with flat shading, encoded by the standard $\text{color}{=}(\mathbf{n}\cdot0.5+0.5)\cdot255$ XYZ$\to$RGB map. Alternative encodings---fixed near/far depth, or world-space/absolute normals---could change the geometry-only utility of these channels and are not explored here.

\paragraph{Chance-corrected agreement (robustness).}
Table~\ref{tab:perdef} reports silver Fleiss' $\kappa$, which deflates under extreme class imbalance. As a prevalence-robust companion we also compute Randolph's free-marginal $\kappa$ per defect: geometry macro $0.52$ vs.\ texture macro $0.18$ on silver---a starker geometry-over-texture gap than Fleiss' $\kappa$ because it removes the base-rate paradox (e.g.\ pose/placement rises from Fleiss $0.09$ to free-marginal $0.88$ at $1.2\%$ prevalence). The texture-below-geometry ordering is consistent across Fleiss' $\kappa$, Randolph's $\kappa$, and expert Cohen's $\kappa$, quantifying the texture label-noise ceiling.

\paragraph{Model versions.}
Table~\ref{tab:model_snapshots} lists the evaluated judges by provider and phase. Names are public model identifiers; screening and model-comparison runs were executed between 2026-05-21 and 2026-06-30 (evaluation configuration selected 2026-06-30). All proprietary models are accessed through hosted APIs, so exact served snapshots may drift over time and are not fully pinnable from the public names alone; to make results reproducible independent of that drift we release every model's raw predictions, and all tables and figures recompute from those predictions without new inference.

\begin{table}[h]
\caption{VLM judges evaluated, by provider and phase. ``Screen'' = eight-model factor-analysis sweep over all 84 designs; ``Compare'' = twelve-model comparison under \texttt{c004}.}
\label{tab:model_snapshots}
\centering\footnotesize
\begin{tabular}{lll}
\toprule
Model & Provider & Phase \\
\midrule
Gemini 3.1 Flash-Lite & Google & Screen + Compare \\
GPT-5 Mini & OpenAI & Screen + Compare \\
GPT-5.4 & OpenAI & Screen + Compare \\
Qwen 3.5 397B-A17B & Alibaba (Qwen) & Screen + Compare \\
Mistral Small 3.1 24B & Mistral & Screen + Compare \\
Gemini 3.1 Pro & Google & Screen + Compare \\
Claude Sonnet 4.6 & Anthropic & Screen + Compare \\
Claude Haiku 4.5 & Anthropic & Screen + Compare \\
Gemini 2.5 Pro & Google & Compare \\
Claude Opus 4.7 & Anthropic & Compare \\
GPT-4o & OpenAI & Compare \\
Qwen2.5-VL-7B & Alibaba (Qwen) & Compare \\
\bottomrule
\end{tabular}
\end{table}

\end{document}